\documentclass[10pt,journal,compsoc]{IEEEtran}

\usepackage{amssymb}
\usepackage{amsfonts}
\usepackage{amsmath}
\usepackage{cite}
\usepackage{bbm}
\usepackage[dvips]{graphicx}
\usepackage{color}
\usepackage{comment}
\usepackage{makecell}
\usepackage{epstopdf}
\usepackage{subfigure}
\usepackage{algorithm}
\usepackage{algorithmic}
\usepackage{booktabs}
\usepackage{ulem}
\usepackage{multirow}
\usepackage{threeparttable}
\usepackage{url}
\let\emph\textit
\begin{document}
\newtheorem{definition}{\it Definition}
\newtheorem{theorem}{\bf Theorem}
\newtheorem{lemma}{\it Lemma}
\newtheorem{corollary}{\it Corollary}
\newtheorem{remark}{\it Remark}
\newtheorem{example}{\it Example}
\newtheorem{case}{\bf Case Study}
\newtheorem{assumption}{\it Assumption}
\newtheorem{property}{\it Property}
\newtheorem{proposition}{\it Proposition}

\newcommand{\hP}[1]{{\boldsymbol h}_{{#1}{\bullet}}}
\newcommand{\hS}[1]{{\boldsymbol h}_{{\bullet}{#1}}}

\newcommand{\ba}{\boldsymbol{a}}
\newcommand{\baq}{\overline{q}}
\newcommand{\bA}{\boldsymbol{A}}
\newcommand{\bb}{\boldsymbol{b}}
\newcommand{\bB}{\boldsymbol{B}}
\newcommand{\bc}{\boldsymbol{c}}
\newcommand{\bcO}{\boldsymbol{\cal O}}
\newcommand{\be}{\boldsymbol{e}}
\newcommand{\bh}{\boldsymbol{h}}
\newcommand{\bH}{\boldsymbol{H}}
\newcommand{\bl}{\boldsymbol{l}}
\newcommand{\bm}{\boldsymbol{m}}
\newcommand{\bn}{\boldsymbol{n}}
\newcommand{\bo}{\boldsymbol{o}}
\newcommand{\bO}{\boldsymbol{O}}
\newcommand{\bp}{\boldsymbol{p}}
\newcommand{\bq}{\boldsymbol{q}}
\newcommand{\bR}{\boldsymbol{R}}
\newcommand{\bs}{\boldsymbol{s}}
\newcommand{\bS}{\boldsymbol{S}}
\newcommand{\bT}{\boldsymbol{T}}
\newcommand{\bw}{\boldsymbol{w}}

\newcommand{\balpha}{\boldsymbol{\alpha}}
\newcommand{\bbeta}{\boldsymbol{\beta}}
\newcommand{\bOmega}{\boldsymbol{\Omega}}
\newcommand{\bTheta}{\boldsymbol{\Theta}}
\newcommand{\bphi}{\boldsymbol{\phi}}
\newcommand{\btheta}{\boldsymbol{\theta}}
\newcommand{\bvarpi}{\boldsymbol{\varpi}}
\newcommand{\bpi}{\boldsymbol{\pi}}
\newcommand{\bpsi}{\boldsymbol{\psi}}
\newcommand{\bxi}{\boldsymbol{\xi}}
\newcommand{\bx}{\boldsymbol{x}}
\newcommand{\by}{\boldsymbol{y}}

\newcommand{\cA}{{\cal A}}
\newcommand{\bcA}{\boldsymbol{\cal A}}
\newcommand{\cB}{{\cal B}}
\newcommand{\cE}{{\cal E}}
\newcommand{\cG}{{\cal G}}
\newcommand{\cH}{{\cal H}}
\newcommand{\bcH}{\boldsymbol {\cal H}}
\newcommand{\cK}{{\cal K}}
\newcommand{\cM}{{\cal M}}
\newcommand{\cO}{{\cal O}}
\newcommand{\cR}{{\cal R}}
\newcommand{\cS}{{\cal S}}
\newcommand{\dcS}{\ddot{{\cal S}}}
\newcommand{\ds}{\ddot{{s}}}
\newcommand{\cT}{{\cal T}}
\newcommand{\cU}{{\cal U}}
\newcommand{\wt}[1]{\widetilde{#1}}

\newcommand{\mA}{\mathbb{A}}
\newcommand{\mE}{\mathbb{E}}
\newcommand{\mG}{\mathbb{G}}
\newcommand{\mR}{\mathbb{R}}
\newcommand{\mS}{\mathbb{S}}
\newcommand{\mU}{\mathbb{U}}
\newcommand{\mV}{\mathbb{V}}
\newcommand{\mW}{\mathbb{W}}

\newcommand{\uq}{\underline{q}}
\newcommand{\ubq}{\underline{\boldsymbol q}}

\newcommand{\red}[1]{\textcolor[rgb]{1,0,0}{#1}}
\newcommand{\gre}[1]{\textcolor[rgb]{0,1,0}{#1}}
\newcommand{\blu}[1]{\textcolor[rgb]{0,0,0}{#1}}

\title{Time-sensitive Learning for Heterogeneous Federated Edge Intelligence}

\author{
Yong~Xiao, \IEEEmembership{Senior~Member,~IEEE}, Xiaohan~Zhang, Yingyu Li, Guangming~Shi, \IEEEmembership{Fellow, IEEE}, Marwan Krunz, \IEEEmembership{Fellow, IEEE}, Diep N. Nguyen, \IEEEmembership{Senior~Member,~IEEE}, and Dinh Thai Hoang, \IEEEmembership{Senior~Member,~IEEE} 

\thanks{*This work has been accepted at IEEE Transactions on Mobile Computing. Copyright may be transferred without notice, after which this version may no longer be accessible. 


Y. Xiao is with the School of Electronic Information and Communications at the Huazhong University of Science and Technology, Wuhan 430074, China, also with the Peng Cheng Laboratory, Shenzhen, Guangdong 518055, China, and also with the Pazhou Laboratory (Huangpu), Guangzhou, Guangdong 510555, China (e-mail: yongxiao@hust.edu.cn).

X. Zhang is with the School of Electronic Information and Communications at the Huazhong University of Science and Technology, Wuhan, China 430074 (e-mail: xiaohanzhang@hust.edu.cn).

Y. Li is with the School of Mechanical Engineering and Electronic Information, China University of Geosciences, Wuhan, China 430074 (e-mail: liyingyu29@cug.edu.cn).

G. Shi is with the Peng Cheng Laboratory, Shenzhen, Guangdong 518055, China, also with the School of Artificial Intelligence, the Xidian University, Xi’an, Shaanxi 710071, China, and also with
the Pazhou Laboratory (Huangpu), Guangzhou, Guangdong 510555, China (e-mail: gmshi@xidian.edu.cn).

M. Krunz is with the Department of Electrical and Computer Engineering, the University of Arizona, Tucson, AZ 85721 (e-mail: krunz@arizona.edu).

D. Nguyen and D. Hoang are with the School of Electrical and Data Engineering, University of Technology Sydney Faculty of Engineering and Information Technology, 120558 Sydney, New South Wales, Australia (e-mail: \{hoang.dinh, diep.nguyen\}@uts.edu.au).



}
}
\maketitle
\begin{abstract}
Real-time machine learning (ML) has recently attracted significant interest due to its potential to support instantaneous learning, adaptation, and decision making in a wide range of application domains, including self-driving vehicles, intelligent transportation, and industry automation. In this paper, we investigate real-time ML in a federated edge intelligence (FEI) system, an edge computing system that implements federated learning (FL) solutions based on data samples collected and uploaded from decentralized data networks, e.g., Internet-of-Things (IoT) and/or wireless sensor networks. FEI systems often exhibit heterogenous communication and computational resource  distribution, as well as  non-i.i.d. data samples arrived at different edge servers, resulting in long model training time and inefficient resource utilization. Motivated by this fact, we propose a time-sensitive federated learning (TS-FL) framework
to minimize the overall run-time for collaboratively training a shared ML model with desirable accuracy. Training acceleration solutions for both TS-FL with synchronous coordination (TS-FL-SC) and asynchronous coordination (TS-FL-ASC) are investigated. To address the straggler effect in TS-FL-SC, we develop an analytical solution to characterize the impact of selecting different subsets of edge servers on the overall model training time. A server dropping-based solution is proposed to allow some slow-performance edge servers to be removed from participating in the model training if their impact on the resulting model accuracy is limited. A joint optimization algorithm is proposed to minimize the overall time consumption of model training by selecting participating edge servers, the local epoch number (the number of model training iterations per coordination), and the data batch size (the number of data samples for each model training iteration). Motivated by the fact that data samples at the slowest edge server may exhibit special characteristics that cannot be removed from model training, we develop an analytical expression to characterize the impact of both staleness effect of asynchronous coordination and  straggler effect of FL on the time consumption of TS-FL-ASC. We propose a load forwarding-based solution that allows a slow edge server to offload part of its training samples to trusted edge servers with higher processing capability. We develop a hardware prototype to evaluate the model training time of a heterogeneous FEI system. Experimental results show that our proposed TS-FL-SC and TS-FL-ASC can provide up to 63\% and 28\% of reduction, in the overall model training time, respectively, compared with traditional FL solutions.
%
%
\end{abstract}


\section{Introduction}

The proliferation of smart applications that require real-time learning, adaptation, and decision making, such as self-driving vehicles, intelligent transportation systems\cite{XY2021AdaptiveFog}, and Tactile Internet\cite{XY2018TactileInternet}, has significantly increased the demand for machine learning (ML)-enabled solutions that support fast proactive learning and model construction 
based on large datasets. 
The so-called {\it real-time ML} has recently attracted significant interest due to its potential to quickly solve unfamiliar problems and self-adapt to unknown situations\cite{nishihara2017real}. Existing ML solutions are often computational demanding and rely on large datasets to be collected and pre-loaded into a centralized location, e.g., cloud data center, and are therefore infeasible for real-time operation. Recent concerns over data privacy further exacerbate the challenge, as some ``local" users do not wish to disclose their datasets to the high-performance data center due to privacy concerns or regulation restrictions.

A highly promising solution is federated learning (FL)\cite{McMahan2017FLfirstpaper}, an emerging ML framework that enables multiple servers to jointly train a shared model without exposing the private data owned by individual users. The key idea is to allow edge servers to train local ML models using their local data samples and periodically coordinate with each other through their locally trained model parameters\cite{yang2019federated,lim2020federated}. 
Federated edge intelligence (FEI)\cite{mills2019communication,deng2020edge}, is an emerging paradigm that focuses on the implementation of FL-based solutions in edge computing systems. FEI has recently been promoted by both industry and academia as a key candidate solution in the next generation mobile technologies, e.g., beyond 5G and 6G\cite{xiao2020optimizing,selfxiao}.

Although FEI has the potential to protect data privacy, enable parallel computation, and avoid the delay of transporting raw data samples to the cloud data center, it brings new challenges when applied to real-time ML applications. First, the performance of FEI can be affected by both computation and communication-related performance metrics, including the processing capabilities of edge servers (clients), data collection and uploading delay of data collecting devices, frequency of coordination during model training, and communication bandwidth between edge servers and the coordinator. These challenges significantly increase the complexity to search for the optimal solution to minimize the overall run-time of training the ML model. For example, reducing the number of local training iterations between successive coordination rounds may accelerate the convergence speed. It however increases the frequency of model coordination which will result in high communication delay.
Similarly, processing more local training samples during each iteration reduces the required number of model coordination rounds to reach the target model accuracy. 
However, it also slows down the data loading speed and increases the computational time in each iteration.
How to jointly optimize computation and communication-related parameters to minimize the overall time consumed for model training is still an open problem.
Furthermore, \blu{traditional FL, especially FL with synchronous coordination, is known to suffer from the {\it straggler effect}\cite{stich2018local}, i.e., the overall ML model training delay is dominated by the slowest edge server. There is still lacking a comprehensive solution that can alleviate the straggler effect and expedite the model training speed in FEI systems under both system heterogeneity, i.e., various hardware and software resources available at different edge servers, and data heterogeneity, i.e., non-i.i.d. distributions of data samples across different edge servers. Recent studies have investigated FL with asynchronous coordination in which each edge server does not have to wait for other edge servers, but can request an instantaneous model update from the coordinator. However, previous works suggest that the asynchronous coordination often results in degraded convergence performance, compared to the synchronous coordination solutions\cite{Han2020AdaptiveFL}, due to the {\it staleness effect}, that is the increasing difference in the model updating frequencies between edge servers with different computational speeds and communication bandwidths will result in out-of-date models at some slow edge servers, especially when the total number of model coordination rounds becomes large. }



\begin{figure}
  \centering
  \includegraphics[width=9 cm]{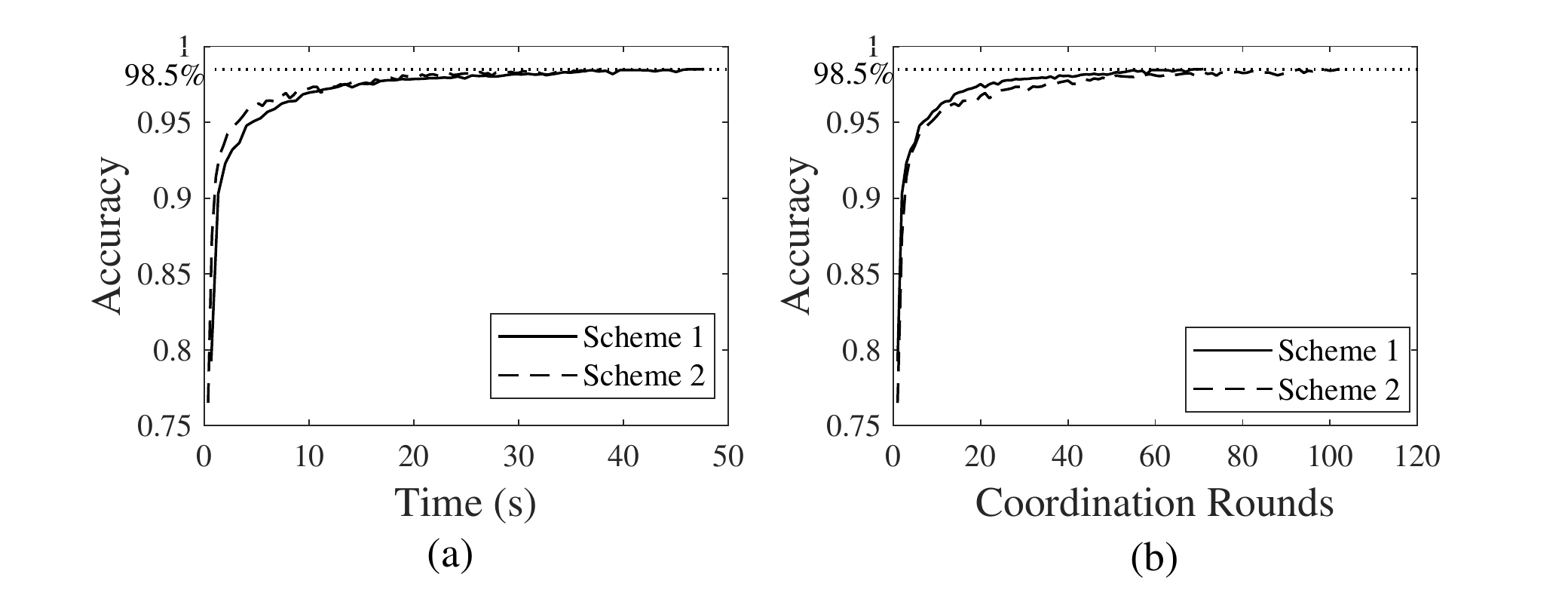}
  \vspace{-0.2in}
  \caption{\blu{(a) Model accuracy vs. training time, and (b) model accuracy vs. number of coordination rounds under two FL model training schemes: {\it Scheme 1} ($200$ local epoch number, $20$ mini-batch size) and {\it Scheme 2} ($200$ local epoch number, $10$ mini-batch size). The target accuracy of the trained model is set to $98.5\%$.}}\label{LantencyandRound}
\end{figure}

In this paper, we propose the time-sensitive FL (TS-FL) framework, \blu{which aims at minimizing the overall runtime for training an ML model with the guaranteed accuracy, the percentage of the correct predictions among all the predictions made by the trained model. In contrast to existing works that focus on reducing the required number of stochastic gradient descent (SGD) coordination rounds for model training \cite{wang2020optimizing}, the overall runtime needed to construct an ML model  is a much more important performance metric in many real-world applications, especially the time-sensitive applications. We observe that, in many practical systems, the required number of SGD coordination rounds and the overall runtime for model training may not follow the same trend, i.e, a smaller number of SGD rounds does not always mean shorter runtime for training the model with the same accuracy. To shed more light on this observation, we conduct experiments on a hardware prototype that consists of $20$ Raspberry Pis (version 4B) serving as edge servers. These mini-computers are connected to a dedicated Raspberry Pi, which serves as the coordinator, via a Wi-Fi router. We compare the required time consumption  (in Fig. \ref{LantencyandRound}(a)) and the number of SGD coordination rounds (in Fig. \ref{LantencyandRound}(b)) for training a shared handwritten digit recognition model based on the MINIST dataset with the same requirement in terms of the model prediction accuracy (e.g., $98.5\%$) under two different schemes of model training parameters, labelled as {\it Schemes 1} and {\it 2}, respectively. We can observe that, although {\it Scheme 2} requires a smaller number of SGD coordination rounds than {\it Scheme 1} for training the model with the required accuracy, it consumes more time for model training than {\it Scheme 1}. In other words, the number of SGD rounds cannot fully characterize the overall runtime of the learning process in a heterogeneous FEI system.}

Motivated by the above observation, in this paper, 
we investigate 
model training time-acceleration solutions for two types of TS-FLs: TS-FL with synchronous coordination (TS-FL-SC) and TS-FL with asynchronous coordination (TS-FL-ASC). In TS-FL-SC, the coordinator updates the global model when it receives the local models from all the edge servers selected for participating in model training. 
To alleviate the straggler effect of TS-FL-SC, we develop an analytical solution that characterizes the impact of selecting different subsets of edge servers on the overall model training time. We then develop a server dropping-based solution in which some slow-performing edge servers are removed from participating in model training so long as their impact on the model accuracy is limited.  A novel algorithm is proposed to jointly optimize the selection of participating edge servers, number of local epochs (i.e., number of model training iterations between two successive coordination rounds), and data batch size (data samples used for each local model training iteration) so as to minimize the overall time of model training. Motivated by the fact that, in some scenarios, data samples at the slowest edge server may exhibit special characteristics that are essential for resulting model accuracy, we consider the TS-FL-ASC, where model coordination among edge servers can be asynchronous. In particular, for TS-FL-ASC, we develop an analytical expression to characterize the {\it staleness effect} of asynchronous coordination systems. 
To reduce staleness and alleviate the straggler effect in this case, we propose a load forwarding-based solution that allows slow edge servers to offload part of their training datasets to other trusted edge servers with high processing capabilities. 
Extensive experiments are conducted based on a hardware prototype consisting of different versions of Raspberry Pi mini-computers (versions 4B and 3A), which we use to simulate model training time under various scenarios. 

\blu{To the best of our knowledge, this is the first work that investigates the time-sensitive FEI systems with precision guarantee by jointly optimizing server dropping, multi-variate parameter control, as well as asynchronous load offloading under both system heterogeneity and data heterogeneity.}

The main contributions of this paper are as follows:
\begin{itemize}
\item[(1)]{\bf Time-sensitive FL for Heterogeneous FEI:} We provide a unified analytical model that quantifies the impact of different training parameters on the time needed for different steps throughout the entire model training process for both TS-FL-SC and TS-FL-ASC.

\item[(2)]\blu{{\bf New Theoretical Results and an Optimization Algorithms:} We derive new theoretical results as well as optimization solutions for both TS-FL-SC and TS-FL-ASC. In particular, for TS-FL-SC, we derive an analytical upper bound to characterize the overall model training time under different selections of participating edge servers as well as combinations of their model training parameters. We also develop an optimization algorithm that can select the optimal subset of participating edge servers and their parameters so as to minimize the model training time for an FEI system with both system and data heterogeneity. For TS-FL-ASC, we derive a new theoretical convergence result with non-independent and identically distributed (non-i.i.d.) data samples at edge servers. We propose a load forwarding-based solution in which slow edge servers can offload part of their training datasets to trusted edge servers with higher processing capability. We prove that the training time minimization problem based on our derived bound is biconvex and then develop a novel algorithm to select the optimal subset of participating edge servers as well as their model training parameters that can minimize the overall model training time.} 

%

\item[(3)]{\bf  Extensive Experiments Based on a Hardware Platform:} We develop a hardware platform consisting of 23 low-cost mini-edge servers (Raspberry Pi mini-computers) with various computing capabilities. Extensive experiments are conducted to evaluate the model training time under various setups. Experimental results have shown that our proposed TS-FL-SC and TS-FL-ASC can, respectively, provide up to $59.31\%$ and $57.24\%$ of reduction in the overall model training time compared to the existing FL solutions.
\end{itemize}

The remainder of this paper is organized as follows. Related works are reviewed in Section \ref{sec_rw}. We introduce the system model and problem formulation in Section \ref{sec_systemmodel}. 
TS-FL-SC and TS-FL-ASC are  introduced in Sections \ref{Sec_S_TSFL} and \ref{sec_A_TSFL}, respectively. Numerical results are presented in Section \ref{sec_simultaion}, and  the paper is concluded in Section \ref{sec_conclusion}.

\section{Related Work}\label{sec_rw}

\noindent{\textbf{FEI in Wireless Networks:}}  
By integrating federated learning into edge intelligence, FEI enables collaborative learning and model construction based on decentralized datasets across a large networking system\cite{selfxiao}. 
Most existing works have focused on optimizing the allocation of communication and computational resources to improve the efficiency of FEI systems. 
The authors in \cite{wang2019adaptive} developed algorithms that can dynamically control the frequency of global aggregation at edge servers to improve the resource utilization and also reduce the energy consumption of a 5G-based edge networking system. Recently, FEI has been also extended into semantic communications\cite{shi2021semantic}, especially the implicit semantic-aware networks which allows the networking systems to automatically infer hidden information such as hidden rules and mechanisms from the communicated messages\cite{XY2023CollaberativeJSAC, XY2022ReasononAir, XY2022ITWSSC}. Although these works have shown significant progress in improving the communication and computational efficiency of FEI networks, they mainly focused on the impact of a single parameter, e.g., model updating frequency, local epoch number, or the selection of edge servers for participating model training, etc. Analytical frameworks that can characterize the impact of various combinations of key parameters on the performance of FEL networking systems are still lacking.

\noindent{\textbf{FEI for Heterogeneous Networks:}} System heterogeneity refers to differences in computation, communication, and storage capacities among edge devices. It is one of the major causes of the straggler effect in FEI \cite{li2020federated,chen2019efficient}. Server (device/client) sampling \cite{li2018federated,McMahan2017FLfirstpaper} is one approach to address system heterogeneity in FEI. For example, the authors in \cite{nishio2019client} adopted an active server sampling method that allows the coordinator to aggregate as many locally trained model parameters as possible within a predefined time window so as to mitigate the influence of heterogeneous devices. However, these works are under the assumption of static system characteristics, which lacking the ability to handle device-specific latency in many practical applications. Data heterogeneity, which reflects the diversity of local data samples across different edge servers in terms of data type and statistical distribution may further aggravate the complexity of problem modeling, theoretical analysis, and evaluation of solutions of an FEI system\cite{smith2017federated,jeong2018communication}. Recent works leveraging on meta-learning and multitask learning \cite{corinzia2019variational,eichner2019semi,khodak2019adaptive} provide feasible solutions for modeling heterogeneous data. For instance, the authors in \cite{corinzia2019variational} proposed a Bayesian network to model the star topology of an FEI system to mitigate  the adverse effect of data heterogeneity. However, most of these solutions are expensive to implement and often involve solving complex non-convex problems, making them impractical for large scale FEI networks.

\noindent{\textbf{FL with Asynchronous Coordination:}} 
Due to the heterogeneity of devices and local datasets, FL with synchronous coordination suffers from the ``straggler effect" that may deteriorate the convergence performance. There are already some recent works trying to alleviate the straggler effect in FEI systems by considering the asynchronous coordination solutions for FL. For example,  the authors in \cite{li2018federated} proposed  FedProx, a variant of FedAvg, in which a proximal term is proposed to balance the trade-off between data and system heterogeneity in FL \cite{li2020federated} as well as ensure convergence for both convex and nonconvex loss functions.  
The authors in \cite{xie2019asynchronous} proposed an asynchronous federated optimization algorithm that could mitigate the staleness of edge servers and guarantee near-linear convergence to a global optimal solution. 
In \cite{stich2018local}, the authors proposed an asynchronous local SGD algorithm to overcome the communication bottlenecks in a large FEI system with i.i.d. distributed data samples across edge servers. 

\begin{figure*}
  \centering
  \includegraphics[width=16 cm]{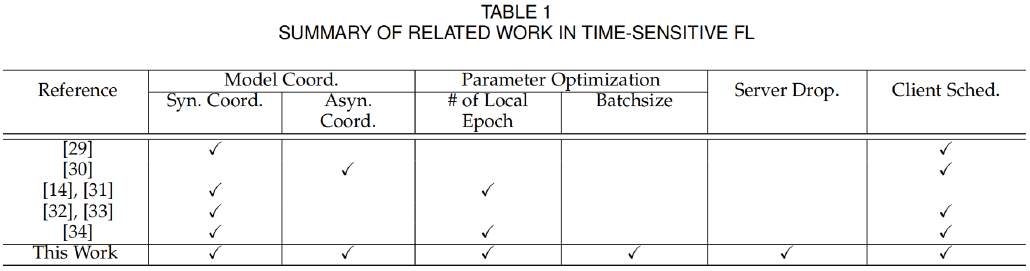}\\
\end{figure*}

\noindent{\blu{\textbf{Time-sensitive FL:}}}
With the popularity of time-sensitive smart applications and services, acceleration solutions that can reduce the overall model training time of FL has attracted significant interest recently. Most of existing works focus on optimizing the model parameters to accelerate the training speed. For example, in \cite{wang2019adaptive,wang_adaptive_2019}, the authors optimized the number of local epochs to reduce the runtime of model training. In \cite{Li2022Power}, the authors proposed a novel multi-armed bandit-based solution to alleviate the straggler effect of FL by optimizing the scheduling of edge server. In \cite{xia2020Multi}, the authors proposed a multi-armed bandit-based online edge server scheduling solution that can accelerate the raining speed without knowing the channel state information and statistic characteristics of edge servers.  In \cite{luo2020cost}, the authors jointly optimized the number of local epochs, scheduling of edge servers, and the maximum number of coordination rounds to minimize the energy consumption and learning time. Different from these existing works, in this paper, we introduce a unified framework for FL with both synchronous coordination and  asynchronous coordination. We propose a framework that can jointly optimize the edge server selection, number of local epochs, batchsizes, as well as training load re-distribution to address the straggler effect for FL with synchronous coordination as well as the staleness effect for FL with asynchronous coordination. We have verified the effectiveness of our proposed framework both theoretically and practically. We summarize the main difference between the existing works and ours in Table 1. 

\section{System Model}
\label{sec_systemmodel}
We first introduce the basic components and the processing steps of an FEI system. We then discuss the possible time consumption in each step of the training process and finally formulate the optimization problem for time-sensitive learning of an FEI system. 

\subsection{FEI Model}

We consider an FEI system consisting of three major components:
\blu{
\begin{itemize}
    \item[(1)] {\it Data collecting network: } It consists of a large number of devices that can collect local data samples to be uploaded to the corresponding edge servers.
    \item[(2)] {\it Edge servers: } A set $\mathcal{K}=\{1,2,...,K\}$ of $K$ edge servers are deployed to store and process the data samples uploaded from the data collecting devices. In this paper, we assume that each edge server is associated with an exclusive set of data collecting devices. Let ${\cal D}_k$ be the distribution of the data samples arrived at the $k$th edge server for $k \in \cal K$. Generally speaking, data samples arrived at different edge servers follow different distributions, i.e., we have ${\cal D}_k \neq {\cal D}_j$ for $k \neq j$ and $k,j \in \cal K$.
    \item[(3)] {\it Coordinator: } It coordinates the model training between edge servers via their intermediate results. It can be deployed at the cloud data center or one of the edge servers. All or a subset of edge servers can be selected to upload the intermediate model training results, e.g., model parameters, to the coordinator once in a while. These edge servers will wait for the coordinator to feedback an updated result, e.g., aggregation of all the received result in the FedAvg algorithm\cite{smith2017federated}, before continuing with their model training process.   
\end{itemize}
}

We consider the standard FL-based model training process described as follows. At the beginning of process, a subset of edge servers can download an initial model from the coordinator. Each edge server will then train a local model based on its local data samples and coordinate its local model training process with others once in a while. We focus on the real-time data uploading and training process. During each round of model training, an edge server first receives a batch of data samples from its associated data collecting devices and updates its local model based on this batch before coordinating with other edge servers.

The main objective 
is to collaboratively train a shared model that minimizes a global objective function $F(\boldsymbol w)$, corresponding to the weighted sum of the local objective functions of edge servers. Formally, we can write
\begin{eqnarray}\label{FL_function}
  \min_{\boldsymbol w \in {\mathbb R}^d} \{F(\boldsymbol w) &=& \sum_{k\in \cal K} p_k F_k(\boldsymbol w)\}
\end{eqnarray}
where $F_k(\boldsymbol w)$ is the local objective function of edge server $k$, $\boldsymbol w$ is the set of parameters of the model with dimensional size $d$, and $p_k$ is the weight of edge server $k$, where $0 \leq p_k \leq 1$ and $\sum_{k=1}^{K} p_k =1$. In some settings, $p_k$ can represent the relative portion of the local dataset of each edge server $k$\cite{McMahan2017FLfirstpaper, yang2020federated}. 

$F_k(\boldsymbol w)$ corresponds to the average loss of prediction over all data samples:
\begin{eqnarray}\label{loss_func}
  F_k(\boldsymbol w) &=& \mathbb{E}_{{x}_{k,j} \sim {\cal D}_k}[{l_k(\boldsymbol w;x_{k,j})}]
\end{eqnarray}
where ${l_k(\boldsymbol w;x_{k,j})}$ is the loss function of edge server $k$ with parameters $\boldsymbol w$ and local data sample $x_{k,j}$.

In a standard FL with synchronous coordination, problem (\ref{FL_function}) is solved by allowing each edge server $k$ to employ local SGDs to iteratively update its local parameters ${\boldsymbol w}_k$ and periodically upload its model to the coordinator for  model aggregation ${\boldsymbol w}=\sum_{k \in \cal{K}}p_k {\boldsymbol w}_{k}$.

In this paper, we consider a more general setting that can support both synchronous and asynchronous coordination. We use subscript $t$ to denote the $t$th iteration of the local model training process at an edge server. Let ${\cal I}^k$ be the set of iteration steps that edge server $k$ uploads its local model to the coordinator. Let ${\boldsymbol v}_{k,t}$ be the updated model received by edge server $k$ from the coordinator during iteration $t$, $t\in {\cal I}^k$. We can write the model updating process as follows
\begin{equation}
{\boldsymbol w}_{k,t+1}=\left\{
\begin{aligned}
& {\boldsymbol w}_{k,t}-\eta_t{\boldsymbol g}_{k,t},  &\mbox{if }t+1 \notin \mathcal{I}^k  \\
 & {\boldsymbol v}_{k,t+1},  &\mbox{if }t+1 \in \mathcal{I}^k
\end{aligned}
\right.
\label{eq_FLModelUpdate}
\end{equation}
where $\eta_t$ is the learning rate. \blu{In this paper, we consider a general scenario in which the learning rate $\eta_t$ can be a time-varying variable. It has already been proved that adopting time-varying learning rate, e.g., diminishing learning rate, is critical to ensure the convergence for most SGD-based algorithmic solutions \cite{zeiler2012adadelta}\cite{you2019does}\cite{lewkowycz2021decay}.} ${\boldsymbol g}_{k,t}$ is the stochastic gradient, calculated based on the locally received data samples:
\begin{eqnarray}\label{SGD_step}
  {\boldsymbol g}_{k,t} = \frac{1}{n_k} \sum_{j=1}^{n_k}\nabla l_k({\boldsymbol w}_{k,t},x_{k,j,t}),
\end{eqnarray}
and ${\boldsymbol v}_{k,t+1}$ is the model aggregated by the coordinator. 

In (\ref{SGD_step}), we assume $n_k$ training data samples $\{x_{k,1,t},x_{k,2,t},\cdots,x_{k,n_k,t}\}$ are computed by edge server $k$ during the $t$th training iteration. We follow the commonly adopted setting\cite{yu2019parallel} and assume data samples at any edge server have unbiased independent stochastic gradients, i.e., $\mathbb{E}_{x_{k,j,t} \sim {\cal D}_k}[{\boldsymbol g}_{k,t}|{\boldsymbol x}_{t-1}]= \nabla F_k({\boldsymbol w}_{k,t})$, where ${\boldsymbol x}_{t-1} = \langle {x}_{k,j,\tau} \rangle_{{k \in \cal{K}},j \in \{1,\cdots,n_k\},\tau \in \{0,\cdots,t-1\} }$.

\begin{figure*}
  \centering
  \includegraphics[width=16 cm]{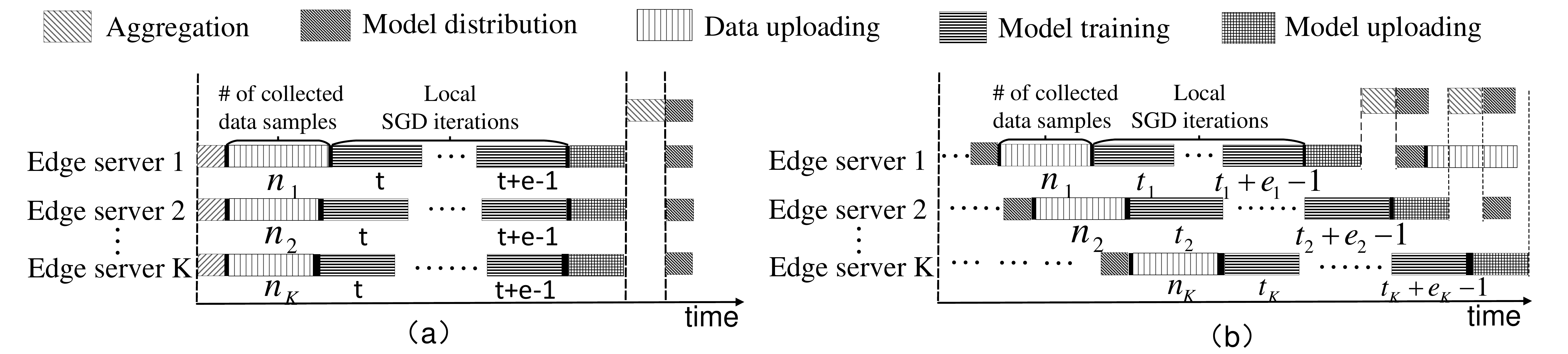}\\
  \caption{Illustration of collaborative model training in an FEI system under: (a) synchronous coordination, and (b) asynchronous coordination.}\label{Fig_SynandAsyn}
\end{figure*}

FL with synchronous and asynchronous coordination is illustrated in Fig. \ref{Fig_SynandAsyn}. With synchronous coordination, the coordinator will periodically aggregate all local models uploaded by the edge servers that are participating in model training. Let $\cal M$ be the subset of edge servers participating in model training, ${\cal M} \subseteq {\cal K}$. We can write $\mathcal{I}^k=\mathcal{I}^j=\mathcal{I}$ for $k \neq j$, $k,j\in {\cal M}$ and ${\boldsymbol v}_{k,t}$ in (\ref{eq_FLModelUpdate}) can be written as 
\begin{eqnarray}
{\boldsymbol v}_{k,t} = \sum_{k \in \cal{M}}p_k ({\boldsymbol w}_{k,t}-\eta_t {\boldsymbol g}_{k,t}), \mbox{for } t \in \mathcal{I}. 
\end{eqnarray}

In the asynchronous coordination case, the coordinator will update the global model whenever it receives a local model from any edge server. In this case, the model training and updating speeds at different edge servers are different, i.e., the $t$th iteration of model training at edge server $k$ may correspond to the $(t+j)$th iteration at edge server $l$. Without loss of generality, we focus on the model updating process of an individual edge server, i.e., edge server $k$. Let $\mathcal{W}_{k,t}^{h} \subseteq \{0,1,2,...,T-1\}$ be the set of local SGD updates of $\{{{\boldsymbol w}_{h,t}}\}_{t \geq 0}$ that have been aggregated and integrated into the global model during the past updating steps and before the $t$th interaction of edge server $k$. Note that $\mathcal{W}_{k,t}^{h}$ can be different from $\mathcal{W}_{k',t}^{h}$ for $k\neq k'$, as different edge servers usually reach iteration $t$ at different times. Since the coordinator keeps aggregating more local updates into the global model, we have $\mathcal{W}_{k,t}^{h} \subseteq \mathcal{W}_{k,t'}^{h}$ for $t' \geq t$. We use $|\mathcal{W}_{k,t}^{h}|$ to represent the cardinality of $\mathcal{W}_{k,t}^{h}$, i.e., the total number of local SGD updates uploaded from edge server $h$ that have already been aggregated into the global model at iteration $t$.
We can then write the updated global model received from the model coordination for edge server $k$ in its $t$th iteration for $t\in {\cal I}^k$ as
\begin{equation}
\begin{aligned}
{\boldsymbol v}_{k,t}={\boldsymbol w}_{0}-\sum_{h\in {\cal M}} p_h \sum_{j \in \mathcal{W}_{k,t}^{h}}{\eta_j \boldsymbol g}_{k,j}, \mbox{for } t \in \mathcal{I}^k.
\end{aligned}
\end{equation}
where ${\boldsymbol w}_{0}$ is the initial model distributed by the coordinator. 



\begin{figure}
  \centering
  \includegraphics[width=5 cm]{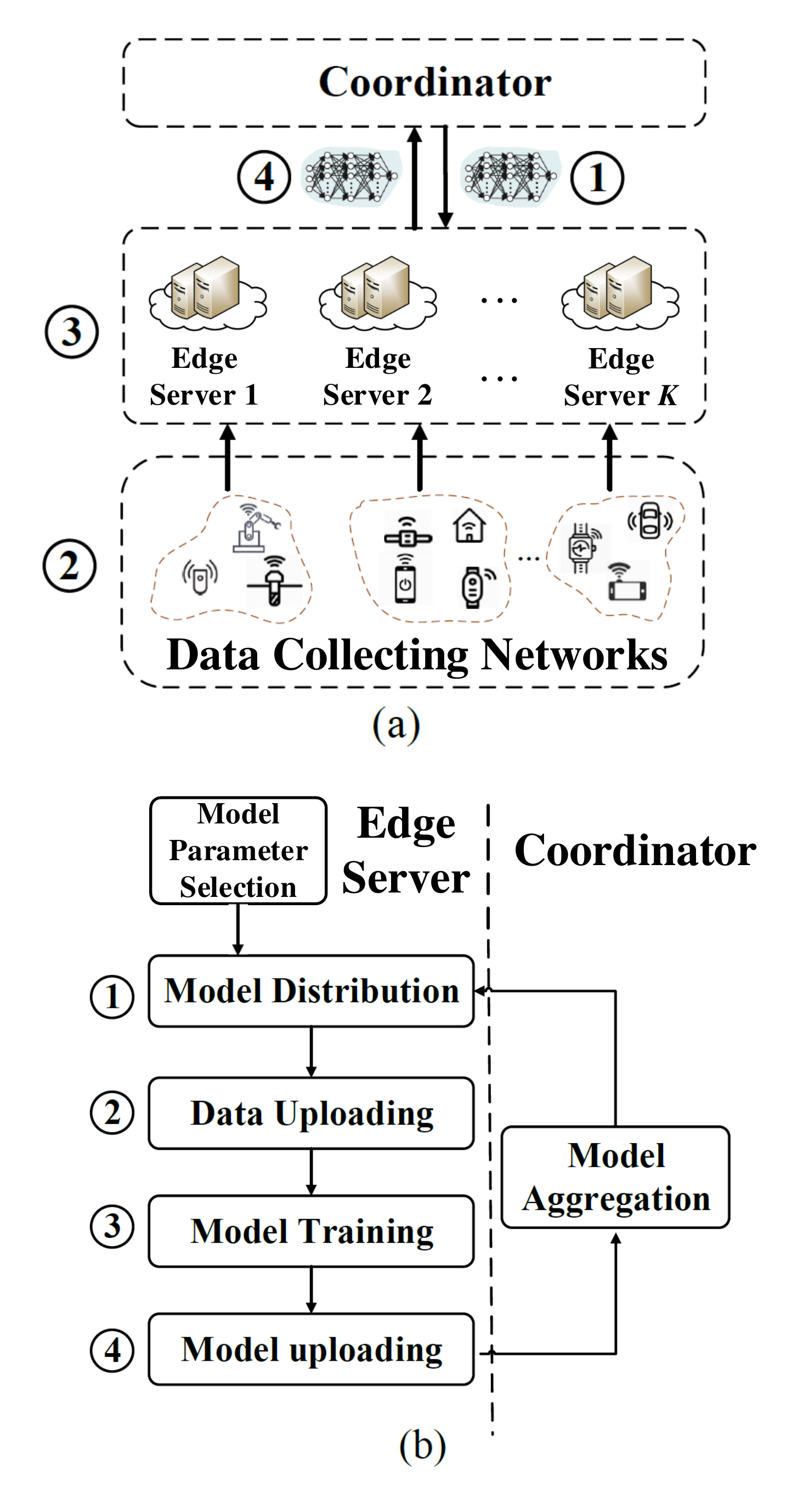}
  \vspace{-0.2in}
  \caption{\blu{An FEI architecture: (a) key system components, and (b) key training procedures.}}
  \label{Fig_FEIprocedure}
  \vspace{-0.2in}
\end{figure}

\subsection{Problem Formulation}
We investigate the time-sensitive learning of an FEI system 
in which the main design objective is to learn an ML model with a target accuracy as quickly as possible. This is different from most existing works that mainly focus on improving the convergence rate\cite{wang2020optimizing}. It has already been observed that the overall time spent on training an ML model not only depends on the number of global and local SGD iterations performed by edge servers, but also on  data collection, model computation, and updating performance.

As illustrated in Fig. \ref{Fig_FEIprocedure}, each round of collaborative training in an FEI system consists of four major steps: (1) model distribution, (2) data uploading, (3) model training, and (4) model updating and aggregation. Next, we discuss the time consumption of each step. 

\noindent
{\bf Model Distribution}: The coordinator updates the global model using (\ref{eq_FLModelUpdate}) and distributes the updated model to a subset $M$ of edge servers. Because in this step the coordinator broadcasts the same model to all selected edge servers, the time consumption does not scale with the number of edge servers and can therefore be considered as a constant, denoted by $\zeta$.

\noindent
{\bf Data Uploading}: Each edge server $k$ will update its local model based on data samples uploaded from its associated local data collecting network.
%
%
We follow the standard FL approach and assume that in each round of model coordination, each edge server first loads the required number of data samples from its associated data collecting network and then randomly samples a mini-batch of the loaded training samples at every local epoch (local model training iteration). Let $e_k$ be the number of local epochs for each round of model updating. Edge server $k$ will load $(n_k e_k)$ training samples in total during each round of model updating. 
Let $a_k$ (in seconds per sample) be the 
sample arrival rate at edge server $k$. 
The total time duration of the data uploading step can be written as 
\begin{eqnarray}\label{data_collecting_delay}
  q_k(n_k,e_k) &=& a_{k} n_k e_k.
\end{eqnarray}

\noindent
{\bf Local Model Training}: Each edge server $k$ performs $e_k$ local SGD steps between two consecutive model updating steps. In addition to the values of $e_k$ and $n_k$, the time needed to train the local model is closely related to the computational capacity of the edge server. As observed in \cite{perrone2019optimal}, the time duration of model training is closely related to the type $\varsigma$ (image, voice, text, etc.) of data samples, the time for processing each data sample by edge server $k$, denoted as $b_{k,\varsigma}$, and the time consumption for computing the gradient and updating the local model weight, denoted as $\beta_{k,\varsigma}$. We can therefore write the time consumption of local model training at edge server $k$ as $\nu_k (e_k,n_k, b_{k,\varsigma},\beta_{k,\varsigma},\varsigma)$.


\noindent
{\bf Model Uploading and Aggregation}: After $e_k$ iterations of local SGD updates, each edge server $k$ will upload its local model to the coordinator for global model updating. In the synchronous coordination scenario, the coordinator will have to wait for all the selected edge servers to upload their updated models before performing aggregation. In this case, the model uploading time mainly depends on the slowest edge server selected to participate in the given round of model coordination. Motivated by the fact that edge servers and the coordination server are usually 
connected through high-speed link (e.g., fiber) with a fixed connecting speed, we follow a commonly adopted setup and assume the time consumption for model uploading is a constant, 
denoted as $u$.


Based on the above analysis, the minimization of the model training time of an FEI system can be formulated as follows:
\begin{eqnarray}\label{Original_obj_func}
 \mbox{\bf (P)}\;\; &\min\limits_{{\boldsymbol e},{\boldsymbol n},\Gamma,{\cal M}}&\mathbb{E}[{ c({\boldsymbol e},{\boldsymbol n},\Gamma,\cal M)}]\nonumber \\
 &\mbox{s.t. }& \mathbb{E}[F({\boldsymbol w}_{{\cal M},{\Gamma}})]-F^*\leq \epsilon \label{eq_ConvConstraint} \\
  && e_k,\Gamma,n_k \in \mathbb{Z}^+ \\
  && \cal M \subseteq \cal K
\end{eqnarray}
where $c({\boldsymbol e},{\boldsymbol n},\Gamma,{\cal M}) = \sum\limits_{r=1}^{\Gamma} \max\limits_{k \in {\cal M}}[ \zeta+u+q_k(n_k,e_k)$ $+$ $\nu_k (e_k,n_k, b_{k,\varsigma},\beta_{k,\varsigma},\varsigma)]$, $\epsilon$ is the precision requirement of the ML model, ${\boldsymbol w}_{{\cal M},{\Gamma}}$ denotes the parameters of the global model trained based on edge servers $\cal M$ after $\Gamma$ rounds of global model coordination, and $F^*=\min\limits_{{\boldsymbol w} \in {\cal R}^d}F(\boldsymbol w)$. 

We can observe from problem ${\bf (P)}$ that one of the key issues that slows down the FL-based model training is the so-called {\it straggler effect}, in which faster edge servers 
will have to wait for {\it stragglers}, i.e., slow edge servers with limited computation and communication capabilities, to finish each round of the global model coordination. 

Solving problem ${\bf (P)}$ involves joint optimization of multiple entangled model training parameters, including $\boldsymbol e$, $\boldsymbol n$, and $\mathcal M$. 
The heterogeneity of the system (e.g., in terms of the computational capacities of different edge servers) and datasets (e.g. non-iid distributions of data samples at different edge servers) further exacerbate the challenge. An analytical framework for characterizing the relationship and understanding the joint impact of these parameters is needed. Joint optimization of all these relevant parameters to improve the speed of model training process in the FEI system with both system and data heterogeneity is still an open problem.

\section{TS-FL with Synchronous Coordination}\label{Sec_S_TSFL}

\subsection{Training Time Modeling and Approximation}
It is known that the model training speed of each individual edge server is dominated by the computational capacity of the edge server as well as the adopted model training parameters\cite{wang2021towards}.
We follow a commonly used setting and assume the time duration of each local model training iteration of a specific edge server is a fixed value under a given combination of model training parameters. 

During each round of model coordination in TS-FL-SC, the coordinator needs to wait for all the selected edge servers to finish uploading their model before performing model aggregation and updating. In this case, each round of model coordination will be dominated by the slowest edge server. 
Without loss of generality, we order the set ${\cal M}$ of $M$ servers from the fastest to the slowest based on their computational capacity and relabel edge servers according to their speed rankings, i.e., we abuse the notation and again use subscript $m$ to denote the $m$th fastest edge server and therefore the set of edge servers is relabeled from the fastest to the slowest as $1, 2, \ldots, M$. 
The overall time consumption during each round of model updating among set $\cal M$ of edge servers is therefore dominated by the slowest edge server $M$. 
For $\Gamma$ iterations of local model training in TS-FL-SC, the training time of the set ${\cal M}$ of edge servers then can be written as  
\begin{eqnarray}
  {\mathbb{E}[c(e,n,\Gamma,\cal M)]} &=& \Gamma \cdot c_{{M}} (e_{{M}}, n_{{M}}, b_{{{M}},\varsigma}, \beta_{{{M}},\varsigma}, \varsigma)
 \label{eq_TotalTimeConsum}
\end{eqnarray}
where $c_{M} (e_M, n_M, b_{M,\varsigma}, \beta_{M,\varsigma}, \varsigma)$ is the time duration of each iteration in the model training of the slowest edge server $M$ under parameters $e_M$ and $n_M$ and a given type of data sample $\varsigma$. $b_{k,\varsigma}$ and $\beta_{k,\varsigma}$ are closely related to the computational capacity of edge server and can therefore considered as a constant under given combination of $e_k$ and $n_k$.

From (\ref{eq_TotalTimeConsum}), 
one can envision two potential solutions to alleviate the straggler effect for TS-FL-SC:
\begin{itemize}
    \item[(1)] {\it Server Dropping}, which involves directly removing the slowest edge server as well as its corresponding data samples from participating in the model training process, and
    \item[(2)] {\it Parameter Optimization}, which focuses on optimizing a combination of parameters, especially $e_k$ and $n_k$, to reduce the time consumed for each iteration and also improve the convergence performance.
\end{itemize}

\blu{Both of the above solutions have their pros and cons. In particular, simply removing the slowest edge server from the model training can reduce the waiting time in each round of coordination. It will however make the resulting model impossible to converge to the global optimal solution with all the data samples from all the edge servers being included in the training process. The parameter optimization solution is easy for implementation. However, it is generally impossible for the coordinator or edge servers to know the optimal combination of all the parameters. 
In the rest of this section, we will develop solutions to the above issues. In particular, we will first consider the server dropping approach. Later on, we will investigate the parameter optimization approach.}

\subsection{Server Dropping}
\subsubsection{Impact of Server Dropping}
Even though server dropping can significantly reduce the overall time consumed by collaborative model training and updating process, it may bias the result and slow convergence speed due to the removal of some data samples from model training. More specifically, recent results\cite{khaled2020tighter} 
suggest that if the data samples from all edge servers follow the same distribution, the FL convergence speed increases almost linearly with the number of edge servers. However, if data samples at different servers follow different distributions (non-i.i.d.), the impact of removing some edge servers from model training on the convergence performance will be much more complex. In particular, it is possible that the data samples at the slowest edge server may exhibit unique features which, when removed from the model training process, may prevent the resulting model from converging to the global optimal solution. In this paper, we focus on such scenarios. 
In this case, we need to first quantify the impact of removing different subsets of edge servers from model training on the convergence performance and then evaluate the overall time consumption of the training process.

\blu{Suppose the optimal model parameter that minimize the global objective function with all $K$ available edge servers being selected is given by 
${\boldsymbol w}^*_{\cK}=\arg\min\limits_{\boldsymbol{w}}{\sum\limits_{k \in \cal K}p_k F_k ({\boldsymbol{w}}_{\cK})}$. To simplify notations, we can also write $F^*_k = F_k ({\boldsymbol{w}}^*_{\cK})$. We then have the following definition. }

\blu{
\begin{definition}
\label{Definition_Impact}
%
We define the {\it impact of server dropping} as follows. The impact of removing a subset ${\cal Q} \subset {\cal K}$ of edge servers and their corresponding data samples from the model training is defined as 
difference between the global optimal objective function with all $K$ edge servers and that with subset ${\cal M} = {\cal K} \backslash {\cal Q}$ of edge servers, given by
\begin{eqnarray}
D^*_{\cal M}(F) = \sum\limits_{k \in \cal M}p_k \left(F_{k}({\boldsymbol w}^*_{\cM})-F^*_{k} \right),
\end{eqnarray}
where  and $\bw^*_{\cM}$ is the optimal models that minimize the weighted sum of the loss functions of edge server set $\cM$, defined as ${\boldsymbol w}_{\cal M}^*=\arg\min\limits_{\boldsymbol w}{\sum\limits_{k \in \cal M}p_k F_k}$.
\end{definition}
}

\blu{One of the key differences of the above definition and the previously proposed concepts of server' impact or contribution\cite{} is that, in our definition, the impact of server dropping depends on the minimized loss functions with the optimal model parameters. In the rest of this section, we first quantify the gap of loss functions caused by server dropping when a given combination of multiple relevant model parameters. We then propose a sample-based pre-training solution for estimating the optimal combination of parameters under a subset of selected edge servers.}

\subsubsection{Theoretic Analysis and Preliminary Experiments}
\blu{We can derive the following theoretical upper bound on the difference of loss functions with and without server dropping.}

\begin{theorem}
\label{Theorem_TSSC_convergence_rate}
Suppose the following assumptions hold: (1) $F_1,F_2,...,F_K$ are all $L-$smooth and $\mu-$convex, i.e., $ {\frac{\mu}{2}}\|{\boldsymbol w}-{\boldsymbol w}'\|^2$ $\leq$ $F_k(\boldsymbol w)-F_k({\boldsymbol w}')$ $+$ $({\boldsymbol w}-{\boldsymbol w}')^T \nabla F_k({\boldsymbol w}')$ $\leq$ ${\frac{L}{2}}\|{\boldsymbol w}-{\boldsymbol w}'\|^2$ for all ${\boldsymbol w},{\boldsymbol w}'\in \mathbb{R}^d$; (2) data samples $x_k$ are uniformly randomly sampled from ${\cal D}_k$; and (3) the stochastic gradient satisfies $\mathbb{E}\|\nabla l_k({\boldsymbol w},x_k)\|^2 \leq G^2$ and $\mathbb{E}\|\nabla l_k({\boldsymbol w},x_k)-\nabla F_k({\boldsymbol w})\|^2 \leq {\sigma_k}^2$.
If $\kappa={\frac{L}{\mu}}$, $\gamma=\max \{8\kappa,e\}$, and the learning rate $\eta_t$ satisfies $\eta_t={2 \over {\mu(\gamma+t)}}$, we have
\begin{eqnarray}
\lefteqn{ \mathbb{E}[F({\boldsymbol w}_{{\cal M},{T}})]-F^* \leq } \label{eq_Theorem1} \\
   && {4\kappa \over \mu({\gamma+T})}({{\sum_{k \in \cal M}{p_k^2{\sigma_k}^2 \over n_k}+8e^2 G^2}}+C_{\cal M} ) +L D^*_{\cal M}(\boldsymbol{w}) \nonumber 
\end{eqnarray}
where ${\boldsymbol w}_{{\cal M},{T}}$ is the model trained by edge servers in the subset ${\cal M}$ after the $T$th local iteration, $F^* = \sum_{k\in {\cal K}}F^*_k$ is the global minimum value of  objective function when all $K$ edge servers are selected to participate in model training, and $C_{\mathcal M}=6L D^*_{\cM}(F)+{\mu^2 (\gamma+1) \over 4}\|{\boldsymbol{w}}_{0}-\boldsymbol{w}_{\mathcal{M}}^*\|^2$.
\end{theorem}
\begin{IEEEproof}
 See Appendix \ref{Appendix_ProofTheorem1}.
\end{IEEEproof}

Inequality (\ref{eq_Theorem1}) provides an upper bound on the gap between the loss function trained with only a subset $\cal M$ of edge servers and that with all $K$ edge servers.
\blu{This gap characterizes the maximum bias of the model trained by only a subset of edge servers, compared to the global optimal model trained with all the available edge servers. As mentioned earlier, under the given dataset and loss function, the value of loss can be used to characterize the prediction accuracy of the output model. In other words, the above gap of loss values can be used to capture the bias in the model prediction accuracy caused by the removal of edge servers. From inequality  (\ref{eq_Theorem1}), we can observe that the gap is closely related to a set of key parameters including the subset of selected edge severs $\cal M$, minibatch size $n_k$, the number of local epochs $e$, and the total number of local SGD iterations $T$. In the rest of this section, we will try to jointly optimize all these parameters to minimize the overall runtime of model training when the bias of the model accuracy  can be controlled within a maximum tolerable level.} 

By substituting $T=e \Gamma$, we can simplify (\ref{eq_Theorem1}) into the following form:
\begin{eqnarray}
\lefteqn{\mathbb{E}[F({\boldsymbol w}_{{\cal M},T})]-F^*} \nonumber \\
&&\;\;\;\; \leq{1 \over e \Gamma}(\sum_{k \in \cal M}{{A} \over n_k}+{B}e^2+{C}_{\cal M})+{D}_{\cal M}
\label{syn_convergence}
\end{eqnarray}
where ${A} = \frac{4\kappa\max\limits_{k \in \cal M}p_k^2\sigma_k^2}{\mu}$, ${B} = \frac{32\kappa G^2}{\mu}$, ${C}_{\cal M} = \frac{24\kappa LD^*_{\cM}(F)}{\mu} +{\mu\kappa (\gamma+1)}\|{\boldsymbol{w}}_{0}-\boldsymbol{w}_{\mathcal{M}}^*\|^2)$ and ${D}_{\cal M} = L D^*_{\cal M}(\boldsymbol{w})$.  
Generally speaking, these values cannot be known by edge servers or by the coordinator. They are, however, important for estimating the overall model training time. In Section \ref{Subsection_PreTraining}, we will discuss how to estimate these constants using a sample-based pre-training approach. 

Substituting (\ref{syn_convergence}) into (\ref{eq_ConvConstraint}), we obtain the following sufficient condition for the convergence constraint of TS-FL-SC:
\begin{eqnarray}\label{syn_sufficient}
{1 \over e \Gamma}(\sum_{k \in \cal M}{{A} \over n_k}+{B}e^2+{C}_{\cal M})+{D}_{\cal M} &\leq& \epsilon.
\end{eqnarray}

The above inequality can be utilized to calculate the possible combination of parameters that achieve a given target of model accuracy. By substituting (\ref{syn_sufficient}) into (\ref{eq_TotalTimeConsum}), we can obtain the following approximated version of  problem ({\bf P}):
\begin{eqnarray}\label{syn_obj}
\mbox{\bf (P1)}\;\; & \min\limits_{e,{\boldsymbol n},\mathcal{M}}& \mathbb{E} [\tilde{c}(e,\boldsymbol{n},{\cal M})] \nonumber\\
&\mbox{s.t.}& {D}_{\cal M}<\epsilon \mbox{ and } e,n_k \in \mathbb{Z}^+, \mbox{ for } {\mathcal M} \subseteq \cal{K} \nonumber
\end{eqnarray}
where 
\begin{eqnarray}\label{approximate_pro}
  \mathbb{E}[\tilde{c}(e,\boldsymbol{n},{\mathcal M})] &=& {{c_{{M}} (e_{{M}}, n_{{M}}, b_{{{M}},\varsigma}, \beta_{{{M}},\varsigma}, \varsigma)} \over e (\epsilon-{D}_{\cal M})} \nonumber \\
  &&\;\;\left(\sum_{k \in \cal M}{{A} \over n_k}+{B}e^2+{C}_{\cal M}\right).
\end{eqnarray}

\begin{figure}
  \centering
  \includegraphics[width=8 cm]{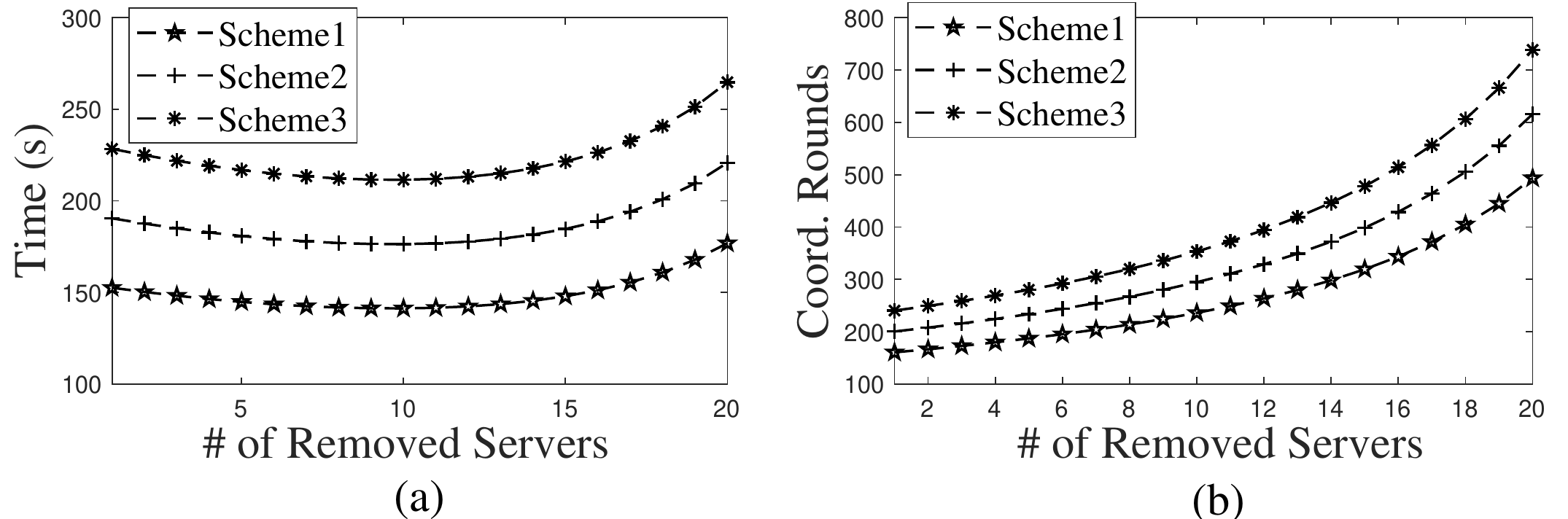}\\
  \caption{\blu{(a) Time duration and (b) number of coordination rounds required to train a model with a target model accuracy when different numbers of slow edge servers have been removed from the training process. Three model training schemes have been compared: {\it Scheme 1}  with $n_k=200$ and $e_k=80$, {\it Scheme 2} with $n_k=200$ and $e_k=100$, and {\it Scheme 3} with $n_k=200$ and $e_k=120$.} }
  \label{straggler_effect}
\end{figure}

\blu{By exploiting the above reformulated problem {\bf (P1)}, we can calculate the optimal combination of parameters that minimize the overall runtime for training a model with the guaranteed output accuracy $\epsilon$. } 

\blu{We can now evaluate the impact of the server dropping on the speed of the model training. In Fig. \ref{straggler_effect}, we present the overall runtime and number of coordination rounds required to train a model with a guaranteed accuracy when different numbers of slowest edge servers are removed. We observe different trends when the required runtime and number of coordination rounds are evaluated for model training. This once again verifies our observations that these two performance metrics are fundamentally different in FEI systems. In particular, in Fig. \ref{straggler_effect}(b), the required number of coordination  rounds decreases when more and more edge servers as well as their corresponding data samples have been removed from participating in the model training. This result is aligned with previously reported results which suggest more data samples from edge servers can always help improving the convergence speed of the model training process\cite{stich2018local}. In Fig. \ref{straggler_effect}(a), however, we observe that the overall runtime for training a model with a guaranteed accuracy can be reduced when only a few slowest edge servers have been removed. As more and more edge servers have been dropped from the model training, the required model training time will eventually start to increase.  This is because, by removing only a few slowest edge servers, the fast edge servers do not have to wait for these slow edge servers in each round of coordination. Therefore, despite the increase of the total number of required coordination rounds, the overall runtime to reach a target model accuracy can still be reduced.  However, when more edge servers as well as their corresponding data samples have been removed from the model training, the adverse effect such as the slowed convergence as well as the bias in the resulting model caused by the removal of data samples will start to dominate the overall model training process. }



\subsection{Parameter Optimization}

\subsubsection{Sample-based Pre-training for Unknown Parameter Estimation}
\label{Subsection_PreTraining}
As mentioned earlier, the objective function $\mathbb{E}[\tilde{c}(e,\boldsymbol{n},{\mathcal M})]$ contains unknown constant parameters that may affect the time consumption of model training. These unknown parameters can be estimated by using a pre-training-based solution described as follows. 

\noindent
{\bf Estimation of Parameters ${A}$, ${B}$ and $C_{\cal M}$:}
We adopt the sampling-based approach in \cite{luo2020cost} to estimate the unknown parameters that affect the training time. The basic idea is to empirically select a set of $I$ different combinations of parameters $\langle e^{(i)},n^{(i)} \rangle$ for $i \in \{1,...,I\}$ to pre-train an ML model  
and use the observed convergence results to estimate the unknown parameters including ${A}$, ${B}$ and $C_{\cal M}$ and $D_{\cal M}$ in the (\ref{syn_sufficient}). We use superscript $(i)$ to denote the $i$-th model pre-training process based on a selection of empirical parameters $(e^{(i)},{n}^{(i)})$ for $1\leq i\leq I$. In the pre-training, the coordinator first pre-select two global loss values $F_a$ and $F_b$ and then records the required numbers of coordination rounds $R_{a}^{(i)}$ and $R_{b}^{(i)}$ to reach the target losses $F_a$ and $F_b$ from the initial global model ${\boldsymbol w}_0$ when employing different combinations of empirically selected parameters. To reduce the time duration of the pre-training process, $F_a$ and $F_b$ can be set to relatively high loss values.

More formally, by applying the convergence bound in (\ref{syn_convergence}) to approximate the global loss value after $R_a^{(i)}$ and $R_b^{(i)}$ coordination rounds, we have the following results
\begin{eqnarray}
  F_a - F^* &\approx& {1 \over e_i R_a^{(i)}}({{A} \over n^{(i)}}+{B}{e^{(i)}}^2+{C}_{\cal M})+{D}_{\cal M},\label{a1} \\
  F_b - F^* &\approx& {1 \over e_i R_b^{(i)}}({{A} \over n^{(i)}}+{B}{e^{(i)}}^2+{C}_{\cal M})+{D}_{\cal M}\label{a2}.
\end{eqnarray}

Combining (\ref{a1}) and (\ref{a2}), we have
\begin{eqnarray}
  e^{(i)}(R_{b}^{(i)} - R_{a}^{(i)}) &\approx&({1 \over F_b - F^*-{D}_{\cal M}}-{1 \over F_a - F^*-{D}_{\cal M}}) \nonumber\\
  &&({{A} \over n^{(i)}}+{B}{e^{(i)}}^2+{C}_{\cal M}).
\end{eqnarray}
By sampling different pairs of $(i,j)$ for $i,j \in \cal I$ and $i \neq j$, we can have
\begin{eqnarray}
{e_i(R_{b}^{(i)} - R_{a}^{(i)}) \over e_j(R_{b}^{(j)} - R_{a}^{(j)})} &\approx& {{{A} \over n^{(i)}}+{B}{e^{(i)}}^2+{C}_{\cal M}\over {{A} \over n^{(j)}}+{B}(1+{1 \over s^{(j)}}){e^{(i)}}^2+{C}_{\cal M}}. \nonumber\\
\label{a3}
\end{eqnarray}

Reorganizing (\ref{a3}), we can write
\begin{eqnarray}
{{ A}{ ({1 \over n^{(i)}}-\chi_{i,j}{1 \over n^{(j)}})}+{ B}{e^{(i)}}^2-\chi_{i,j}(1+{1 \over s^{(j)}}){e^{(i)}}^2} \nonumber \\
\approx C_{\cal M}(\chi_{i,j}-1).
\label{upper_estimate}
\end{eqnarray}
where $\chi_{i,j}\triangleq{e^{(i)}(R_{b}^{(i)}-R_{a}^{(i)}) \over e^{(j)}(R_{b}^{(j)}-R_{a}^{(j)})}$. Repeating the above steps for all the different combinations of parameters, we can eventually obtain estimated solutions for the unknown constants ${A}$, ${B}$ and $C_{\cal M}$ using standard linear fitting method, e.g., least square method.

\noindent
{\bf Estimation of Parameter ${D}_{\cal M}$:}
Based on the estimated parameters ${A}$, ${B}$ and $C_{\cal M}$, we can calculate ${D}_{\cal M}$ as follows:
\begin{eqnarray}
{D}_{\cal M} \approx F_b - F^* - {1 \over e_i R_b^{(i)}}({{A} \over n^{(i)}}+{B}e^{(i)2}+{C}_{\cal M}).
\end{eqnarray}

\noindent
{\bf Estimation of Parameters $\zeta$, $a_{k,\varsigma}$, $b_{k,\varsigma}$, $\beta_{k,\varsigma}$, and $u$:} 
To define the local model training delay $\left(\nu_k (e,n_k, b_{k,\varsigma},\beta_{k,\varsigma}, \varsigma)+u\right)$, we adopt a commonly adopted setting \cite{perrone2019optimal} and assume the delay under fixed hardware and software environments, can be represented into the following form:
\begin{eqnarray}\label{comp_time}
\nu_k (e_k,n_k, b_{k,\varsigma},\beta_{k,\varsigma}, \varsigma)=e_k(b_{k,\varsigma}n_k+\beta_{k,\varsigma}).
\end{eqnarray}

Suppose the coordinator can record the time consumption of each coordination round for each selected edge server. We can therefore obtain the average time consumption ${\bar c}_{k}(e^{(i)}, n^{(i)})$ under a given pair of parameters $(e^{(i)},n^{(i)})$ for $i \in \{1,2,...,I\}$. 
We can then have
\begin{eqnarray}
  \zeta+\alpha_k e^{(i)} n^{(i)} + \beta_k e^{(i)}+u &\approx& {\bar c}_{k}(e^{(i)}, n^{(i)}), \nonumber \\
  &&\; \forall i \in \{1,2,...,I\}.
  \label{latency_estimate}
\end{eqnarray}

By jointly solving all $I$ equations  in (\ref{latency_estimate}), we can obtain estimated values of  unknown parameter $\zeta$, $\alpha_k=a_{k,\varsigma}+b_{k,\varsigma}$, $\beta_k=\beta_{k,\varsigma}$, and $u$. The detailed procedures of the pre-training for parameter estimation is presented in Algorithm 1.


\begin{algorithm}[!t]
  \caption{Sample-based Pre-training Algorithm}\label{main_Algorithm}
  \begin{algorithmic}[1]
    \STATE {\bf Input:} Target loss values: $F_a, F_b$, and $I$ empirically selected combinations of training parameters, the subset $\mathcal M$ of edge servers;
    \STATE {\bf Output:} ${A}$, ${B}$, ${C}_{\cal M}$, ${D}_{\cal M}$, $\zeta, \alpha_k, \beta_k,$ and $u$;

    \FOR {$i=\{1,...,I\}$}
    \STATE Empirically choose $(e^{(i)},n^{(i)})$ for each edge server $k \in \cal M$;
    \STATE Record values $R_{a}^{(i)}$ and $R_{b}^{(i)}$ when target loss $F_a$ and $F_b$ are reached;
    \STATE Record the average time duration ${\bar c}_{k}(e^{(i)},n^{(i)})$ of edge server $k$ to complete each round of model coordination;
    \ENDFOR
    \STATE Estimate the unknown parameters ${A}$, ${B}$, ${C}_{\cal M}$, and ${D}_{\cal M}$ 
    using (\ref{upper_estimate});
    \STATE Estimate the unknown parameters $\zeta, \alpha_k, \beta_k,$ and $u$ using (\ref{latency_estimate}); 
    \STATE {\bf Return} ${A}$, ${B}$, ${C}_{\cal M}$, ${D}_{\cal M}$, $\zeta, \alpha_k, \beta_k,$ and $u$;
  \end{algorithmic}
\end{algorithm}

\begin{algorithm}[!]
  \caption{Alternate Convex Search Algorithm}\label{SYN_solution}
  \begin{algorithmic}[1]
  \STATE \textbf{Input:} Parameters: ${A}$, ${B}$, ${C}_{\cal M}$, ${D}_{\cal M}$, $\zeta$, $\alpha_k$, $\beta_k$, $u$;
  \STATE \textbf{Initialization:} Initial model parameters $z_{0}=(e_0,{n}_0)$; Stopping criterion $\theta$; Domains ${\cal Z}_{n}$ and ${\cal Z}_{e}$; $i \leftarrow 0$;
  \STATE \textbf{Output:} Solution $(e^*,{n}^*,{\cal M})$;
  \WHILE {$\mathbb{E}[\tilde{c}(z_{i-1},{\cal M})]-\mathbb{E}[\tilde{c}(z_i,{\cal M})] \geq \theta$}

    \STATE Find $e^*$of $\tilde{c}(e,{n}_i,{\cal M})$ in the search domain ${\cal Z}_{e}$ when give $s_i$ and ${n}_i$ as well as ${\cal M}$ and perform $e_{i+1}\leftarrow e^*$;
    \STATE Find $n^*$ of $\tilde{c}(e_{i+1},n,{\cal M})$ in the search domain ${\cal Z}_{n}$ when given $\cal M$ and perform $n^{i+1}\leftarrow n^*$;
    \STATE $z_{i+1}\leftarrow (e_{i+1},{n}_{i+1})$, $i\leftarrow i+1$;
    \ENDWHILE
    \STATE {\bf Return} the solution $(e_{i},{n}_{i},{\cal M})$.
  \end{algorithmic}
\end{algorithm}
\noindent

\subsubsection{Parameter Optimization Algorithm}

As mentioned earlier, the  model training time duration is closely related to the key training parameters, especially $n_k$ and $e_k$. Unfortunately, these two parameters are entangled with each other in affecting the overall model training time. For example, reducing the local epoch number $e_k$ at each edge server $k$ may accelerate the convergence of global model. It will however increase the frequency of model coordination, resulting in high communication overhead and coordination delay. Similarly, loading more training data samples $e_k n_k$ and choosing a large mini-batch size $n_k$ at each training iteration at an edge server $k$ will reduce the total number of required iterations for the model to converge to a satisfactory result.
It will however slow down the data loading time and also extend computational time at each iteration of the local model training process. 
Therefore, it is generally difficult to jointly optimize both $n_k$ and $e_k$ to minimize the overall time duration for training a satisfactory model.
Fortunately, we can prove that the objective function of problem {\bf (P1)} is a biconvex optimization problem when $n_k=n$ for every $k \in \mathcal{M}$ under a given subset $\mathcal M$ of edge servers. In particular, we can prove the following result.



\begin{lemma}
\label{Lemma_SC_biconvex}
Suppose $n_k=n$ for every $k \in \mathcal{K}$ and changing $n$ does not affects the rank of model training times among edge servers, problem {\bf (P1)} is biconvex over $e$ and $n$.
\end{lemma}
\begin{IEEEproof}
See Appendix \ref{proofLemma1}.
\end{IEEEproof}


Motivated by the above observation, we can adopt a commonly adopted iterative optimization method called alternate convex search (ACS) to calculate the optimal solution of $e^*$ and ${\boldsymbol n}^*$ to minimize the objective function $\mathbb{E}[\tilde{c}(e,\boldsymbol{n},{\cal M})]$. We present the detailed procedures of the ACS approach in Algorithm 2.  


\section{TS-FL with Asynchronous Coordination}\label{sec_A_TSFL}

As mentioned earlier, TS-FL-SC with server dropping may result in a biased model at the end of the model training process and therefore may not suitable for the scenarios when the data samples at the slowest edge server possess unique characteristics that are important for the model. In this section, we investigate model training acceleration solutions for TS-FL-ASC in which each edge server can upload its local model to the coordinator right after finishing the required number of local training iterations and can receive an instantaneous update of the global model from the coordinator.  


\subsection{Modeling the Staleness Effect}
%
Since TS-FL-ASC allows each edge server to obtain an instantaneous update of the global model at any time during the training process, it will have the potential to avoid waiting for the slow edge servers.
Unfortunately, it is known that the convergence performance of the asynchronous coordination is adversely affected by the so-called {\it staleness effect}, that is the gap between the model updating iterations at the slowest edge server and that at the fastest edge server increases with the total number of iterations, resulting in slow convergence or even divergence of the global model in each round of model aggregation. 

To quantify the impact of the staleness effect on the convergence of model training, we introduce the following virtual global model updating sequence $\{\bar{\boldsymbol w}_t\}_{t=1, \ldots, T}$ as an equivalent form of a centralized mini-batch SGD process updated by a set of decentralized edge servers following the asynchronous coordination as TS-FL-ASC, i.e., we define the virtual sequence of TS-FL-ASC as follows:
\begin{eqnarray}
\bar{\boldsymbol w}_0&\triangleq&{\boldsymbol w}_0, \\
{\bar{\boldsymbol w}}_t &\triangleq& {\boldsymbol w}_0 - \sum_{k=1}^{K} p_k \sum_{j=0}^{t-1}\eta_j{\boldsymbol g}_{k,j}.
\end{eqnarray}

We can observe that $\bar{\boldsymbol w}_t$ satisfies $\bar{\boldsymbol w}_t= \frac{1}{K} \sum^{K}_{k=1} p_k {\boldsymbol w}_{k,t}$.
Note that the virtual sequence is only introduced for our analysis and it does not need to be calculated explicitly.

We also define an upper bound $\tau$ of staleness as the maximum gap between the numbers of iterations locally performed by different edge servers, i.e., we introduce the following constraint for TS-FL-ASC:
\begin{eqnarray}
\max \limits_{h \in \mathcal{K}}{\Lambda(\mathcal{W}_{k,t}^{h})}-\min\limits_{h \in \mathcal{K}}{\Lambda(\mathcal{W}_{k,t}^{h})}\leq \tau
\label{eq_Staleness}
\end{eqnarray}
where $\Lambda(\mathcal{W}_{k,t}^{h})$ is the index of local iterations reported by edge server $h$ to the coordinator when edge server $k$ uploads its local model in the $t$th iteration.


We can then prove the following result.
\begin{lemma}\label{lemma_proof}
Suppose $x_k$ is data samples following distribution ${\cal D}_k$. We assume the following conditions hold: (1) the stochastic gradient satisfies $\mathbb{E}\|\nabla l_k({\boldsymbol w},x_k)\|^2 \leq G^2$; (2) the staleness is upper bounded by (\ref{eq_Staleness}); (3) learning rate $\eta_t$ is non-increasing and satisfies $\eta_t \leq 2 \eta_{H+\tau+t}$ for all $t>0$ where $H\triangleq\max\limits_{k \in \cal K}e_k$. The gap between the local model trained at edge server $k$ and the (virtual) global  model is upper bounded by
\begin{eqnarray}\label{lemma2_con1}
\mathbb{E}(\|\bar{\boldsymbol w}_t-{\boldsymbol w}_{k,t}\|^2) \leq 24 \eta_{t}^2 G^2 (2\tau^2+e_k^2).
\end{eqnarray}

Similarly, we can write the upper bound of the gap between weighted sum of all the local models at edge servers and the global model as follows:
\begin{eqnarray}\label{lemma2_con2}
\mathbb{E}({\sum_{k=1}^{K}p_k \|\bar{\boldsymbol w}_t-{\boldsymbol w}_{k,t}\|^2}) \leq 24 \eta_{t}^2 G^2 (2\tau^2+\sum_{k=1}^{K}p_k e_k^2)
\end{eqnarray}
where  which represents the weighted average of models of all edge servers after the $t$th SGD step.
\end{lemma}
\begin{IEEEproof}
See Appendix \ref{ProofLemma2}.
\end{IEEEproof}

We can observe from the above result that the upper bound of the staleness effect $\tau$ directly affects the convergence of the model training process.

Based on the above lemma, we can prove the following result about the convergence of TS-FL-ASC.

\begin{theorem}
\label{Theorem_ConvergenceStaleness}
Suppose the following assumptions hold: (1) $F_1,F_2,...,F_K$ are all $L-$smooth and $\mu-$convex, i.e., $ {\mu \over 2}\|{\boldsymbol w}-{\boldsymbol w}'\|^2$ $\leq$ $F_k(\boldsymbol w)-F_k({\boldsymbol w}')$ $+$ $({\boldsymbol w}-{\boldsymbol w}')^T \nabla F_k({\boldsymbol w}')$ $\leq$ ${L \over 2}\|{\boldsymbol w}-{\boldsymbol w}'\|^2$ for all ${\boldsymbol w},{\boldsymbol w}'\in \mathbb{R}^d$; (2) Data sample $x_k$ is uniformly randomly sampled from ${\cal D}_k$; and (3) the stochastic gradient satisfies $\mathbb{E}\|\nabla l_k({\boldsymbol w},x_k)\|^2 \leq G^2$ and $\mathbb{E}\|\nabla l_k({\boldsymbol w},x_k)-\nabla F_k({\boldsymbol w})\|^2 \leq \sigma_k^2$. 
By setting $\kappa={L \over \mu}$, $\gamma=\max \{8\kappa,\tau+H\}$ and $\eta_t={2 \over {\mu(\gamma+t)}}$, we have
\begin{equation}\label{stale_convergence_bound}
   \mathbb{E}[F({\boldsymbol w}_T)-F^*]\leq{\kappa \over \gamma+T}({2B \over \mu}+{\mu\gamma \over 2} \mathbb{E}\|{\boldsymbol w}_0-{\boldsymbol w}^*\|^2)
\end{equation}
where $B=\sum_{k=1}^{K}{{p_k^2\sigma_k^2 \over n_k}} +6L(F^*-\sum_{k=1}^K p_k F_k^*)+48G^2(2\tau^2+\sum_{k=1}^{K}p_k e_k^2)$, ${\boldsymbol w}^*$ is the global optimal model, ${\boldsymbol w}_T$ is weighted sum of all the local models trained by edge servers after $T$ iterations for $T \in {\cal I}^k$ and $k\in\cal K$.
\end{theorem}
\begin{IEEEproof}
See Appendix \ref{ProofTheorem2}.
\end{IEEEproof}

From the above result, we  can also observe that, when $\tau \ge \sqrt{T}$, the upper bound of convergence result in (\ref{stale_convergence_bound}) may not converge 
when the number of iterations $T$ becomes large. This once again verifies our previous result that increasing the upper bound of the staleness $\tau$ may result in degradation of convergence speed of the model training process. 

One potential solution to alleviate the staleness effect is to decrease the model training and updating speed of the fast edge servers to reduce the upper bound of the staleness within a certain limit. This however will result in more time consumption in the model training process and therefore may not suitable for some time-sensitive applications.



\subsection{Load Forwarding}
In this subsection, we propose a load forwarding-based solution to reduce the staleness by allowing some slow edge servers to offload part of its data samples to some trusted edge servers with high processing capability. To simplify our description and also due to the fact that, in many practical systems, allowing a slow edge server to reveal its data samples to multiple fast edge severs may raise security concern, in the rest of this section, we mainly focus on the load forwarding solution within a trusted edge server pair consisting of a relatively fast edge server and a slower edge server. \blu{This is reasonable for many practical network systems in which, to further improve the robustness of its networking systems, a single network operator or service provider may sign contract with at least one other service provider for resource sharing when necessary. For example, major telecommunication operators in US and Europe have already signed contract with at least one other operator for various forms of resource sharing including infrastructure and spectrum sharing \cite{Bourreau2021InfraSharingOperator}.} We write an edge server pair as $\langle k, k' \rangle$ where, without loss of generality, we label the slow edge server as $k$ and fast edge server as $k'$.
We abuse the notation and use $k$ to denote edge server pair $\langle k, k' \rangle$. Let $\cal P$ be the set of edge server pairs that support load forwarding, i.e., $k \in {\cal P}$.

Note that in each edge server pair, during each round of model coordination, the slow edge server only needs to send a batch of its local data samples to the fast edge server.
The fast edge server $k'$ can then directly obtain the updated model from the coordinator, perform model training based on the received batch of data samples, 
and finally upload the updated model to the coordinator. 
In other words, the fast edge server will take over the local training process for the slow edge servers without requiring any model transfer within the edge server pair. We define the data communication delay caused by  forwarding a batch of data samples from edge servers $k$ to $k'$ as $o_{k,k'} = \lambda_k n_k e_k$ where $\lambda_{k}$ is the data forwarding speed for sending each data sample from edge servers $k'$ to $k$.

We write $\alpha_{k,k'}$ as the portion of model updating rounds of the slow edge server $k$ that are helped by the fast edge server $k'$. We can then write the overall time consumption of model training with load forwarding as
\begin{eqnarray}
\lefteqn{c({\boldsymbol e},{\boldsymbol n},{\boldsymbol \Gamma},{\cal P},{\boldsymbol \alpha})} \\
&=& \max\limits_{k \in {\cal P}}\{{\Gamma_k}(1-\alpha_{k,k'})c_k, ({\Gamma_{k'}}c_{k'}+{\Gamma_{k}}\alpha_{{k},{k'}}c_{k,k'})\} \nonumber
\end{eqnarray}
where ${\boldsymbol e} = \langle e_k\rangle_{k \in {\cal K}}$, ${\boldsymbol n} = \langle n_k\rangle_{k \in {\cal K}}$, ${\boldsymbol \alpha} = \langle \alpha_{k, k'}\rangle_{k \in {\cal K}}$, ${\boldsymbol \Gamma} = \langle \Gamma_k\rangle_{k \in {\cal K}}$, $c_k=\zeta+q_k(n_k,e_k)+\nu_k(e_k, n_k)+u$ is the time required to complete each round of model updating, and $c_{{k},k'}$ $=$ $(\zeta+q_{{k}}(n_{k},e_{k})$ $+$ $o_{{k},k'}(n_{k},e_{k})+\nu_{k'}(e_{k}, n_{k} )+u)$ is overall time duration for edge server $k'$ to help edge server $k$ in training its local model.

We can observe that in an ideal condition, after applying the load forwarding, the model training time of fast and slow edge servers in each edge server pair becomes equal, i.e., ${\Gamma_k}(1-\alpha_{k,k'})c_k
={\Gamma_{k'}}c_{k'}+{\Gamma_{k}}\alpha_{k,{k'}}c_{k,k'}$. In this case, we can write the optimal  value $\alpha^*_{k,k'}$ as
\begin{eqnarray}
\alpha^*_{k,k'} = \begin{aligned}
&\dfrac{{\Gamma_k}c_k-{\Gamma_{k'}}c_{k'}}{{\Gamma_k}(c_k+c_{k,k'})},\;\;\; \forall k\in {\cal P}. 
\end{aligned}
\end{eqnarray}

We can therefore rewrite the original optimization problem {\bf (P)} as follows:
\begin{eqnarray}
 \mbox{\bf (P2)}\;\; &\min\limits_{{\boldsymbol e},{\boldsymbol n},\Gamma,{\cal P},{\boldsymbol \alpha}}&\mathbb{E}[{ c({\boldsymbol e},{\boldsymbol n},{\boldsymbol \Gamma},\cal P,{\boldsymbol \alpha})}] \nonumber \\
 &\mbox{s.t. }& \mathbb{E}[F({\boldsymbol w}_{{\cal P},{\boldsymbol \Gamma}})]-F^*\leq \epsilon,\label{eq_AsynConstraint} \\
  && e_k,\Gamma_k,n_k \in \mathbb{Z}^+, \\
  && 0 \leq \alpha_{k,k'} \leq 1.
\end{eqnarray}
where ${\boldsymbol w}_{{\cal P},{\boldsymbol \Gamma}}$ is the model trained by a set ${\cal P}$ of edge server pairs in iteration ${\boldsymbol \Gamma}$. 


In the rest of this section, we discuss how to solve problem {\bf (P2)} using server dropping and parameter optimization.

\subsection{Server Dropping}
Similar to TS-FL-SC, it is also possible to further reduce the model training time in TS-FL-ASC by removing some slow edge servers from participating in the model training process.
%

We can prove the following upper bound of the convergence rate of TS-FL-ASC when a subset ${\cal M}$ of edge servers are selected to participate in the model training.


\begin{theorem}\label{Theorem_TSASC_convergence_rate}
Suppose the following assumptions hold: (1) $F_1,F_2,...,F_K$ are all $L-$smooth and $\mu-$convex, i.e., $ {\mu \over 2}\|{\boldsymbol w}-{\boldsymbol w}'\|^2$ $\leq$ $F_k(\boldsymbol w)$ $-$ $F_k({\boldsymbol w}')$ $+$ $({\boldsymbol w}-{\boldsymbol w}')^T \nabla F_k({\boldsymbol w}')$ $\leq$ ${L \over 2}\|{\boldsymbol w}-{\boldsymbol w}'\|^2$ for all ${\boldsymbol w},{\boldsymbol w}'\in \mathbb{R}^d$; (2) data samples $x_k$ are uniformly and randomly sampled from ${\cal D}_k$; (3) the stochastic gradient satisfies $\mathbb{E}\|\nabla l_k({\boldsymbol w},x_k)-\nabla F_k({\boldsymbol w})\|^2$ $\leq$ ${\sigma_k}^2$ and $\mathbb{E}\|\nabla l_k({\boldsymbol w},x_k)\|^2 \leq G^2$; and (4) The staleness is bounded by (\ref{eq_Staleness}). If $\kappa={L \over \mu}$, $\gamma=\max \{8\kappa,\tau+H\}$, and $\eta_t={2 \over {\mu(\gamma+t)}}$, then we have
\begin{eqnarray}
\label{eq_ConvergenceSevDropTSASC}
\lefteqn{\mathbb{E}[F({\boldsymbol w}_{{\cal M}, T})]-F^* \leq L D_{\cal M}^{*}(\boldsymbol w)}\\
   &+&{4\kappa \over \mu(\gamma+T)}({{\sum\limits_{k \in \cal M}}{p_k^2{\sigma_k^2} \over n_k}+48{\sum\limits_{k \in \cal M}}{p_k{e_k}^2}G^2}+ C_{\cal M})\nonumber
\end{eqnarray}
where $C_{\cal M}=96G^2{\tau}^2+6L D_{\cal M}^{*}(F)+{\mu^2 \gamma\over4}||{\boldsymbol{w}}_{0}-\boldsymbol{w}_{\mathcal M}^*||^2$ and $D_{\cal M}^{*}(\boldsymbol w)$ is the difference of objective function between models  collaboratively trained by edge server sets $\cal K$ and $\cal M$.
\end{theorem}
\begin{IEEEproof}
  See Appendix \ref{ProofTheorem}.
\end{IEEEproof}

Theorem \ref{Theorem_TSASC_convergence_rate} characterizes the impact of both server dropping and staleness bound $\tau$ on the convergence performance of the FEI system. We can also observe that the convergence upper bound in TS-FL-ASC is relatively more loose, compared to that of TS-FL-SC presented in Theorem \ref{Theorem_TSSC_convergence_rate}, due to the staleness effect which is reflected by $\tau$ in the first term of $C_{\cal M}$.  

We can then follow the same line as TS-FL-SC to model the overall time duration of training a satisfactory model using TS-FL-ASC as follows.
%
%
We first rewrite (\ref{eq_ConvergenceSevDropTSASC}) in \ref{Theorem_TSASC_convergence_rate} into the following simplified form:
\begin{eqnarray}\label{TSFL_ASC_form}
\lefteqn{\mathbb{E}[F({\boldsymbol w}_{{\cal M}, T})]-F^*} \\
&\leq& {1 \over T}(\sum_{k \in \cal M}{{A'} \over n_k}+{B'}\sum_{k\in \mathcal {M}}p_k e_k^2+{C'}_{\cal M})+{D'}_{\cal M} \nonumber 
\end{eqnarray}
where ${A'}=\frac{4\kappa\max\limits_{k \in \cal K}p_k^2 \sigma_k^2}{\mu}$, ${B'}=\frac{192\kappa G^2}{\mu}$, ${C'}_{\cal M}=\frac{384\kappa G^2{\tau}^2}{\mu}+\frac{24\kappa LD^*_{\cM}(F)}{\mu}+{\mu\kappa \gamma}||{\boldsymbol{w}}_{0}-\boldsymbol{w}_{\mathcal M}^*||^2$ and ${D'}_{\cal M}=L D^*_{\cal M}(\boldsymbol{w})$. 

By substituting (\ref{TSFL_ASC_form}) into constraint (\ref{eq_AsynConstraint}), we can obtain the following 
constraint of TS-FL-ASC:
\begin{eqnarray}
{1 \over T}(\sum_{k \in \cal M}{{A'} \over n_k}+{B'}\sum_{k\in \mathcal {M}}p_k e_k^2+{C'}_{\cal M})+{D'}_{\cal M} &\leq& \epsilon.
\label{eq_TSFLASC_Approx}
\end{eqnarray}

By replacing the constraint (\ref{eq_AsynConstraint}) in Problem {\bf ({P3})} with (\ref{eq_TSFLASC_Approx}), we can obtain an  approximated version of problem {\bf ({P3})} as follows:
\begin{eqnarray}
\mbox{\bf (P4)} \;\; &\min\limits_{\boldsymbol{e},{\boldsymbol n}}&\mathbb{E} [\tilde{c}(\boldsymbol{e},\boldsymbol{n},{\cal M},{\boldsymbol{\alpha}})]\\
&\mbox{s.t.}& {D'}_{\cal M}<\epsilon,\mbox{ and }0 \leq \alpha_{k,k'} \leq 1 \mbox{ and } e_k,n_k \in \mathbb{Z}^+,\nonumber
\end{eqnarray}
where $\mathbb{E}[\tilde{c}(\boldsymbol{e},\boldsymbol{n},{\mathcal M},{\boldsymbol{\alpha}})]$ is given by
\begin{eqnarray}\label{approximate_ASC}
  \lefteqn{\mathbb{E}[\tilde{c}(\boldsymbol{e},\boldsymbol{n},{\mathcal M},{\boldsymbol{\alpha}})]= { {{A'} \over n_k}+{B'}\sum_{k\in \mathcal {M}}p_k e_k^2+{C'}_{\cal M} \over \epsilon-{D'}_{\cal M}}}\\
  &&({\frac{(1-\alpha_{{M},{M}'})c_{{M}}(e_{{M}},n_{{M}})}{e_{{M}}}}
  +\frac{\alpha_{{M}',{M}} c_{{M}',{M}}(e_{{M}'},n_{{M}'})}{e_{{M}'}})\nonumber
\end{eqnarray}
and $M$ is the slowest edge server pair to finish each round of model updating for $1\le M\le |{\cal P}|$.

\subsection{Parameter Optimization} 

\subsubsection{Unknown Parameter Estimation via Pre-training}
To solve problem {\bf (P4)}, we need to first estimate the impact of various parameters on the convergence of model training. We follow the same line as TS-FL-SC to estimate these parameters by pre-running model training process with different combinations of parameters.

\noindent
{\bf Estimation of Parameters ${A'}$, ${B'}$, ${C'}_{\cal M}$, ${D'}_{\cal M}$:} We empirically select $I$ different combinations of parameters $\langle e^{(i)},n^{(i)} \rangle$ for $i \in \{1,...,I\}$ to pre-train the ML model and use the observed convergence results to estimate these parameters. Under any given training parameters $\langle e^{(i)},n^{(i)} \rangle$, the convergence upper bound of TS-FL-ASC can be rewrite as:
\begin{eqnarray}
\mathbb{E}[F({\boldsymbol w}_T)]-F^* \leq {1 \over e\Gamma}\left({{A'} \over n^{(i)}}+{B'}e^{(i)2}+{C'}_{\cal M}\right)+{D'}_{\cal M}
\end{eqnarray}
where $\Gamma_k=\Gamma=\dfrac{T}{e^{(i)}}$ for all $k \in \mathcal{M}$. We can observe that the above upper bound is in the same form as (\ref{syn_convergence}). We can therefore use the same parameter estimation procedures described in Section \ref{Subsection_PreTraining} to estimate ${A'}$, ${B'}$, ${C'}_{\cal M}$, ${D'}_{\cal M}$. We omit the details due to the space limit.

\noindent
{\bf Estimation of Parameters $\zeta$, $a_{k,\varsigma}$, $b_{k,\varsigma}$, $u$:} These parameters are closely related to the computational and communication capacity of edge servers and can be obtained by adopting the same procedures discussed in Section \ref{Subsection_PreTraining} and Algorithm 1. We again omit the details due to the limit of space.

\subsubsection{Parameter Optimization Algorithm}
Similar to the TS-FL-SC, we can again 
prove that the objective function of problem (\ref{approximate_ASC}) is a biconvex optimization problem when $e_k=e$ and $n_k$ for every server $k \in \cal M$ and the subset $\mathcal M$ edge servers participating in the model training is fixed. The following lemma can be proved by following the same line as Lemma \ref{Lemma_SC_biconvex}. We omit the proof due to the limit of space.

\begin{lemma}
Suppose $n_k=n$ and $e_k=e$ for every $k \in \mathcal{K}$ and changing $n$ does not affects the rank of model training times of edge servers, problem (\ref{approximate_ASC}) is biconvex over $e$ and $n$.
\end{lemma}

We can again adopt an ACS-based approach to calculate the optimal solution of $e^*$ and $n^*$ to minimize the overall model training time $\mathbb{E}[\tilde{c}(\boldsymbol{e},\boldsymbol{n},{\mathcal M},{\boldsymbol{\alpha}})]$ for TS-FL-ASC.

\section{Numerical Result}\label{sec_simultaion}
\begin{figure}[ht]
  \centering
  \includegraphics[width=8 cm]{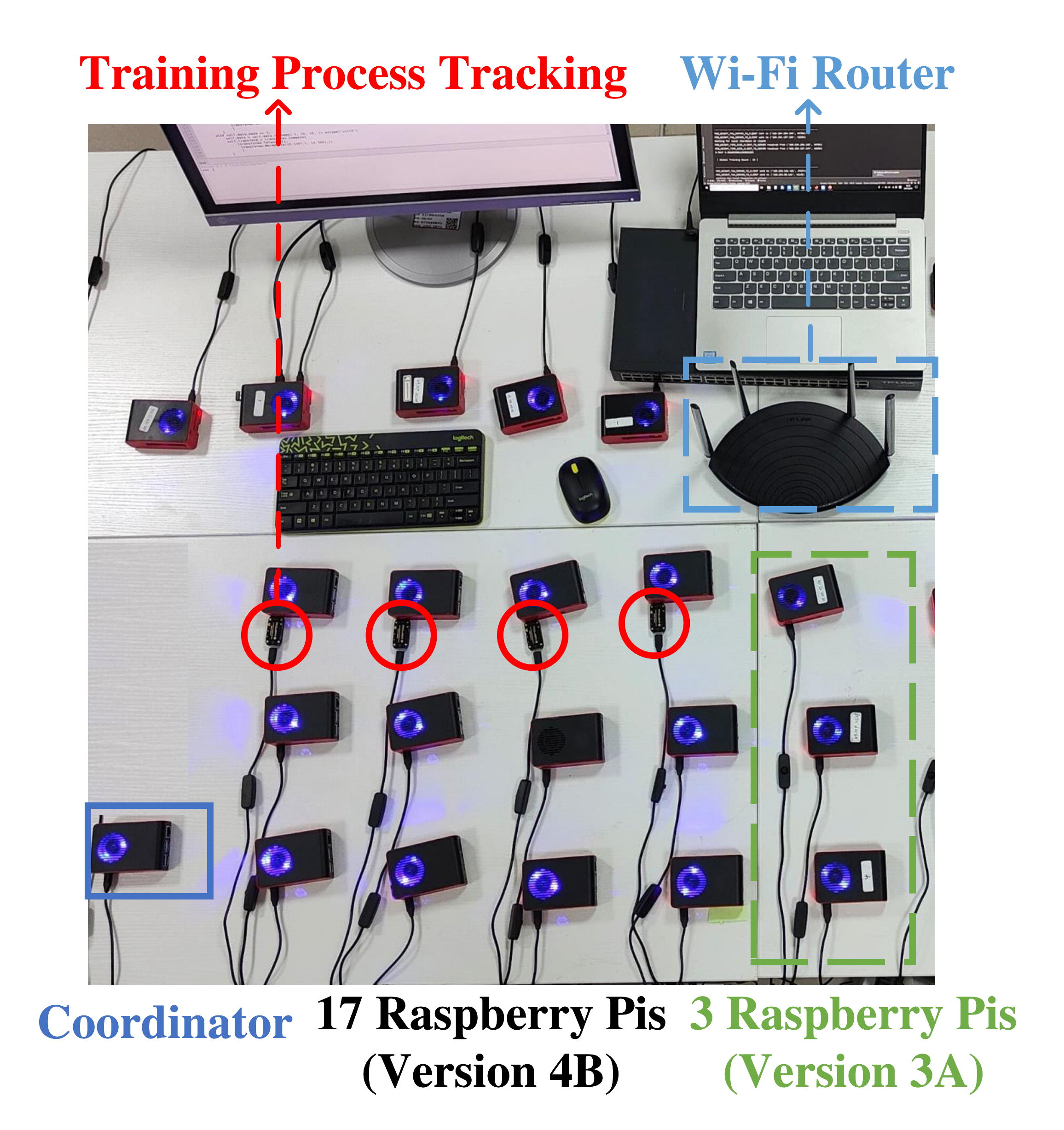}\\
  \caption{Hardware platform consisting of $20$ Raspberry Pis serving as edge servers and one Raspberry Pi serving as the coordinator. Edge servers and coordinator are connected with each other via Wi-Fi links.}
  \label{raspi}
\end{figure}
\subsection{Experimental Setup}

\begin{table}[htbp]
 \centering
 \caption{Experimental Setup}
 \begin{tabular}{|p{2.5cm}|p{2cm}|p{2cm}|}
  \hline
  \multirow{2}{*}{Architecture Setup}&\multicolumn{2}{|c|}{Model Type}\\
  \cline{2-3}
  ~& \multicolumn{2}{|c|}{Multinomial Logistic Regression} \\
  \hline
  Dataset &MNIST & Traffic \\
  \hline
  Input Size& $784\times1$& $2500\times1$\\
  \hline
  Output Size& $10\times1$& $6\times1$\\
  \hline
  Activation Function & \multicolumn{2}{|c|}{Sigmoid}\\
  \hline
  Optimizer& \multicolumn{2}{|c|}{SGD, learning rate 0.01 with decay rate 0.995}\\
  \hline
 \end{tabular}\label{network}
\end{table}

\subsubsection{Hardware Platform} 
To evaluate the performance of our proposed solutions, we develop a hardware platform consisting of 20 Raspberry Pi mini-computers as edge servers, including 17 version 4B with a Quad-core Cortex-A72 (ARM V8) 64-bit SoC operated on 1.5GHz, and 3 version 3A with a Quad-core Cortex-A53 (ARM V8) 64-bit SoC operated on 1.4GHz, as illustrated in Figure \ref{raspi}. All the edge servers are connected to a coordinator (another Raspberry Pi version 4B) via a TP-LINK Wi-Fi Router.
To estimate the time duration of the model training process, we installed 
a multi-function USB multi-meter POWER-Z KM001C to the power port of each edge server to measure and keep track of voltage, current, and power during the entire model training process. We set the sampling rate of each multi-meter to 1 kHz.


\subsubsection{Experimental Setup} We conduct our experiment based on two commonly used datasets: (1) handwritten digit dataset, called MNIST\cite{lecun1998gradient}, which contains $60,000$ images training data samples and $10,000$ testing data samples, and (2) real world network traffic dataset, called ISCX VPN-nonVPN, which consists of $60,000$ traffic data samples for model training and other $12,000$ for testing\cite{draper2016characterization}.
To guarantee the convex property of the classification problem, we follow a commonly adopted setting and use  multinomial logistic regression model. For all the experiments, the global model is initialized with the same weight ${\boldsymbol w}_0$ and the learning rate is set to $\eta_0=0.1$ with a fixed decay $0.995$ per round. To simulate the data heterogeneity scenario, each edge server only has data samples associated with a limited number of randomly selected labels, e.g., in MNIST dataset, each edge server only has images of a limited number of digits, and in traffic dataset, each edge server can only store a limited number of types of traffic data samples. In this section, we compare five different scenarios:

\noindent {\bf Scenario (a)}: we consider handwritten digit dataset and assume each edge server only has data samples associated with a single digit and the data samples of the three low-performance edge servers, i.e., the stragglers (Raspberry Pi 3A), are associated with non-unique digits, i.e., there exist other edge servers (Raspberry Pi 4B) with data samples of the same digits. Since MNIST dataset only has 10 digits 0-9 and we have 20 edge servers, at least two edge servers have data samples associated with the same digit. In this case, even if all three slow edge servers have been removed from participating the model training, there are still able to find data samples associated with the removed digits in the rest of the high-performance edge servers.

\noindent {\bf  Scenario (b)}: we follow the same setting as Scenario (a) but the three low-performance edge servers (Raspberry Pi 3A) consist of data samples associated with unique digits. In this case, if these low-performance edge servers have been removed from the training process, there is impossible to find data samples with the removed digits in the rest of the edge servers.

\noindent {\bf  Scenario (c)}: we consider handwritten digit dataset and assume each edge server has data samples randomly sampled from any five digits and the data samples of the three low-performance edge servers are associated with non-unique digits.

\noindent {\bf  Scenario (d)}: we consider traffic dataset and assume each edge server has data samples randomly sampled from any two types of data traffic and the data samples of the three low-performance edge servers are associated with non-unique types of data traffics.

\noindent {\bf  Scenario (e)}: we consider traffic dataset and assume each edge server has data samples randomly sampled from any four types of data traffic and the data samples of the three low-performance edge servers are associated with non-unique types of data traffics.

\subsection{Measuring Time Consumption of Model Training}

\begin{figure*}[t]
\subfigure[]{
  \begin{minipage}[t]{0.3\linewidth}
   \centering
   \includegraphics[width=5cm]{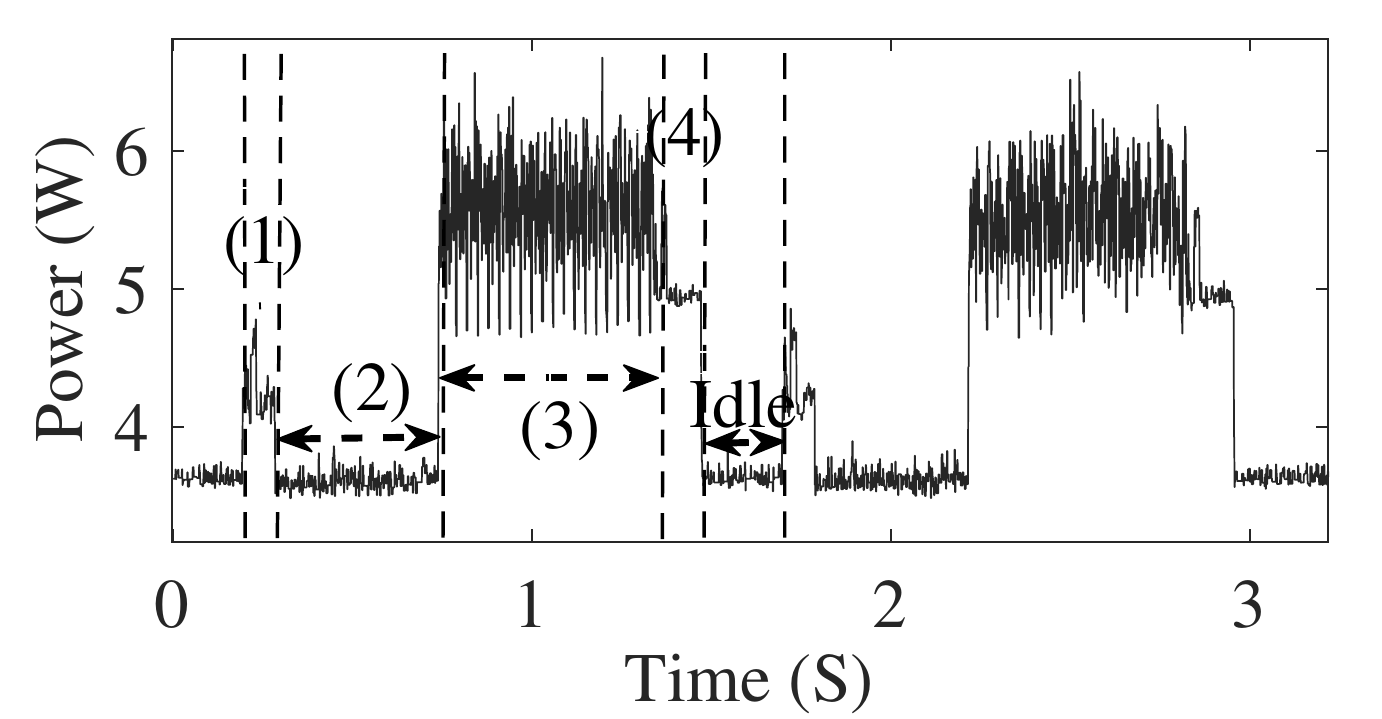}{}
  \end{minipage}
  \label{STATE}%
}
\subfigure[]{
  \begin{minipage}[t]{0.3\linewidth}
   \centering
   \includegraphics[width=5cm]{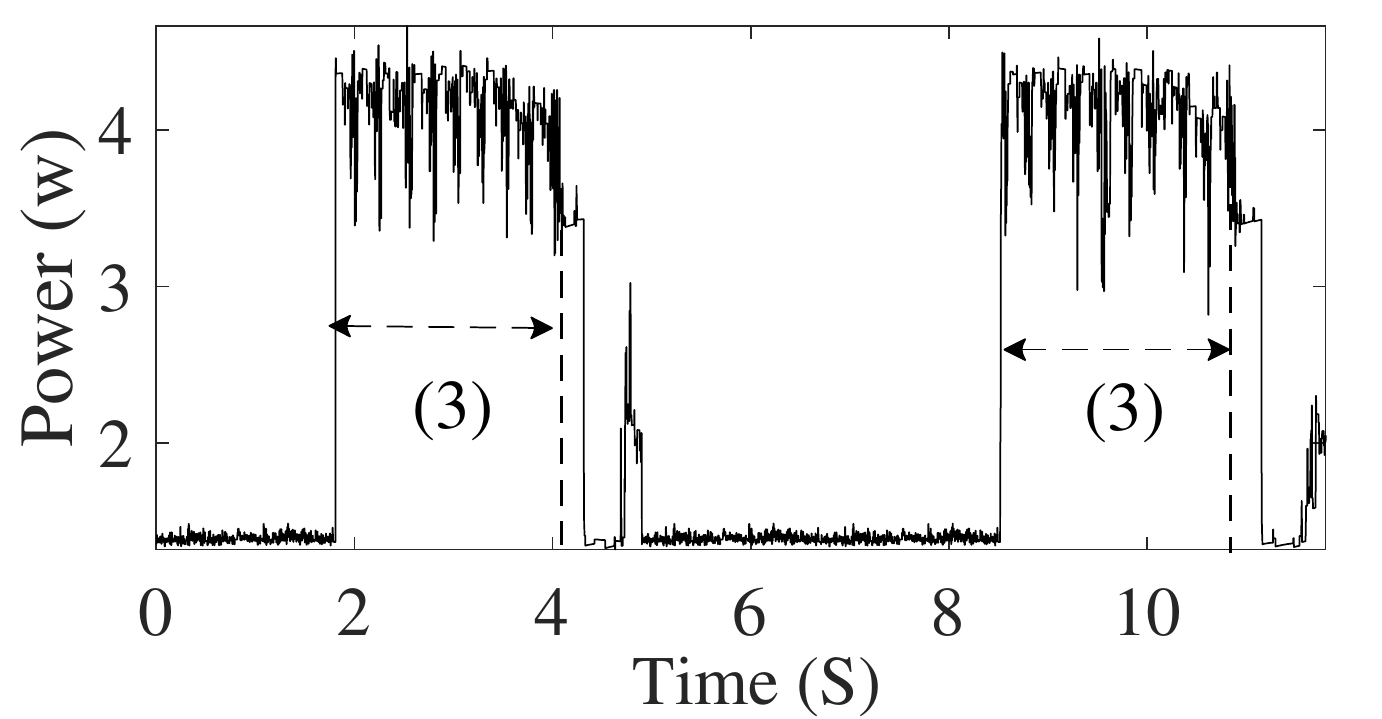}{}
  \end{minipage}%
  \label{Fig_PowerTraceB3000E10}
}
\subfigure[]{
  \begin{minipage}[t]{0.3\linewidth}
   \centering
   \includegraphics[width=5cm]{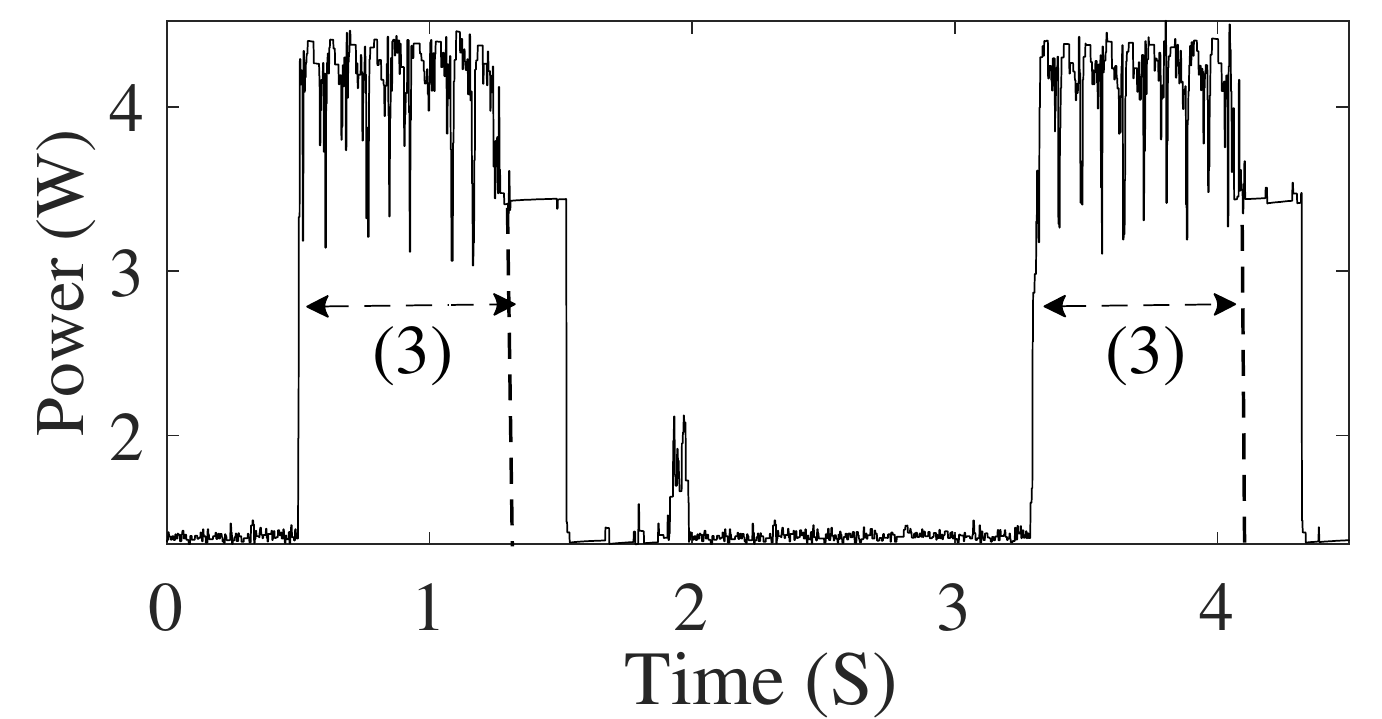}
  \end{minipage}%
  \label{Fig_PowerTraceB1000E10Power}
}
\caption{(a) Recorded power tracking traces of an edge server (Raspberry Pi $4$B) 
with four highlighted steps in each round of model coordination: {\it (1)} model distribution, {\it (2)} data uploading, {\it (3)} local model training, and {\it (4)} model uploading. (b) Comparison of time duration in step {\it (3)} under training parameters $(e=10,n=3000)$, and (c) $(e=10,n=1000)$.}  
\label{local_figure}
\end{figure*}

We record data traces of the power dynamics during the FL model training process performed at each edge server. We can observe a clear dynamic pattern being repeated at each round of model training. We can also observe a clear difference in the power levels at different steps of the model  training process as illustrated in Fig. \ref{STATE}. In particular, we can observe the  four training steps in each round of model training process which are described as follows.


\noindent
{\bf Step (1)--Time duration of model distribution:} 
The coordinator broadcasts the updated global model to a selected subset of the edge servers at the beginning of each round of model training process.  
In Fig. \ref{STATE}, we can observe that 
the time consumption of the model distribution through a wireless network (e.g. 2.4GHz WiFi in our case) is approximately $\zeta=0.2$ sec.

\noindent
{\bf Step (2)--Time duration of data uploading:} 
To simulate the delay caused by uploading of data samples from the data collecting devices to the edge servers, we set the data arrival rate at each high-performance edge server to $50$MB/s (the common data rate of Wi-Fi links) and that at each low-performance edge server to $5$MB/s (the common data rate for Bluetooth connections). Since, in our experiment, all the dataset samples have been pre-loaded to each edge server, to simulate the data uploading delay, we set a pre-calculated idle time at the beginning of every round of model training process as shown in Fig. \ref{STATE}. 

\begin{table*}[!htbp]
 \centering
 \caption{Recorded time duration (in sec) of local model training step (step (3)) under different setups}
\begin{tabular}{|c||c|c|c|c|c|c|c|c|c|c}
\toprule
\multirow{3}*{Raspberry Pi (4B)}&$(n,e)$&(400,10)&(400,20)&(400,30)&(1000,10)&(1000,20)&(1000,30)\\
\cline{2-8}
&MNIST&0.0618&0.1236&0.1856&0.1472&0.2908&0.4377\\
\cline{2-8}
&Traffic&0.3700&0.7402&1.1092&0.4749&0.9473&1.4191\\
\bottomrule
\toprule
\multirow{3}*{Raspberry Pi (3A)}&$(n,e)$&(400,10)&(400,20)&(400,30)&(1000,10)&(1000,20)&(1000,30)\\
\cline{2-8}
&MNIST&0.3815&0.7615&1.1408&0.8013&1.6003& 2.4014
\\
\cline{2-8}
&Traffic&6.0476&12.0934&18.1386&6.5669&13.1309&19.6958\\
\bottomrule
\end{tabular}\label{table_com}
\end{table*}

\begin{table}[!htbp]
 \centering
 \caption{Fitted parameters in the time duration model of local model training step}
\begin{tabular}{|c||c|c|c|c|c}
\toprule
\multirow{3}*{Raspberry Pi (4B)}&Dataset type $\varsigma$&$b_{k,\varsigma}$&$\beta_k$\\
\cline{2-4}
&MNIST&$1.4*10^{-5}$&$5.2*10^{-4}$\\
\cline{2-4}
&Traffic&$1.7*10^{-5}$&0.03\\
\bottomrule
\toprule
\multirow{3}*{Raspberry Pi (3A) }&Dataset type&$b_{k,\varsigma}$&$\beta_k$\\
\cline{2-4}
&MNIST&$7*10^{-5}$&$0.01$\\
\cline{2-4}
&Traffic&$8.6*10^{-5}$&0.57\\
\bottomrule
\end{tabular}\label{paramters_estmated}
\end{table}

\noindent
{\bf Step (3)--Time duration of local model training:} Once edge server $k$ obtains the model distributed by the coordinator, it will perform $e$ local SGD iterations to update its model. 
In our data traces in Fig. \ref{STATE}, we can observe  that each local iteration performed by an edge server corresponds to a peak in power level. 
For $e$ local iterations, we can observe $e$ peaks during the local model training step. The value of $n_k$ mainly affects the duration of each local iteration (time duration between two consecutive peaks) as shown in Fig. \ref{Fig_PowerTraceB3000E10} and Fig. \ref{Fig_PowerTraceB1000E10Power}.  In Table \ref{table_com}, we present the time duration of local training step under different combinations of $e$ and $n_k$ at different edge servers (i.e. Raspberry Pi (version $4$B) and (version $3$A)) based on different datasets. We can observe that for a fixed $n_k$, the time duration of local model training step increases almost linearly with $e$. Similarly, when $e$ is fixed, the time consumption of each local iteration increases almost linearly with $n_k$. To summarize, the time duration of local model training step can be modeled as
\begin{eqnarray}
  \nu_k (e_k,n_k|\varsigma) &=& e_k(b_{k,\varsigma} n_k+\beta_k).
\end{eqnarray}
where $b_{k,\varsigma}$ characterizes the time duration  for edge server $k$ to process each training data sample and $\beta_{k,\varsigma}$ is the time duration for performing local SGD and updating the local models. We present the fitted parameters of $b_{k,\varsigma}$ and $\beta_k$ under different setup in Table \ref{paramters_estmated}.

\noindent
{\bf Step (4)--Time duration of model uploading:} Each edge server uploads its local model to the coordinator after $e$ local SGD iterations. In our experiment, all edge servers and the coordinator are connected to a Wi-Fi router, we can observe that the time duration of model uploading does not affected by model training parameters such as $n_k$ or $e$. We can therefore assume the time duration of this step is a constant, i.e., $u=0.2$s as observed in Fig. \ref{STATE}.

\subsection{Experimental Results}

\begin{figure}[ht]
\subfigure[]{
  \begin{minipage}[t]{0.45\linewidth}
   \centering
   \includegraphics[width=4cm]{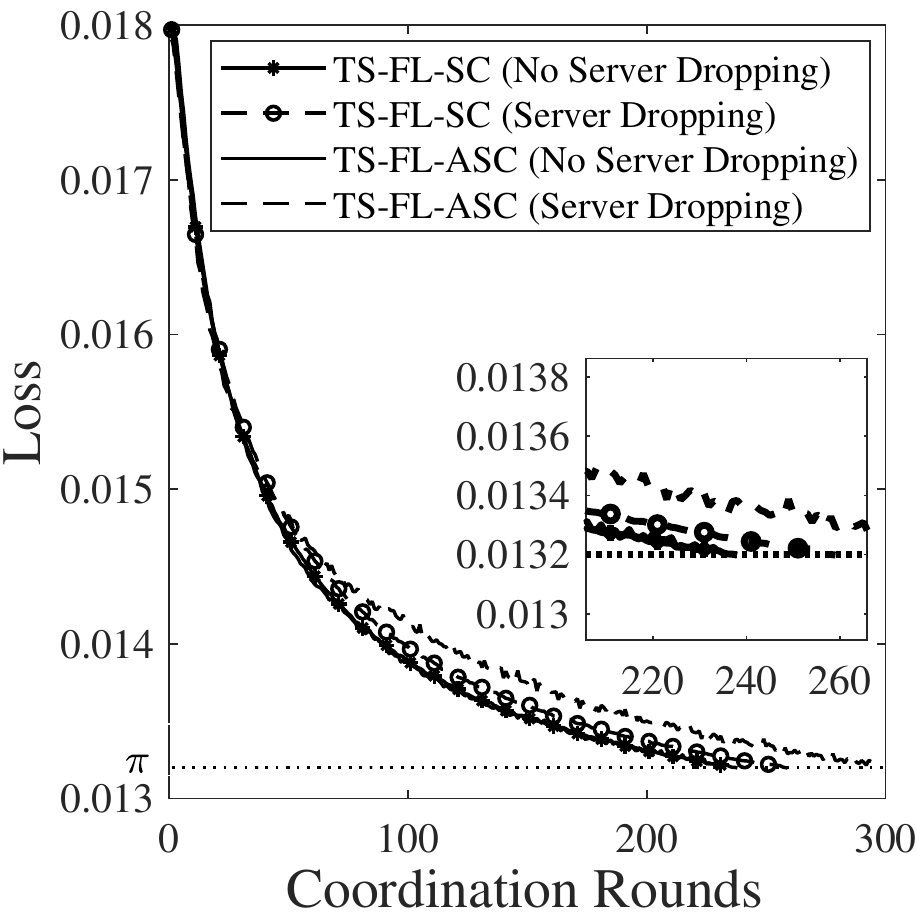}\label{conv_a}
    \vspace{-0.14in}
  \end{minipage}%
}
\subfigure[]{
  \begin{minipage}[t]{0.45\linewidth}
   \centering
   \includegraphics[width=4cm]{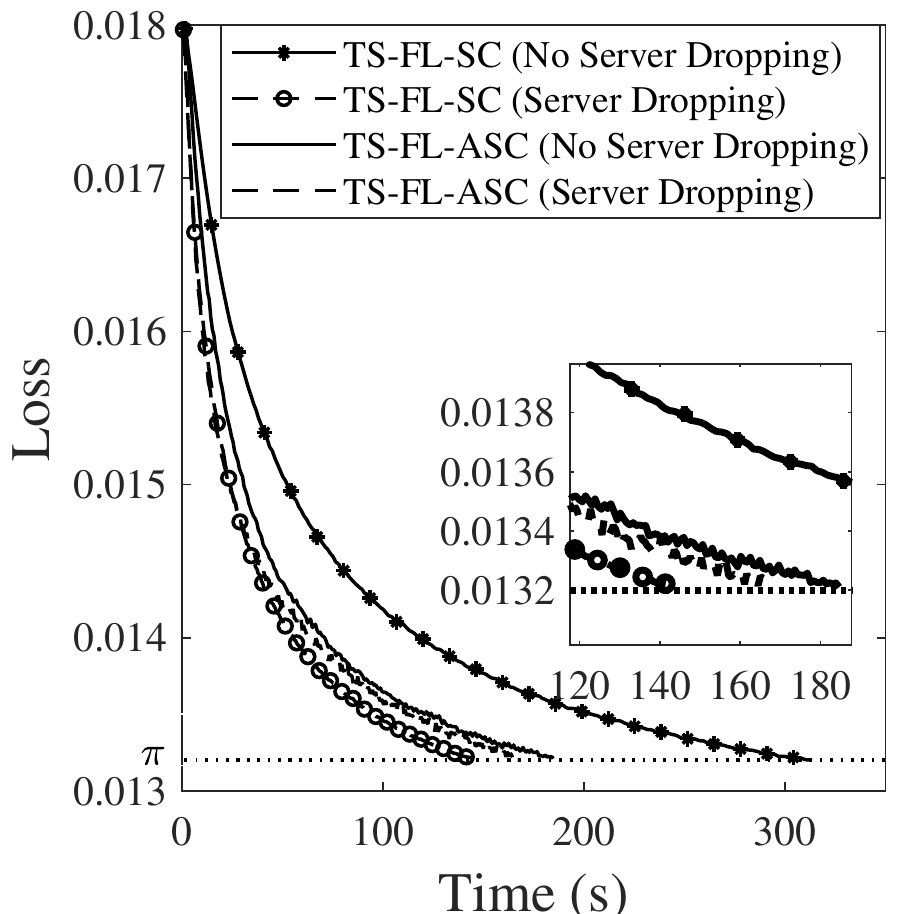}\label{time_a}
   \vspace{-0.14in}
  \end{minipage}%
  }
  \label{Fig_ScenarioA}
\caption{(a) Number of coordination rounds and (b) model training time of TS-FL-SC and TS-FL-ASC under scenario (a).}
\end{figure}

\begin{figure}[ht]
\subfigure[]{
  \begin{minipage}[t]{0.45\linewidth}
   \centering
   \includegraphics[width=4cm]{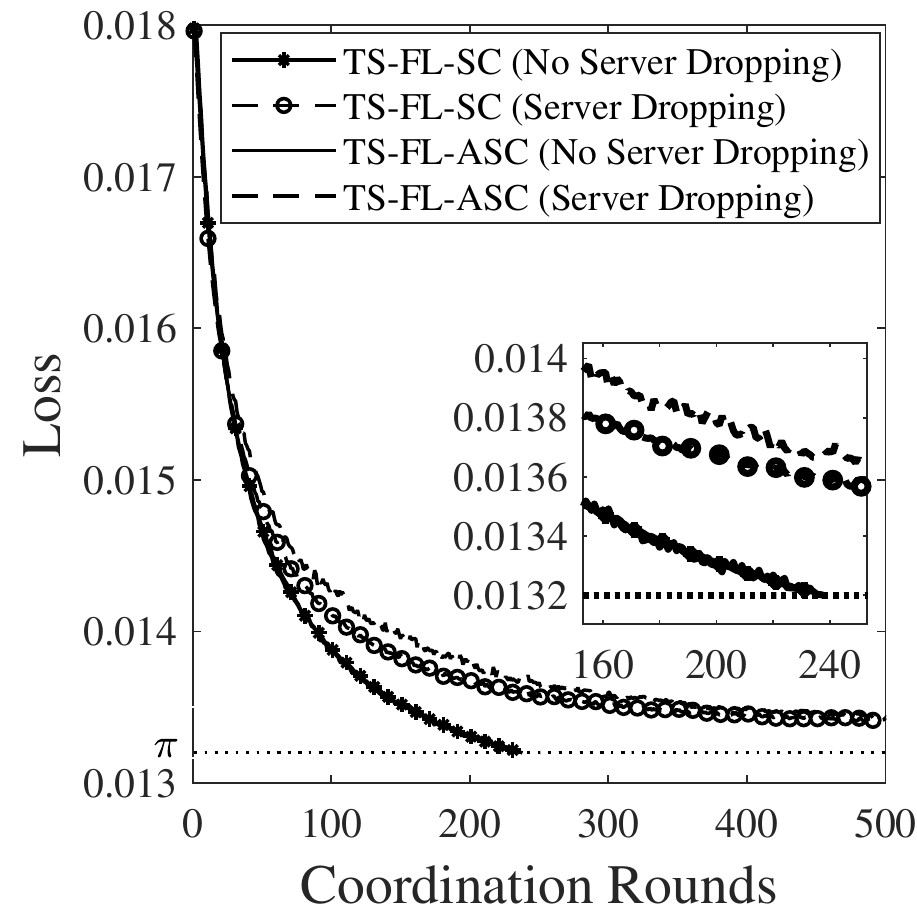}\label{conv_b}
    \vspace{-0.14in}
  \end{minipage}%
}
\subfigure[]{
  \begin{minipage}[t]{0.45\linewidth}
   \centering
   \includegraphics[width=4cm]{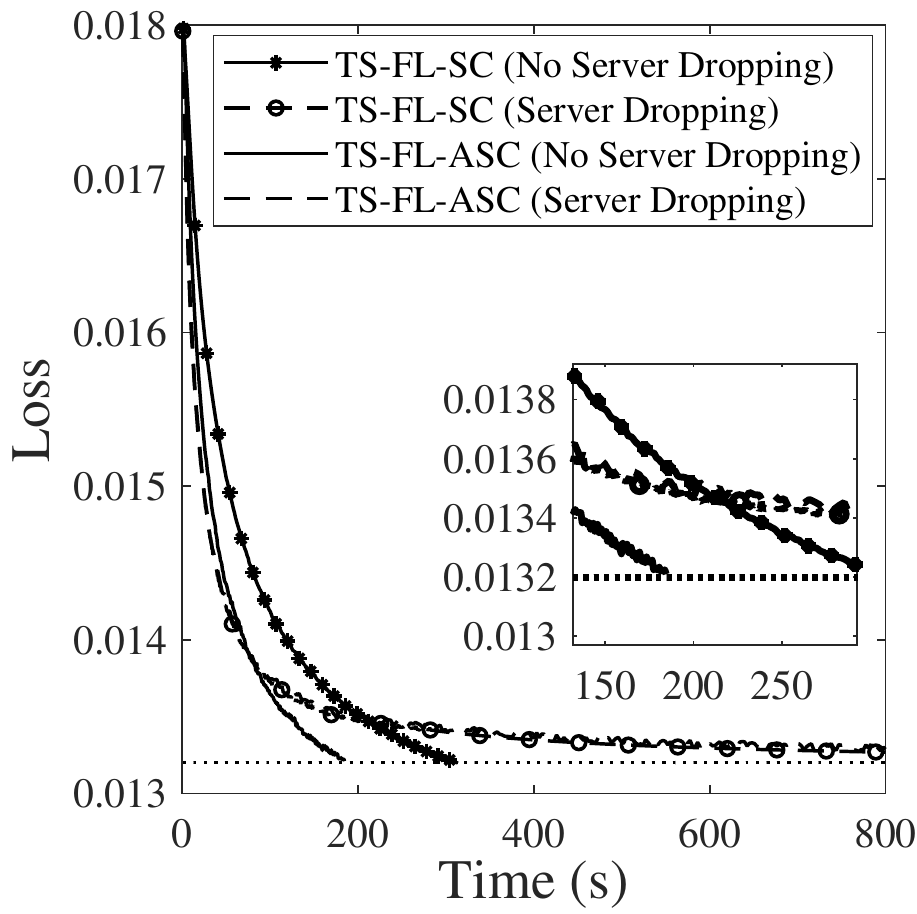}\label{time_b}
   \vspace{-0.14in}
  \end{minipage}%
  }
\caption{(a) Number of coordination rounds and (b) model training time of TS-FL-SC and TS-FL-ASC under scenario (b).}
\label{Fig_ScenarioB}
\end{figure}

\begin{figure}[ht]
\subfigure[]{
  \begin{minipage}[t]{0.45\linewidth}
   \centering
   \includegraphics[width=4cm]{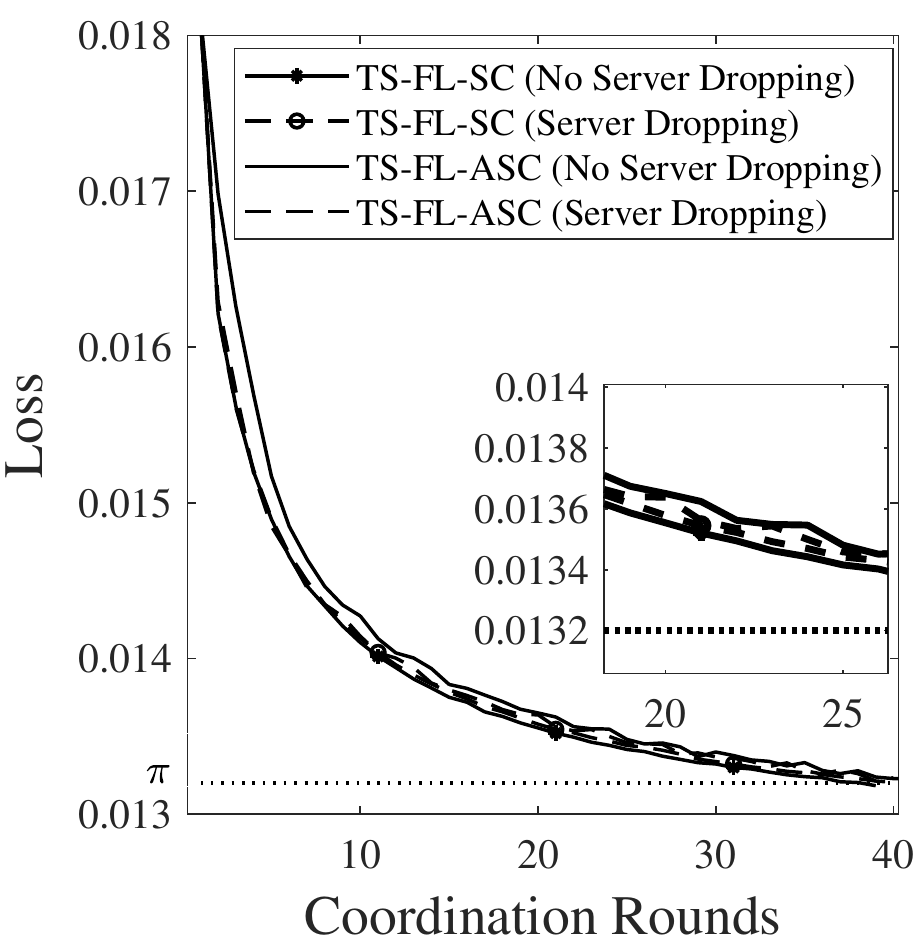}\label{conv_c}
    \vspace{-0.14in}
  \end{minipage}%
}
\subfigure[]{
  \begin{minipage}[t]{0.45\linewidth}
   \centering
   \includegraphics[width=4cm]{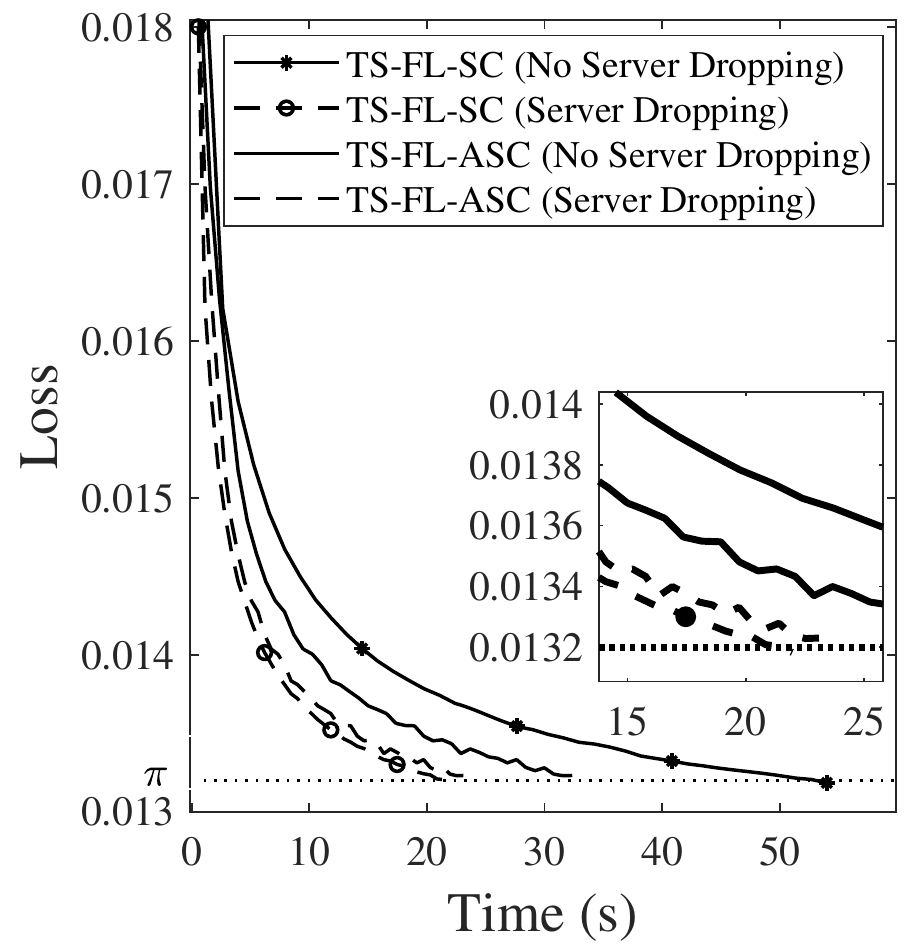}\label{time_c}
   \vspace{-0.14in}
  \end{minipage}%
  }
\caption{(a) Number of coordination rounds and (b) model training time of TS-FL-SC and TS-FL-ASC under scenario (c).}
\label{Fig_ScenarioC}
\end{figure}

\begin{figure}[ht]
\subfigure[]{
  \begin{minipage}[t]{0.45\linewidth}
   \centering
   \includegraphics[width=4cm]{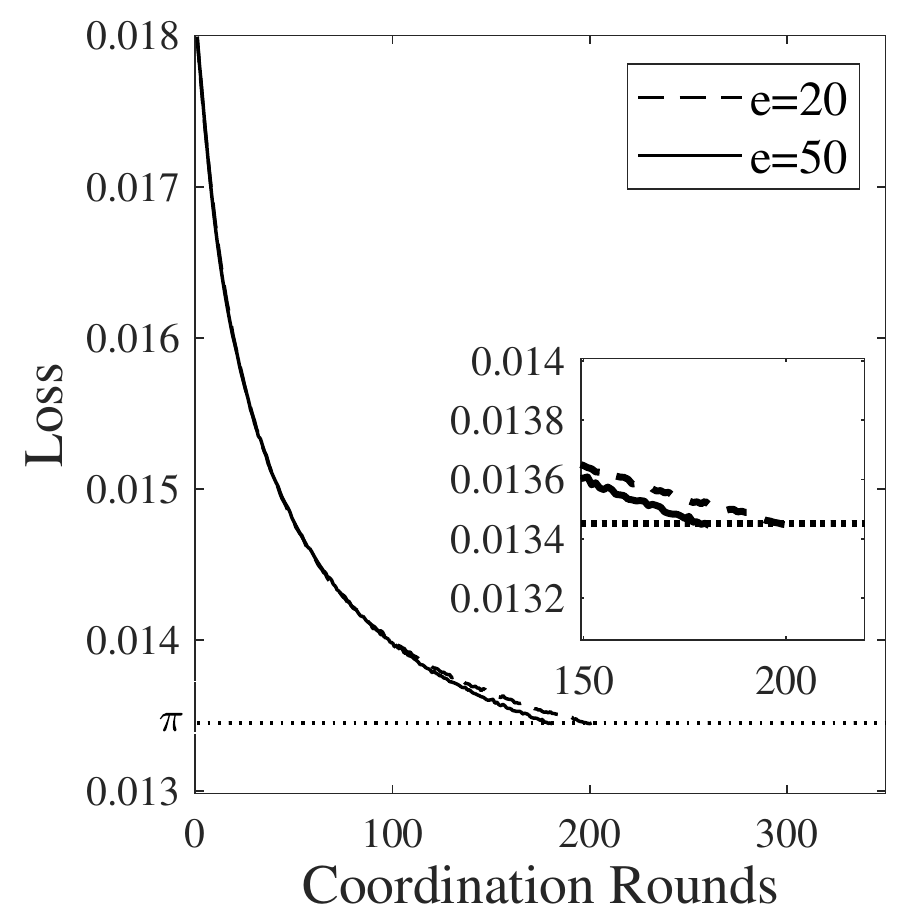}\label{conv_e}
   \vspace{-0.14in}
  \end{minipage}%
  }
\subfigure[]{
  \begin{minipage}[t]{0.45\linewidth}
   \centering
   \includegraphics[width=4cm]{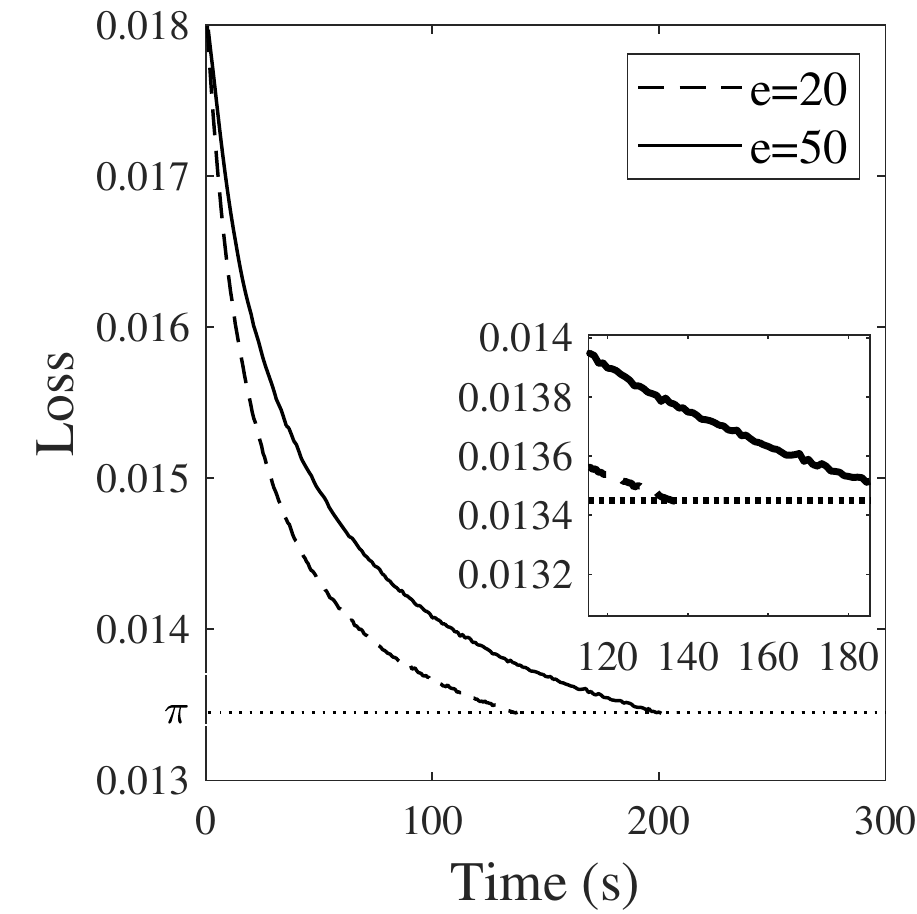}\label{time_e}
    \vspace{-0.14in}
  \end{minipage}%
}
\caption{(a) Number of coordination rounds and (b) model training time of TS-FL-SC with server dropping under different local epoch numbers $e$ in scenario (a).}
\label{Fig_LossVSLocalEpochScenarioA}
\end{figure}

\begin{figure}[ht]
\subfigure[]{
  \begin{minipage}[t]{0.45\linewidth}
   \centering
   \includegraphics[width=4cm]{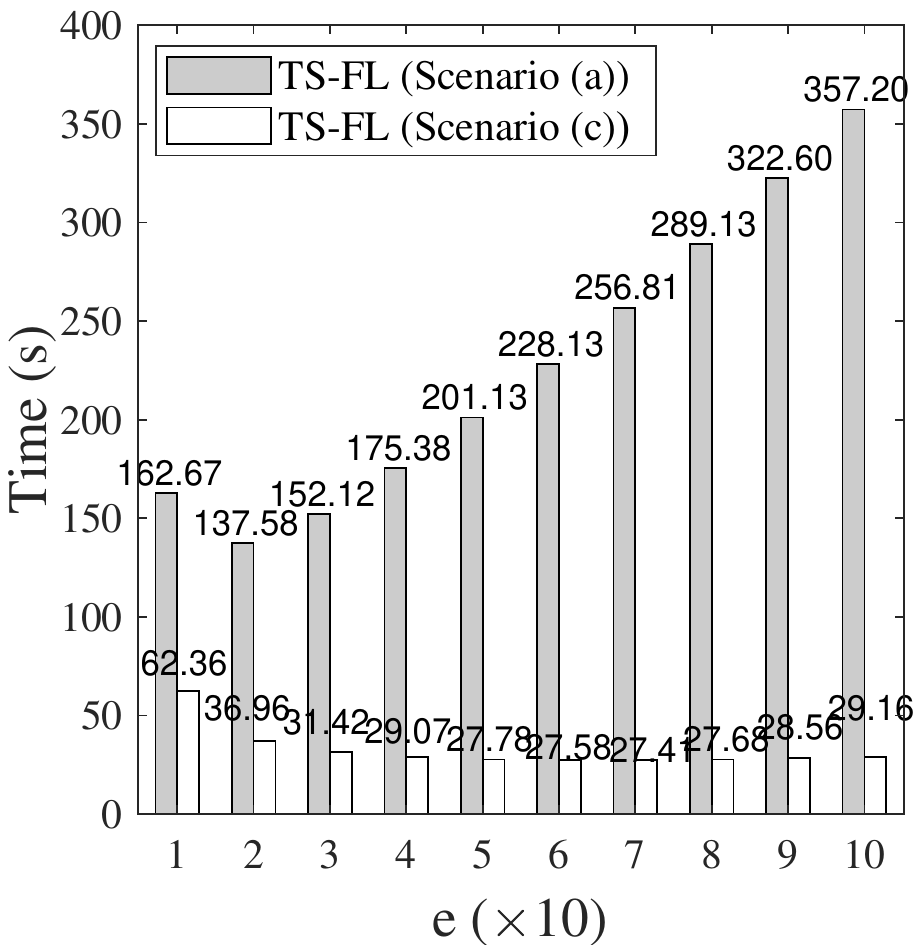}\label{mnist_e}
    \vspace{-0.14in}
  \end{minipage}%
}
\subfigure[]{
  \begin{minipage}[t]{0.45\linewidth}
   \centering
   \includegraphics[width=4cm]{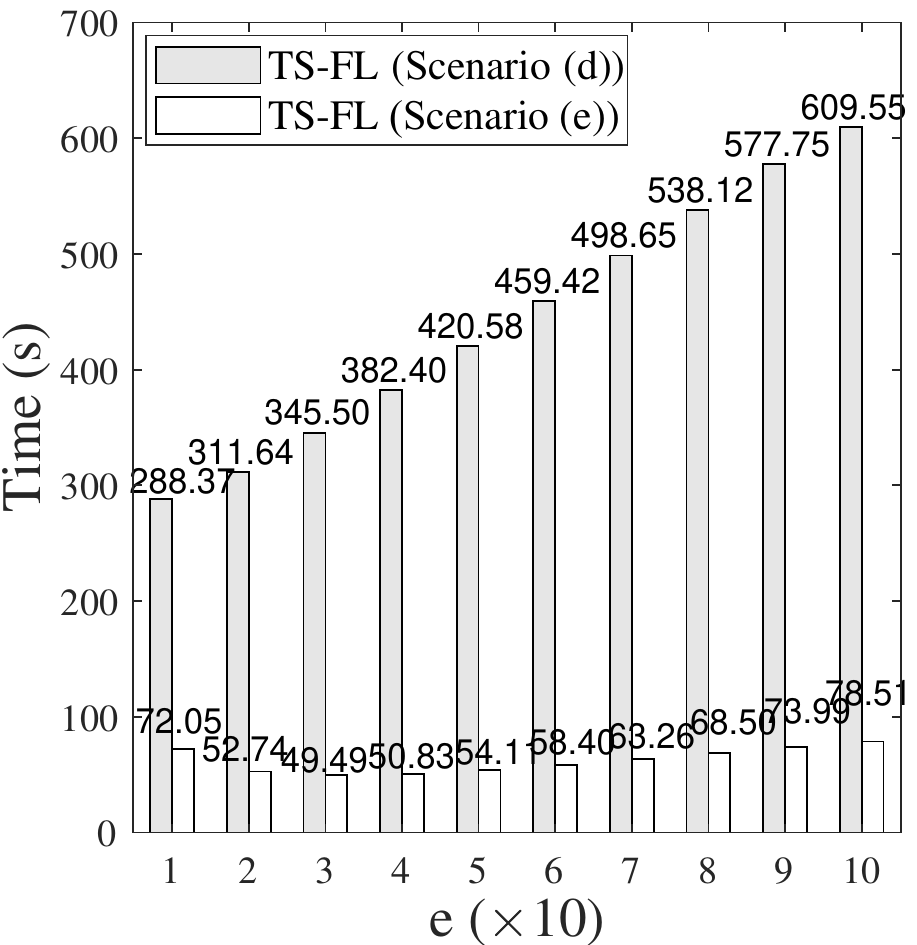}\label{traffic_e}
   \vspace{-0.14in}
  \end{minipage}%
  }
\caption{Required time consumption for training a model with target loss $(\pi=0.0132)$ under different local epoch numbers $e$ and a fixed mini-batch size $n=500$ based on (a) handwritten digit dataset and (b) traffic dataset.}
\label{Fig_TimeVSepoch}
\vspace{-0.1in}
\end{figure}

\begin{figure}[ht]
\subfigure[]{
  \begin{minipage}[t]{0.45\linewidth}
   \centering
   \includegraphics[width=4cm]{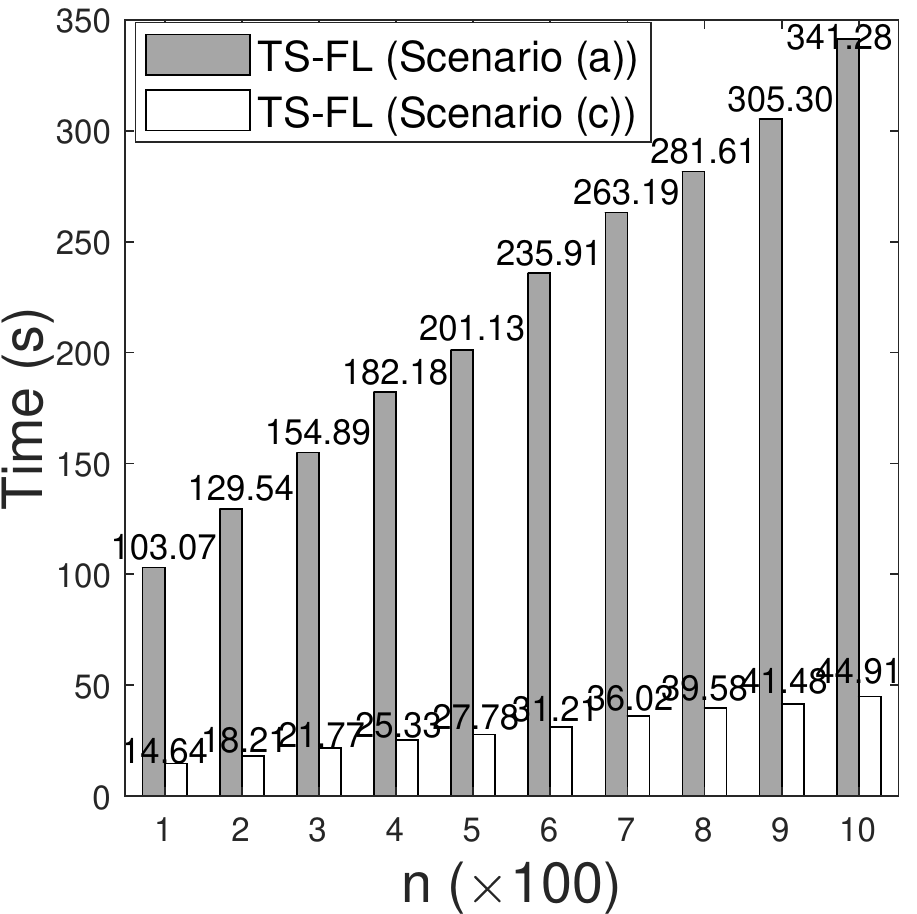}\label{mnist_n}
    \vspace{-0.14in}
  \end{minipage}%
}
\subfigure[]{
  \begin{minipage}[t]{0.45\linewidth}
   \centering
   \includegraphics[width=4cm]{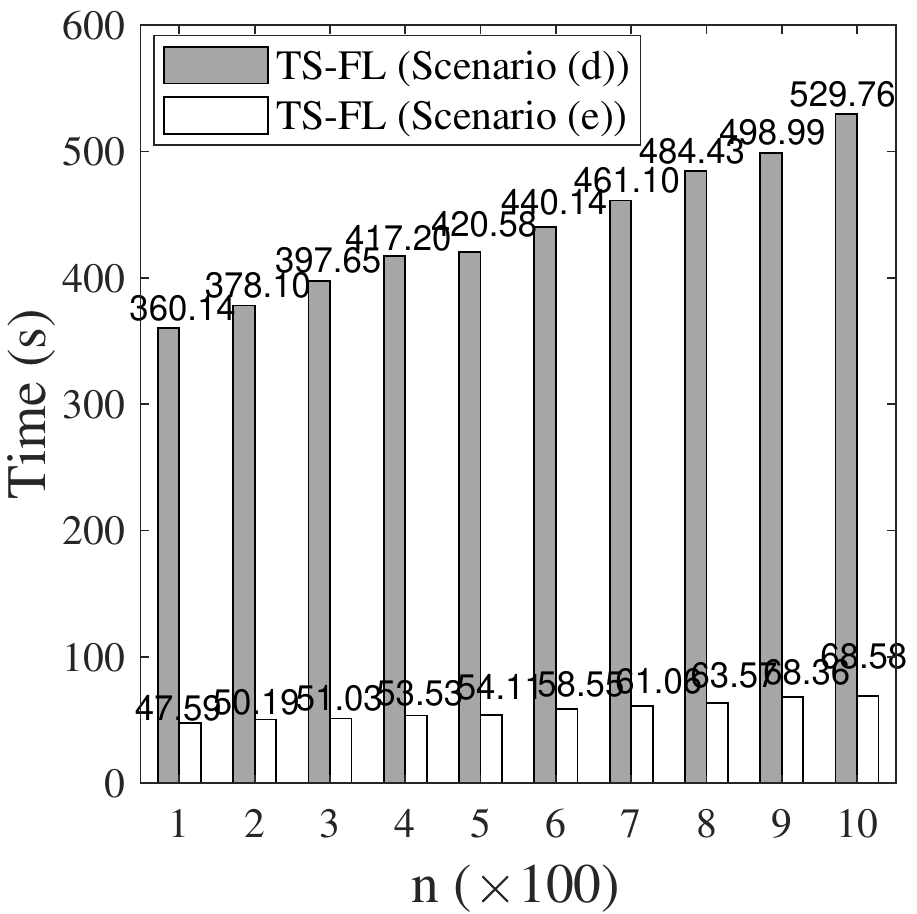}\label{traffic_n}
   \vspace{-0.14in}
  \end{minipage}%
  }
\caption{Required time consumption for training a model with target loss $(\pi=0.0132)$ under different mini-batch sizes $n$ and a fixed epoch number $e=50$ based on (a) handwritten digit dataset and (b) traffic dataset.}
\label{Fig_TimeVSbatchsize}
\vspace{-0.1in}
\end{figure}

\begin{table*}[t]
\footnotesize
\centering
\blu{
\caption{COMPARISON OF MODEL TRAINING TIME OF SCHEDULING-BASED SOLUTION AND SERVER DROPPING}
\renewcommand\arraystretch{1}
\begin{tabular}{ccp{1.3cm}<{\centering}p{1.3cm}<{\centering}p{1.3cm}<{\centering}p{1.3cm}<{\centering}p{1.3cm}<{\centering}p{1.3cm}<{\centering}p{1.3cm}<{\centering}p{1.3cm}}
\hline
Datasets & (e, n) & (10, 200) & (20, 200) & (30, 200) & (40, 200) & (50, 200)& (60, 200)\\ \hline \hline
\multirow{3}{*}{ MNIST} &
     \multicolumn{1}{|c} {Sched.-based}  & 293.12  & 269.64  &402.23   & 534.51 &667.76 &799.39 \\
    \cline{2-8}
     &
     \multicolumn{1}{|c} {Server Drop.}  & 162.67  &137.58   & 152.12  & 175.38 & 201.13 & 228.13\\
    \cline{2-8}
    &  \multicolumn{1}{|c} {Time Saving} &\bf 130.45  &\bf  132.06  &\bf 250.11   &\bf  359.13&\bf  466.63& \bf 571.26 \\
     \hline
\multirow{3}{*}{Traffic} &
     \multicolumn{1}{|c} {Sched.-based}  & 1372.32  & 2835.49  & 4189.72  & 5550.56 & 6915.60 & 8280.64\\
    \cline{2-8}
     &
     \multicolumn{1}{|c} {Server Drop.}  & 288.37  & 311.64  &345.50   &382.40  &420.58 &459.65 \\
    \cline{2-8}
    &  \multicolumn{1}{|c} {Time Saving} &\bf 1083.95  & \bf 2523.85  & \bf 3844.22  & \bf 5168.16 & \bf 6495.02& \bf 7820.99 \\
     \hline
\end{tabular}
}
\label{Table_ModelTrainingTimeCompare}
\end{table*}


As mentioned earlier, the convergence rate and time consumption of model training are two different concepts. The later is more complex and can be  affected jointly by computation and communication delays influenced  by multiple parameters. In Fig. \ref{Fig_ScenarioA}, we compare the required number of coordination rounds  and  time duration of training a model at a target loss $(\pi=0.0132)$ \blu{(corresponding to the model accuracy of 83.30\% and 83.50\% for MNIST and traffic datasets, respectively)}  using TS-FL-SC and TS-FL-ASC with and without using server dropping under scenario (a) where the low-performance  edge servers do not have any uniquely labeled data samples. We can observe that the values of the loss functions under different numbers of coordination rounds and model training times exhibit different trends. In particular, in terms of convergence rate, the TS-FL-SC without server dropping has the fastest convergence, followed by TS-FL-ASC without server dropping. In other words, the server dropping cannot accelerate the convergence rate of an FEI system when measuring based on the number of coordination rounds. However, when we consider the time duration of the model training process, server dropping can significantly reduce the required time duration for training a satisfactory model (with target loss value). As observed in Fig. \ref{Fig_ScenarioA}(b), the TS-FL-SC with server dropping can reduce the overall time required for model training almost by half, compared to TS-FL-SC without dropping any low-performance servers. We can also observe that TS-FL-ASC can also reduce the time duration of model training, especially when all edge servers are selected to participate in the model training. The performance improvement provided by TS-FL-ASC, compared to TS-FL-SC, is reduced when server dropping is enabled. This is because the slowest edge server always dominates the overall time consumption of the model training and therefore when being removed from the model training process, the load forwarding between the slow and fast edge servers will have a limited effect in reducing the model training time.

In Fig. \ref{Fig_ScenarioB}, we present the number of coordination rounds and time consumption required for training a satisfactory model with a target loss $(\pi=0.0132)$ under scenario (b). Recall that in scenario (b), some low-performance edge servers have data samples with unique labels. In this case, we can observe that removing the low-performance edge servers from participating in model training will slow the convergence rate, i.e., neither TS-FL-SC nor TS-FL-ASC can converge to the target loss within 500 rounds of coordination and if without server dropping, both solutions can obtain the target model after around 220 coordination rounds. This observation becomes more noticeable if we consider the required time for training a satisfactory model. More specifically, in Fig. \ref{Fig_ScenarioB}(b), we can observe that the TS-FL-ASC without server dropping is the fastest solution to train the model, followed by TS-FL-SC without server dropping. The later solution results in almost 40\% more time consumption for training the same model.

In Fig. \ref{Fig_ScenarioC}, we evaluate the required coordination rounds and time for reaching a target loss $(\pi=0.0132)$ under scenario (c). In this case, we can observe that in terms of required number of coordination rounds for training a model, the  convergence performance of different solutions are quite similar, i.e., the difference in the required numbers of coordination rounds for reaching the same target loss  when applying different solutions  is within 2-3 coordination rounds. However, if we consider the overall time consumption for training the model, different solutions exhibits quite different convergence performance. More specifically, server dropping reduces the model training time in TS-FL-SC and TS-FL-ASC by around 63\% and 28\%, respectively. This once again verifies our previous observation that the model training time and convergence rate are different concepts and the former metric provides more important insight, especially in the smart applications and servers requiring time-sensitive learning and model updating.

As mentioned earlier, the model training time is closely related to two important model training parameters local epoch number $e$ and mini-batch size $n$. In particular, in Sections \ref{Sec_S_TSFL} and \ref{sec_A_TSFL}, we have proved that if the overall time required for training any desirable model can be approximated by our derived convergence bound, the model training time can be considered as biconvex over $n$ and $e$. In other words, if the approximated model training time is convex over $n$ (or $e$) under any given $e$ (or $n$). Let us now verify this results by comparing the overall time  under different $e$ and $n$. In particular, in Fig. \ref{Fig_TimeVSepoch} and \ref{Fig_TimeVSbatchsize}, we compare the real model training time recorded in our hardware platform  for reaching a target loss value $(\pi=0.0132)$ when either $e$ or $n$ is fixed while the other variables are changing into different values. In Fig. \ref{Fig_TimeVSepoch}, we can observe that the real model training time verifies our theoretical results and the overall time required for training a satisfactory model exhibits convexity in our recorded traces. 
More specifically, there exists an optimal value of $e$ which can minimize time consumption of model training under a given $n$. More specifically, the optimal epoch number $e^*$ that can reach the target accuracy fastest in scenario (a) and scenario (c) in handwritten digit dataset are $20$ and $60$ with the recorded total time duration for model training at $137.58$s and $27.41$s, respectively. For the traffic dataset, the minimum model training time is achieved by selecting the smallest epoch number $e^*=10$ in scenario (d) resulting $288.37$s for training the model. In scenario (e) however, the minimum model training time is achieved at $e^*=30$ resulting in $49.49$s. 
In Fig. \ref{Fig_TimeVSbatchsize}, we can observe that the required time for training the model increases almost linearly with the mini-batch sizes $n$. This is because a large mini-batch sizes not only means longer delay for data sample uploading from data collecting devices to the edge servers but also more time consumed for model training at each edge server.

\blu{
To compare our proposed solution with the recent state-of-the-art, in Table \ref{Table_ModelTrainingTimeCompare}, we have presented the overall time consumption of model training process using our proposed solution and the scheduling-based approach proposed in \cite{luo2020cost} under two different datasets MNIST and traffic data with target model accuracy at 83.30\% and 80.43\% in scenarios (a) and (d), respectively. Since the scheduling-based approach in \cite{luo2020cost} is only applicable in the synchronous coordination scenario, to make the result comparable, we also present the model training time of server dropping in TS-FL-SC scenario. We can observe that the proposed server dropping solution can significantly reduce the overall runtime when training a model with the same guaranteed accuracy, especially when the local model training and updating speed difference between slow and the fast edge servers is large. In particular, our proposed server dropping solution can reduce up to 71.43\% and 94.45\% model training times in MNIST and traffic dataset, respectively, compared to the scheduling-based approach.
} 

\section{Conclusion}\label{sec_conclusion}
This paper investigated the real-time learning for FEI systems with system and dataset heterogeneity. A novel framework, called TS-FL, was proposed to minimize the overall run-time for collaboratively training a shared ML model with desirable accuracy. Training acceleration solutions for both TS-FL-SC and TS-FL-ASC were proposed.
For the TS-FL-SC, a server dropping-based solution was proposed to allow some slow-performance edge servers to be removed from participating in the model training if their impact on the resulting model accuracy is limited. A joint optimization algorithm is proposed to minimize the overall time consumption of model training by selecting participating edge servers, the local epoch number (the number of model training iterations per coordination), and the data batch size (the number of data samples for each model training iteration). For the TS-FL-ASC, a load forwarding-based solution was proposed to allow the slow edge server to offload part of its training samples to trusted edge servers with higher processing capability. New theoretical convergence bound for TS-FL-SC and TS-FL-ASC are derived for model training based on non-i.i.d. datasets. We develop a hardware prototype to evaluate the model training time of a heterogeneous FEI system. Experimental results show that our proposed TS-FL-SC and TS-FL-ASC can provide up to 63\% and 28\% of reduction, in the overall model training time, respectively, compared with traditional FL solutions.

\section*{Acknowledgment}
Y. Xiao was supported in part by the National Natural Science Foundation of China under grant 62071193 and the Key R \& D Program of Hubei Province of China under grants 2021EHB015 and 2020BAA002. G. Shi was supported in part by the National Natural Science Foundation of China under grants 61871304, 61976169. Y. Xiao and G. Shi were supported in part by the major key project of Peng Cheng Laboratory (No. PCL2021A12). M. Krunz was supported by the National Science Foundation (grants 1910348, 1731164, 1813401, 2229386, and IIP-1822071) and by the Broadband Wireless Access \& Applications Center (BWAC). Any opinions, findings, conclusions, or recommendations expressed in this paper are those of the author(s) and do not necessarily reflect the views of NSF.

\bibliography{ref}
\bibliographystyle{IEEEtran}

\begin{IEEEbiography}[{\includegraphics[width=1.1in,height=1.3in,clip,keepaspectratio]{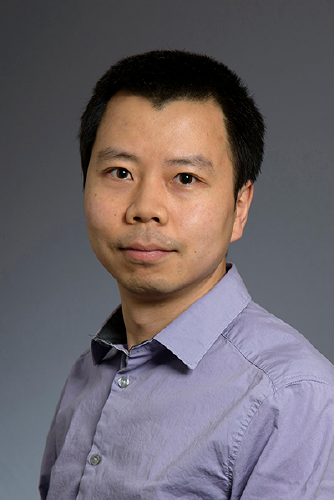}}]{Yong Xiao}(Senior Member, IEEE) received his B.S. degree in electrical engineering from China University of Geosciences, Wuhan, China in 2002, M.Sc. degree in telecommunication from Hong Kong University of Science and Technology in 2006, and his Ph. D degree in electrical and electronic engineering from Nanyang Technological University, Singapore in 2012. He is now a professor in the School of Electronic Information and Communications at the Huazhong University of Science and Technology (HUST), Wuhan, China. He is also with Peng Cheng Laboratory, Shenzhen, China and Pazhou Laboratory (Huangpu), Guangzhou, China. He is the associate group leader of the network intelligence group of IMT-2030 (6G promoting group) and the vice director of 5G Verticals Innovation Laboratory at HUST. Before he joins HUST, he was a  research assistant professor in the Department of Electrical and Computer Engineering at the University of Arizona where he was also the center manager of the Broadband Wireless Access and Applications Center (BWAC), an NSF Industry/University Cooperative Research Center (I/UCRC) led by the University of Arizona. His research interests include machine learning, game theory, distributed optimization, and their applications in cloud/fog/mobile edge computing, green communication systems, wireless communication networks, and Internet-of-Things (IoT).
\end{IEEEbiography}


\begin{IEEEbiography}[{\includegraphics[width=1.1in,height=1.3in,clip,keepaspectratio]{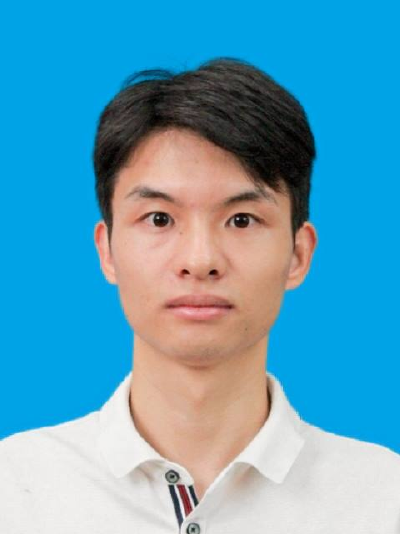}}]{Xiaohan Zhang} received his B.S. degree in electronic information engineering from China University of Geosciences, Wuhan, China in 2017, M.Sc. degree in information and communication engineering from Huazhong University of Science and Technology, Wuhan, China in 2022. His research interests include federated learning and cloud/edge computing networks.
\end{IEEEbiography}

\begin{IEEEbiography}[{\includegraphics[width=1.1in,height=1.3in,clip,keepaspectratio]{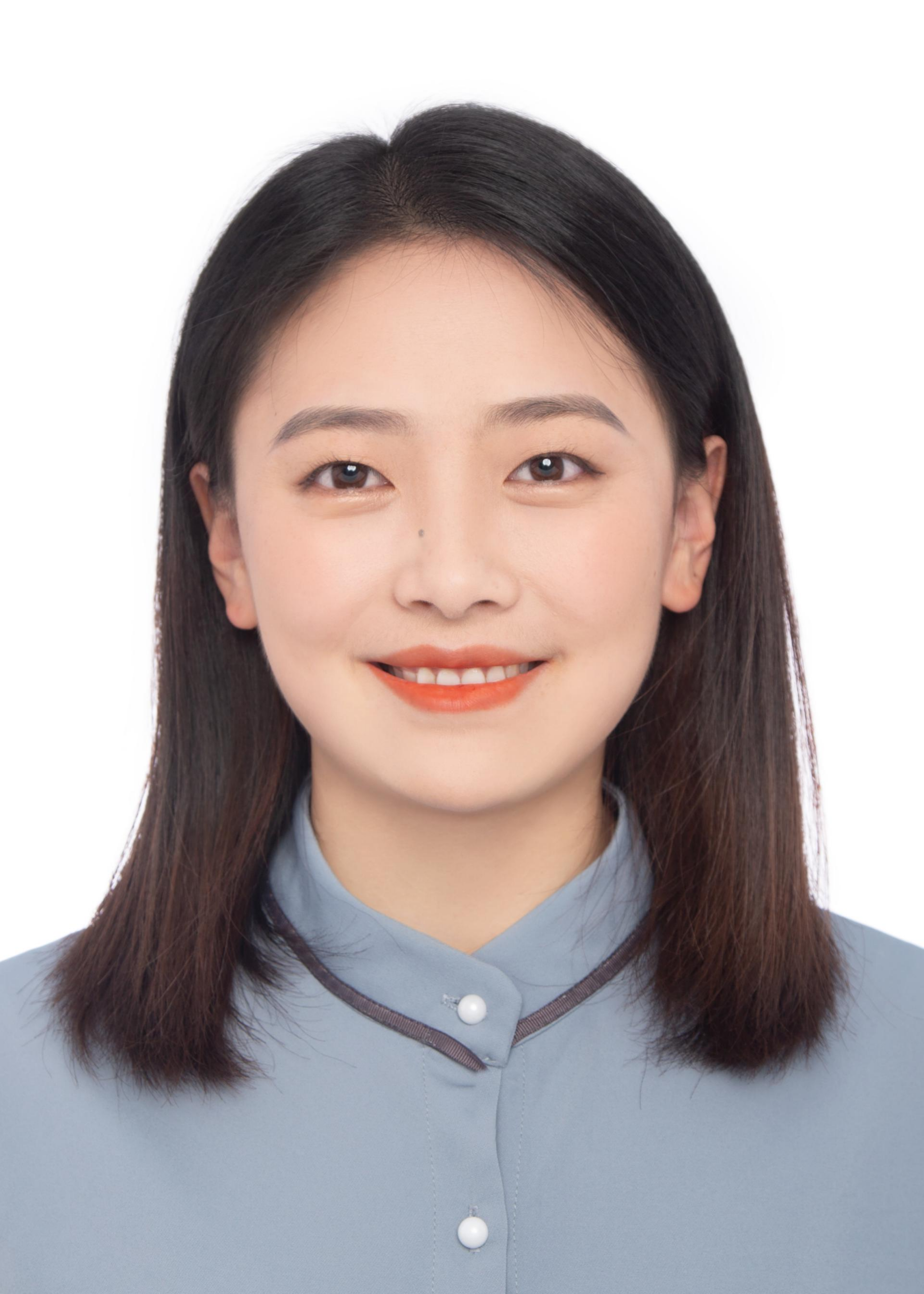}}]{Yingyu Li} (Member, IEEE) received the B.Eng. degree in electronic information engineering and the Ph.D. degree in circuits and systems from the Xidian University, Xi'an, China, in 2012 and 2018, respectively. From 2014 to 2016, she was a Research Scholar with the Department of Electronic Computer Engineering at the University of Houston, TX, USA. She was a postdoctoral researcher in the School of Electronic Information and Communications at Huazhong University of Science and Technology from 2018 to 2021. She is now an associate professor at the School of Mechanical Engineering and Electronic Information, China University of Geosciences (Wuhan). Her research interests include semantic communications, edge intelligence, green communication networks, and IoT.
\end{IEEEbiography}

\begin{IEEEbiography}[{\includegraphics[width=1.1in,height=1.3in,clip,keepaspectratio]{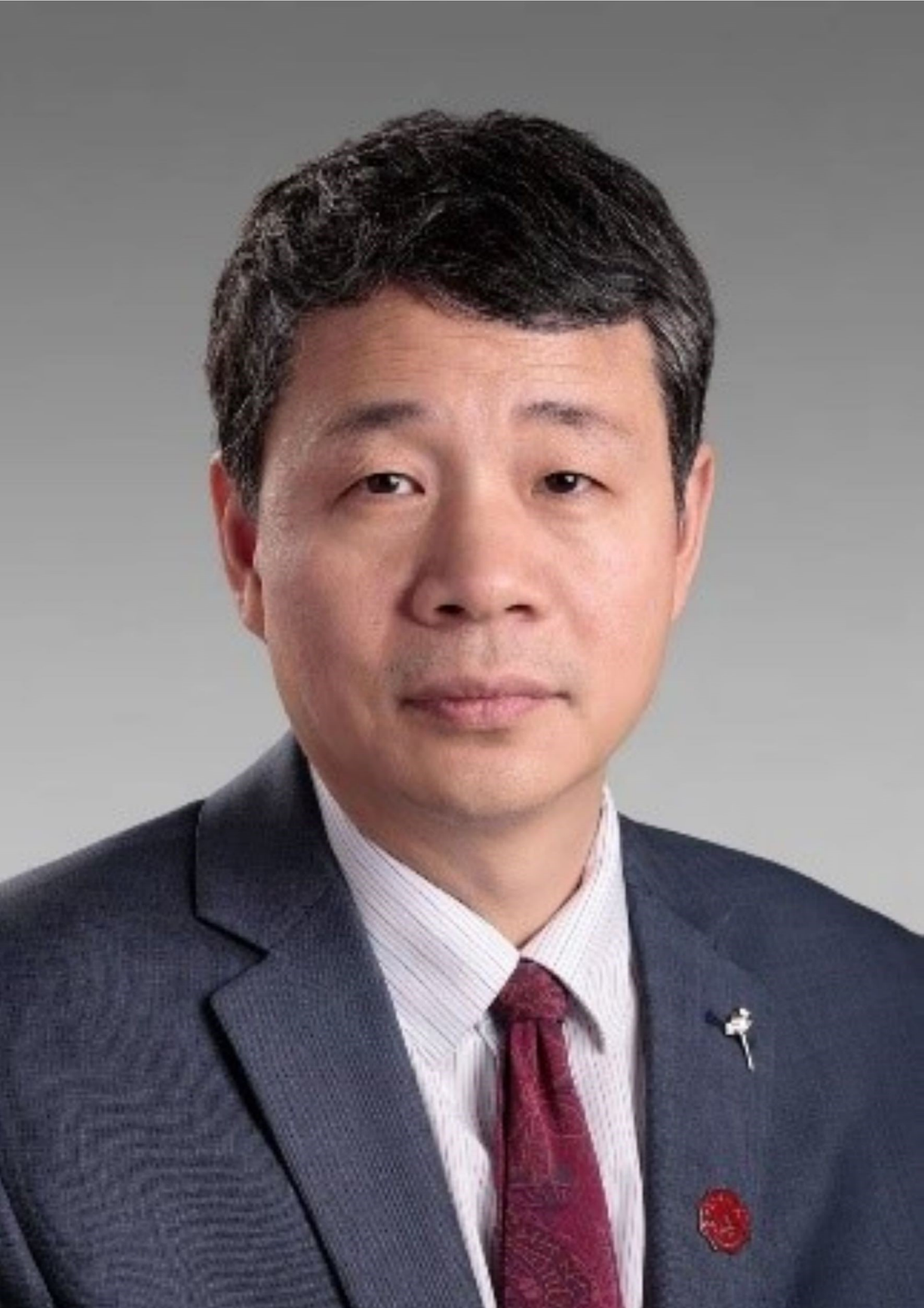}}]{Guangming Shi} (Fellow, IEEE) received the M.S. degree in computer control and the Ph.D. degree in electronic information technology from Xidian University, Xi’an, China, in 1988, and 2002, respectively. He was the vice president of Xidian University from 2018 to 2022. Currently, he is  the Vice Dean of Peng Cheng Laboratory and a Professor with the School of Artificial Intelligence, Xidian University. He is an IEEE Fellow, the chair of IEEE CASS Xi’an Chapter, senior member of ACM and CCF, Fellow of Chinese Institute of Electronics, and Fellow of IET. He was awarded Cheung Kong scholar Chair Professor by the ministry of education in 2012. He won the second prize of the National Natural Science Award in 2017. His research interests include Artificial Intelligence, Semantic Communications, and Human-Computer Interaction.
\end{IEEEbiography}

\begin{IEEEbiography}[{\includegraphics[width=1.1in,height=1.3in,clip,keepaspectratio]{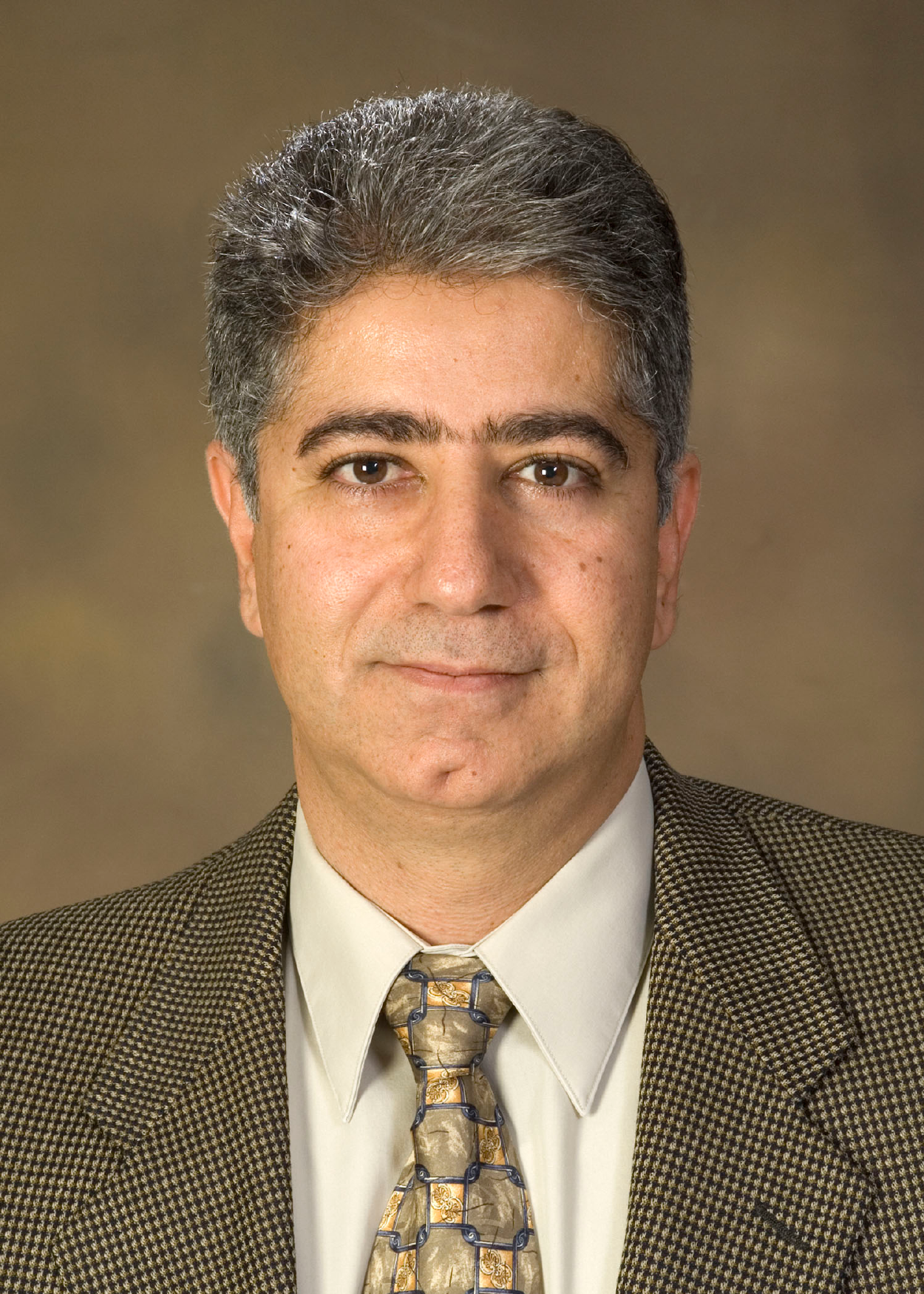}}]{Marwan Krunz}(Fellow, IEEE) is a Regents Professor at the University of Arizona. He holds the Kenneth VonBehren Endowed Professorship in ECE and is also a professor of computer science. He directs the Broadband Wireless Access and Applications Center (BWAC), a multi-university NSF/industry center that focuses on next-generation wireless technologies. He also holds a courtesy appointment as a professor at University Technology Sydney. Previously, he served as the site director for  Connection One, an NSF/industry-funded center of five universities and 20+ industry affiliates. Dr. Krunz’s research is in the fields of wireless communications, networking, and security, with recent focus on applying AI and machine learning techniques for protocol adaptation, resource management, and signal intelligence. He has published more than 320 journal articles and peer-reviewed conference papers, and is a named inventor on 12 patents. His latest h-index is 60. He is an IEEE Fellow, an Arizona Engineering Faculty Fellow, and an IEEE Communications Society Distinguished Lecturer (2013-2015). He received the NSF CAREER award. He served as the Editor-in-Chief for the IEEE Transactions on Mobile Computing. He also served as editor for numerous IEEE journals. He was the TPC chair for INFOCOM’04, SECON’05, WoWMoM’06, and Hot Interconnects 9. He was the general vice-chair for WiOpt 2016 and general co-chair for WiSec’12. Dr. Krunz served as chief scientist/technologist for two startup companies that focus on 5G and beyond wireless systems.
\end{IEEEbiography}

\begin{IEEEbiography}[{\includegraphics[width=1.1in,height=1.3in,clip,keepaspectratio]{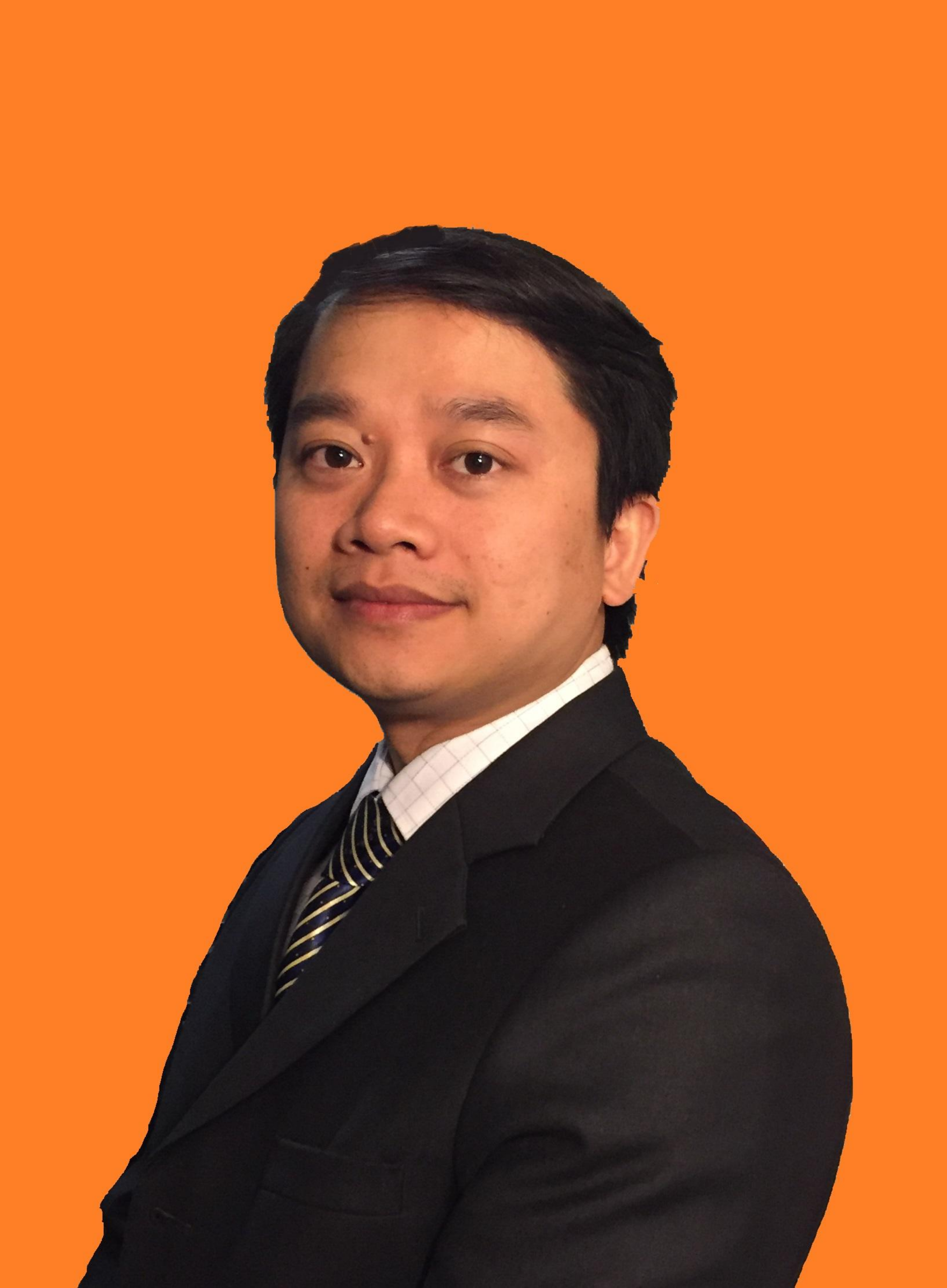}}]{Diep N. Nguyen} (Senior Member, IEEE) received the M.E. degree in electrical and computer engineering from the University of California at San Diego (UCSD), La Jolla, CA, USA, in 2008, and the Ph.D. degree in electrical and computer engineering from The University of Arizona (UA), Tucson, AZ, USA, in 2013. He is currently a Faculty Member with the Faculty of Engineering and Information Technology, University of Technology Sydney (UTS), Sydney, NSW, Australia. Before joining UTS, he was a DECRA Research Fellow with Macquarie University, Macquarie Park, NSW, Australia, and a Member of the Technical Staff with Broadcom Corporation, San Jose, CA, USA, and ARCON Corporation, Boston, MA, USA, and consulting the Federal Administration of Aviation, Washington, DC, USA, on turning detection of UAVs and aircraft, and the U.S. Air Force Research Laboratory, Wright-Patterson Air Force Base, OH, USA, on anti-jamming. His research interests include computer networking, wireless communications, and machine learning application, with emphasis on systems’ performance and security/privacy. Dr. Nguyen received several awards from LG Electronics, UCSD, UA, the U.S. National Science Foundation, and the Australian Research Council. He is currently an Editor, an Associate Editor of the IEEE Transactions on Mobile Computing, IEEE Communications Surveys \& Tutorials (COMST), IEEE Open Journal of the Communications Society, and Scientific Reports (Nature's).
\end{IEEEbiography}

\begin{IEEEbiography}[{\includegraphics[width=1.1in,height=1.3in,clip,keepaspectratio]{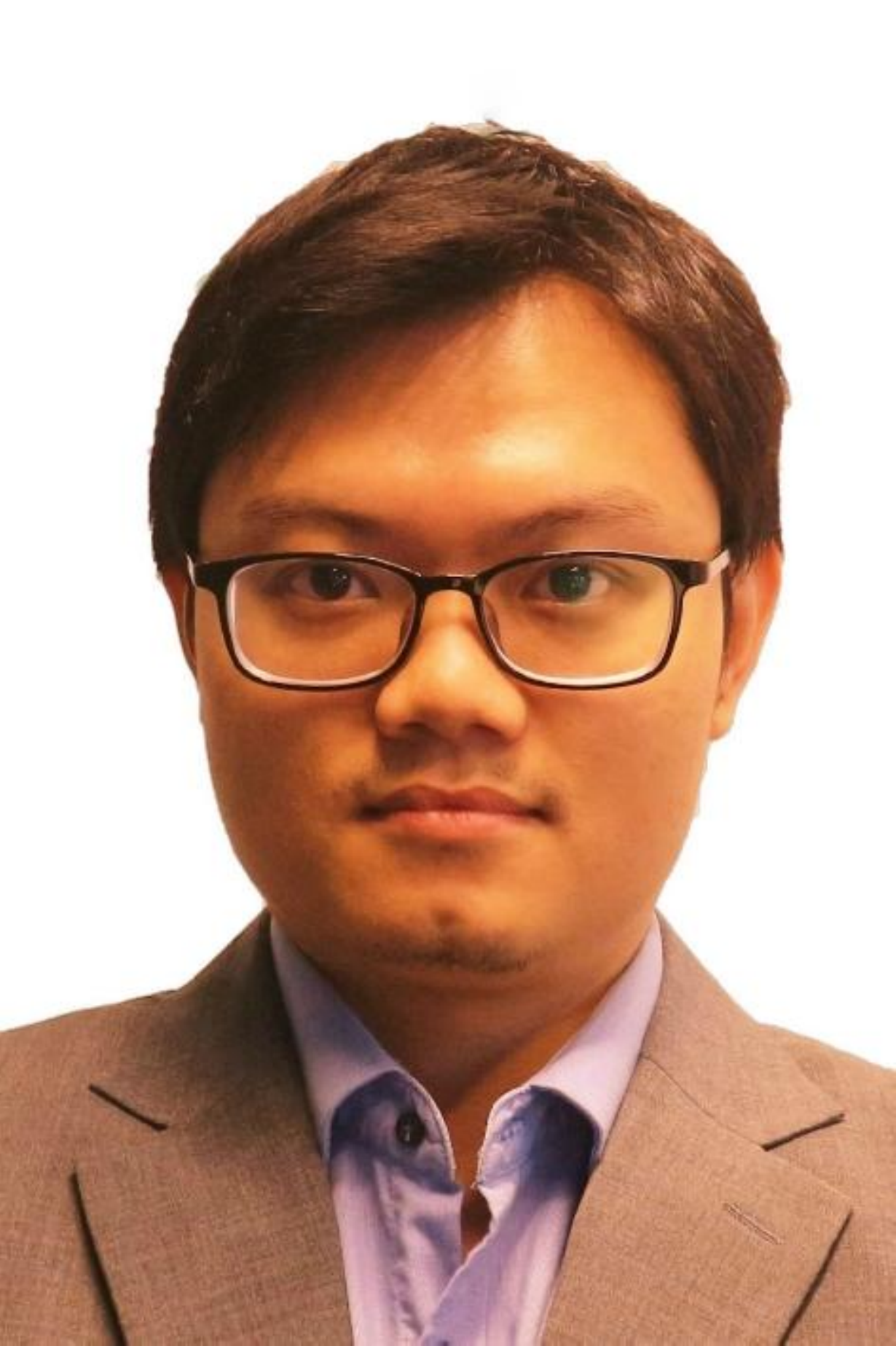}}]{Dinh Thai Hoang} is currently a faculty member at the School of Electrical and Data Engineering, University of Technology Sydney, Australia. He received his Ph.D. in Computer Science and Engineering from the Nanyang Technological University, Singapore, in 2016. His research interests include emerging wireless communications and networking topics, especially machine learning applications in networking, edge computing, and cybersecurity. He has received several awards, including the Australian Research Council and IEEE TCSC Award for Excellence in Scalable Computing (Early Career Researcher). He is an Editor of IEEE Transactions on Wireless Communications, IEEE Transactions on Cognitive Communications and Networking, IEEE Transactions on Vehicular Technology, and Associate Editor of IEEE Communications Surveys \& Tutorials.
\end{IEEEbiography}

\newpage

\appendices
\section{Proof of Theorem \ref{Theorem_TSSC_convergence_rate}}
\label{Appendix_ProofTheorem1}

Let us now derive the convergence rate of TS-FL-SC with server dropping.
To facilitate our analysis, we introduce a virtual sequence  ${\bar{\boldsymbol{w}}_{{\cal M},t}}=\sum_{k\in \mathcal{M}}{p_k \boldsymbol{w}_{k,t}}$ to represent the (virtual) sequence of equivalent global model if all the local models at a subset $\cal M$ of edge servers can be aggregated at every local SGD iteration. Note that, 
after $T$ SGD iterations, the resulting global model aggregated by the coordinator should be equivalent to the most updated model in the virtual sequence, i.e.,  ${{\boldsymbol{w}}_{{\cal M},T}}={\bar{\boldsymbol{w}}_{{\cal M},T}}$. We can therefore focus on deriving the convergence rate of the virtual sequence during the rest of this proof.

Note that our proposed TS-FL-SC with server dropping is different from the traditional FL with partial server participation. In the latter solution, every edge server will have equal chance of being  selected to participate in the model training process and therefore no edge servers will be removed from the entire model training process.

One of the key issue of TS-FL-SC with server dropping is to estimate the gap between the model ${\bar{\boldsymbol{w}}_{{\cal M},t}}$ trained with a subset $\cal M$ of edge servers and the global optimal model $\boldsymbol{w}_{\cal K}^*$ trained with the set $\cal K$ of all $K$ edge servers, i.e.,  $\|{\bar{\boldsymbol{w}}_{{\cal M},t}}-\boldsymbol{w}_{\cal K}^*\|^2$. 

In \cite{li2019convergence}, it has proved the following upper bound for model training with a set $\cal M$ edge server with full participation.

\begin{lemma}\cite[Theorem 1]{li2019convergence}
\label{Lemma_convergence_rate}
Suppose the following assumptions hold: (1) $F_1,F_2,...,F_K$ are all $L-$smooth and $\mu-$convex, i.e., $ {\frac{\mu}{2}}\|{\boldsymbol w}-{\boldsymbol w}'\|^2$ $\leq$ $F_k(\boldsymbol w)-F_k({\boldsymbol w}')$ $+$ $({\boldsymbol w}-{\boldsymbol w}')^T \nabla F_k({\boldsymbol w}')$ $\leq$ ${\frac{L}{2}}\|{\boldsymbol w}-{\boldsymbol w}'\|^2$ for all ${\boldsymbol w},{\boldsymbol w}'\in \mathbb{R}^d$; (2) data samples $x_k$ are uniformly randomly sampled from ${\cal D}_k$; and (3) the stochastic gradient satisfies $\mathbb{E}\|\nabla l_k({\boldsymbol w},x_k)\|^2 \leq G^2$ and $\mathbb{E}\|\nabla l_k({\boldsymbol w},x_k)-\nabla F_k({\boldsymbol w})\|^2$ $\leq$ ${\sigma_k}^2$.
If $\kappa={\frac{L}{\mu}}$, $\gamma=\max \{8\kappa,e\}$, and the learning rate $\eta_t$ satisfies $\eta_t={2 \over {\mu(\gamma+t)}}$, we have
\begin{eqnarray}
\mathbb{E}(||{\bar{\boldsymbol{w}}_{{\cal M},t}}-\boldsymbol{w}_{\mathcal{M}}^*||^2) &\leq& \Delta
\label{Li_convergence}
\end{eqnarray}
where $\Delta$ is given by
\begin{eqnarray}
\Delta&=&\dfrac{4}{\mu^2 (\gamma+t)}({{\sum_{k \in \cal M}{p_k^2{\sigma_k}^2 \over n_k}+8e^2 G^2}}+6L D^*_{\cM}(F)\nonumber \\
&&+{\mu^2 (\gamma+1) \over 4}\|{\boldsymbol{w}}_{0}-\boldsymbol{w}_{\mathcal{M}}^*\|^2).
\end{eqnarray}
\end{lemma}


Decomposing the term $||{\bar{\boldsymbol{w}}_{{\cal M},t}}-\boldsymbol{w}_{\mathcal{M}}^*||^2$, and we can write
\begin{eqnarray}
\lefteqn{\mathbb{E}||{\bar{\boldsymbol{w}}_{{\cal M},t}}-\boldsymbol{w}_{\mathcal{M}}^*||^2-\mathbb{E}(||{\bar{\boldsymbol{w}}_{{\cal M},t}}-\boldsymbol{w}_{\cal K}^*||^2)} \nonumber \\
 &=& 2\langle{\bar{\boldsymbol{w}}_{{\cal M},t}}-\boldsymbol{w}_{\mathcal{M}}^*+\boldsymbol{w}_{\mathcal{M}}^*-\boldsymbol{w}_{\cal K}^*,\boldsymbol{w}_{\cal K}^*-\boldsymbol{w}_{\mathcal{M}}^*\rangle \nonumber \\
 && +\|\boldsymbol{w}_{\cal K}^*-\boldsymbol{w}_{\mathcal{M}}^*\|^2 \nonumber \\
&=& 2\langle{\bar{\boldsymbol{w}}_{{\cal M},t}}-\boldsymbol{w}_{\mathcal{M}}^*,\boldsymbol{w}_{\cal K}^*-\boldsymbol{w}_{\mathcal{M}}^* \rangle{-\|\boldsymbol{w}_{\cal K}^*-\boldsymbol{w}_{\mathcal{M}}^*\|^2}
\label{2ab}\\
&\geq&{-{\mathbb{E}{||\bar{\boldsymbol{w}}_{{\cal M},t}}-\boldsymbol{w}_{\mathcal{M}}^*||^2}}-2{{||\boldsymbol{w}_{\cal K}^*-\boldsymbol{w}_{\mathcal{M}}^*||^2}} \label{conc_1}
\end{eqnarray}
where (\ref{2ab}) comes from the fact that $2ab \geq -a^2-b^2$.

Combining Lemma \ref{Lemma_convergence_rate} and equation (\ref{conc_1}), we have
\begin{eqnarray}
\mathbb{E}(||{\bar{\boldsymbol{w}}_{{\cal M},t}}-\boldsymbol{w}_{\cal K}^*||^2) &\leq& 2\Delta + 2 {||\boldsymbol{w}_{\cal K}^*-\boldsymbol{w}_{\mathcal{M}}^*||^2}.
\label{eq_gapMK}
\end{eqnarray}

Combining (\ref{eq_gapMK}) and the $L$-smooth property of $F$, we can obtain
\begin{eqnarray}
\mathbb{E}(F(\bar{\boldsymbol{w}}_{{\cal M},t}))-F^* &\leq&  {L \over 2} \mathbb{E}(||{\bar{\boldsymbol{w}}_{{\cal M},t}}-\boldsymbol{w}_{\cal K}^*||^2) \\
&\leq& L\Delta+L||\boldsymbol{w}_{\cal K}^*-\boldsymbol{w}_{\mathcal{M}}^*||^2. \label{conc_theorem1_2}
\end{eqnarray}

This concludes the proof.

\section{Proof of Lemma 1}\label{proofLemma1}

Suppose $n_k = n$ for each $k \in \mathcal{K}$ and changing $n$ does not affect the ranking of model training time among edge servers, we rewrite the objective function (\ref{approximate_pro}) as follows:
\begin{eqnarray*}
\mathbb{E}[\tilde{c}(e, n, \mathcal{M})]=\frac{\zeta+\alpha_{\tilde{M}} e n+\beta_{\tilde{M}} e+u}{e\left(\epsilon-{D}_{\mathcal{M}}\right)}\left(\frac{{A}}{n}+{B} e^{2}+{C}_{\mathcal{M}}\right).
\end{eqnarray*}

Let us first prove the above objective function is convex over $e$ for the given $n$ and $\mathcal{M}$.

We rewrite the objective function into the following simplified form:
\begin{eqnarray}
  \mathbb{E}[\tilde{c}(e, n, \mathcal{M})] &=& (A'+B' e)({C' \over e}+D' e),
  \label{eq_approxobj}
\end{eqnarray}
where $A'=\frac{\zeta+u}{\left(\epsilon-{D}_{\mathcal{M}}\right)}$, $B'=\frac{n\alpha_{\tilde{M}} +\beta_{\tilde{M}}}{\left(\epsilon-{D}_{\mathcal{M}}\right)}$, $C^{\prime} \triangleq \frac{{A}}{n}+{C}_{\mathcal{M}}$, and $D'={B}$ can be considered as positive constants.

To prove the convexity of (\ref{eq_approxobj}), we take the second order derivative of $\mathbb{E}[\tilde{c}(e, n, \mathcal{M})]$ over $e$ as follows:
\begin{eqnarray}
\frac{\partial^{2} \mathbb{E}[\tilde{c}(e, \boldsymbol{n}, \mathcal{M})]}{\partial^{2} e}=2 B^{\prime} D^{\prime}+2 \frac{A^{\prime} C^{\prime}}{e^{3}}>0,
\end{eqnarray}

We can therefore claim (\ref{eq_approxobj}) is a convex function over $e$.

Let us now prove the objective function in (\ref{approximate_pro}) is convex over $n$ under the given $e$ and $\mathcal{M}$. Similarly, we can rewrite the objective function into the following form:
\begin{eqnarray}
\mathbb{E}[\tilde{c}(e, n, \mathcal{M})]=\left(A^{\prime \prime}+B^{\prime \prime} n\right)\left(\frac{C^{\prime \prime}}{n}+D^{\prime \prime}\right)
\label{eq_approxobj_n}
\end{eqnarray}
where $A^{\prime \prime}=\frac{\zeta+\beta_{\tilde{M}} e+u}{e\left(\epsilon-{D}_{\mathcal{M}}\right)}$, $B^{\prime \prime}=\frac{\alpha_{\tilde{M}} }{\left(\epsilon-{D}_{\mathcal{M}}\right)}$, $C^{\prime \prime}={A}$ and $D^{\prime \prime}={B} e^{2}+{C}_{\mathcal{M}}$ can be considered as positive constants. By taking the second derivative over (\ref{eq_approxobj_n}), we can have
\begin{eqnarray}
\frac{\partial^{2} \mathbb{E}[\tilde{c}(e, n, \mathcal{M})]}{\partial^{2} n}=2 \frac{A^{\prime \prime} C^{\prime \prime}}{n^{3}}>0,
\end{eqnarray}
which indicates the function in (\ref{eq_approxobj_n}) is also convex of $n$.

According to the definition of biconvexity, we can claim that the objective function (\ref{approximate_pro}) is biconvex over $e$ and $n$. This concludes the proof.

\section{Proof of Lemma 2}\label{ProofLemma2}

Let us first define $t_k \leq t$ as the most recent iteration of each edge server $k$ that is updated by the global model coordination, i.e., $t_k \in \mathcal{I}^k$ and $t-t_k \leq e_k$.

We can then decompose term $\|\bar{\boldsymbol w}_t-{\boldsymbol w}_{k,t}\|^2$ into the following form:
\begin{eqnarray}
  \|\bar{\boldsymbol w}_t-{\boldsymbol w}_{k,t}\|^2 &\leq& 3(\|{\boldsymbol w}_{k,t} -{\boldsymbol w}_{k,t_k}\|^2+ \|{\boldsymbol w}_{k,t_k}-\bar{\boldsymbol w}_{t_k}\|^2 \nonumber \\
  &&+\|\bar{\boldsymbol w}_{t_k}-\bar{\boldsymbol w}_t\|^2),
  \label{eq_wtminuswk}
\end{eqnarray}
where ${\bar{\boldsymbol w}}_t \triangleq {\boldsymbol w}_0 - \sum_{k=1}^{K} p_k \sum_{j=0}^{t-1}\eta_j{\boldsymbol g}_{k,j}$.

We first analyze  the first  term $\|{\boldsymbol w}_{k,t} -{\boldsymbol w}_{k,t_k}\|^2$ on the left-hand-side of equation (\ref{eq_wtminuswk}) which characterizes the local model training since last global model coordination at iteration $t_k$. 
We can derive the following bound for $\|{\boldsymbol w}_{k,t} -{\boldsymbol w}_{k,t_k}\|^2$ as follows:
\begin{eqnarray}
  \|{\boldsymbol w}_{k,t} -{\boldsymbol w}_{k,t_k}\|^2 &=&\|\sum_{j=t_k}^{t-1}\eta_j \nabla {\boldsymbol g}_{k,j}\|^2  \label{eq_wtminuswk_sgd}\\
  &\leq& \eta_{t_k}^2 (t-t_k)^2 G^2 \label{eq_wtminuswk_assum1}\\
  &\leq& \eta_{t_k}^2 e_k^2 G^2 \label{eq_wtminuswk_assum2}\\
  &\leq& 4\eta_{t}^2 e_k^2 G^2 \label{eq_wtminuswk_assum3}
\end{eqnarray}
where (\ref{eq_wtminuswk_sgd}) is obtained by the model updating rule in (\ref{eq_FLModelUpdate}), (\ref{eq_wtminuswk_assum1}) is obtained by Assumption 1, (\ref{eq_wtminuswk_assum2}) comes from tha fact that $t-t_k\leq e_k$ and $\eta_t$ is non-increasing, and (\ref{eq_wtminuswk_assum3}) is based on the property of learning rate 
$\eta_t \leq 2 \eta_{H+\tau+t}$ for all $t>0$.

For the second term on the left-hand-side of equation (\ref{eq_wtminuswk}), we can derive the following bound: 
\begin{eqnarray}
\|\bar{\boldsymbol w}_t-\bar{\boldsymbol w}_{t_k}\|^2 &=& \|\sum_{h=1}^{K} p_h \sum_{j=t_k}^{t-1}{\eta_j {\boldsymbol g}_{k,j}}\|^2 \label{eq_barwtminuswtk_def}\\
&\leq& \eta_{t_k}^2 e_k^2 G^2 \label{eq_barwtminuswtk_assum1}\\
&\leq& 4\eta_{t}^2 e_k^2 G^2 \label{eq_barwtminuswtk_assum2}
\end{eqnarray}
where (\ref{eq_barwtminuswtk_def}) is the definition of virtual sequence, (\ref{eq_barwtminuswtk_assum1}) and (\ref{eq_barwtminuswtk_assum2}) follow the same line as (\ref{eq_wtminuswk_assum2}) and (\ref{eq_wtminuswk_assum3}).

The third term on the left-hand-side of equation (\ref{eq_wtminuswk}) captures the divergence between the virtual sequence and  global model at the coordinator. We can write
\begin{eqnarray}
\lefteqn{\|{\boldsymbol w}_{k,t_k}-\bar{\boldsymbol w}_{t_k}\|^2=\|{\boldsymbol v}_{k,t_k}-\bar{\boldsymbol w}_{t_k}\|^2} \label{eq_wktkminuswtk_v}\\
&&=\|\sum_{h=1}^{K} p_h \sum_{j \in \mathcal{W}_{k,t_k}^{h}}\eta_j {\boldsymbol g}_{h,j}-\sum_{h=1}^{K} p_h \sum_{j=0}^{t_k-1}\eta_j  {\boldsymbol g}_{h,j}\|^2 \label{eq_wktkminuswtk_def}\\
&&\leq 16\eta_{t}^2 \tau^2 G^2. \label{eq_wktkminuswtk_constraint}
\end{eqnarray}
where (\ref{eq_wktkminuswtk_v}) is based on the model updating rule in (\ref{eq_FLModelUpdate}), (\ref{eq_wktkminuswtk_def}) follows the definition of ${\boldsymbol v}_{k,t_k}$ and $\bar{\boldsymbol w}_{t_k}$, (\ref{eq_wktkminuswtk_constraint}) is according to constraint in (\ref{eq_Staleness}) and the property of learning rate used in (\ref{eq_wtminuswk_assum3}).   

Combining (\ref{eq_wtminuswk_assum3}), (\ref{eq_barwtminuswtk_assum2}), and (\ref{eq_wktkminuswtk_constraint}), we have
\begin{eqnarray*}
\|\bar{\boldsymbol w}_t-{\boldsymbol w}_{k,t}\|^2 &\leq& 3\left({\|{\boldsymbol w}_{k,t} -{\boldsymbol w}_{k,t_k}\|^2 + \|{\boldsymbol w}_{k,t_k}-\bar{\boldsymbol w}_{t_k}\|^2} \right.\\
&& \left.{ +\|\bar{\boldsymbol w}_{t_k}-\bar{\boldsymbol w}_t\|^2 }\right)  \leq 8\eta_{t}^2 G^2 (2\tau^2+e_k^2).
\end{eqnarray*}

We can also obtain the following results:
\begin{eqnarray*}\label{lemma2_con}
\mathbb{E}\left[\sum_{k=1}^{K}p_k \|\bar{\boldsymbol w}_t-{\boldsymbol w}_{k,t}\|^2\right] &\leq& 24 \eta_{t}^2 G^2 \left(2\tau^2+\sum_{k=1}^{K}p_k e_k^2\right).
\end{eqnarray*}

This concludes the proof.

\section{Proof of Theorem 2}\label{ProofTheorem2}
To derive the convergence upper bound of TS-FL-ASC, we derive the convergence of $\mathbb{E}\left[\|\bar{{\boldsymbol w}}_{t+1}-{\boldsymbol w}_{\mathcal{K}}^*\|^2\right]$.

Let us first introduce the following result which has already been proved in \cite{li2019convergence}. 

\begin{lemma}\cite[Lemma 1, 2]{li2019convergence}
\label{Lemma_ConvergenceStaleness}
Suppose the following assumptions hold: (1) $F_1,F_2,...,F_K$ are all $L-$smooth and $\mu-$convex, i.e., $ {\mu \over 2}\|{\boldsymbol w}-{\boldsymbol w}'\|^2$ $\leq$ $F_k(\boldsymbol w)-F_k({\boldsymbol w}')$ $+$ $({\boldsymbol w}-{\boldsymbol w}')^T \nabla F_k({\boldsymbol w}')$ $\leq$ ${L \over 2}\|{\boldsymbol w}-{\boldsymbol w}'\|^2$ for all ${\boldsymbol w},{\boldsymbol w}'\in \mathbb{R}^d$; (2) data sample $x_k$ is uniformly randomly sampled from ${\cal D}_k$; and (3) the stochastic gradient satisfies $\mathbb{E}\|\nabla l_k({\boldsymbol w},x_k)\|^2 \leq G^2$ and $\mathbb{E}\|\nabla l_k({\boldsymbol w},x_k)-\nabla F_k({\boldsymbol w})\|^2$ $\leq$ $\sigma_k^2$. 
By setting $\kappa={L \over \mu}$, $\gamma=\max \{8\kappa,\tau+H\}$ and $\eta_t={2 \over {\mu(\gamma+t)}}$, we have
\begin{eqnarray*}
\mathbb{E}\left[\|\bar{\boldsymbol w}_{t+1}-\boldsymbol{w}_{\cal K}^*\|^2\right] &\leq& (1-\eta_t \mu)\mathbb{E}\left[\|\bar{\boldsymbol w}_{t}-\boldsymbol{w}_{\cal K}^*\|^2\right] \\
&& +\eta_t^2 \sum_{k=1}^{K}{{p_k^2\sigma_k^2 \over n_k}}+6\eta_t^2 L D_{\cal K}^*(F) \nonumber \\
&& +2\mathbb{E}\left[{\sum_{k=1}^{K}p_k \|\bar{\boldsymbol w}_t-{\boldsymbol w}_{k,t}\|^2}\right],
\end{eqnarray*}
\end{lemma}

In Lemma \ref{lemma_proof}, we have derived a bound for the term $\mathbb{E}\left[{\sum_{k=1}^{K}p_k \|\bar{\boldsymbol w}_t-{\boldsymbol w}_{k,t}\|^2}\right]$. Substituting (\ref{lemma2_con2}) into Lemma \ref{Lemma_ConvergenceStaleness}, we have
\begin{eqnarray}
\mathbb{E}\|\bar{\boldsymbol w}_{t+1}-\boldsymbol{w}_{\cal K}^*\|^2 \leq (1-\eta_t \mu)\mathbb{E}\|\bar{\boldsymbol w}_{t}-\boldsymbol{w}_{\cal K}^*\|^2+\eta_t^2 B,
\label{eq_barwtminuswK}
\end{eqnarray}
where
\begin{eqnarray}
B=\sum_{k=1}^{K}{{p_k^2\sigma_k^2 \over n_k}} +6L D_{\cal K}^*(F) +48G^2(2\tau^2+\sum_{k=1}^{K}p_k e_k^2).
\end{eqnarray}

Let $\Delta_t=\mathbb{E}\left[\|\bar{\boldsymbol w}_{t}-\boldsymbol{w}_{\cal K}^*\|^2\right]$ and $\xi = \max\{{4B \over \mu^2},\gamma\|{\boldsymbol w}_0-\boldsymbol{w}_{\cal K}^*\|^2\}$. Based on the definition of $\xi$, if $t=0$, we have $\Delta_{t} \leq {\xi \over \gamma + t}$.
%
We can then write $\Delta_{t+1}$ as the following form:
\begin{eqnarray}
\Delta_{t+1} &\leq& (1-\eta_t \mu)\Delta_t+\eta_t^2 B \label{eq_Deltatplus1}\\
&\leq& (1-{2 \over \gamma +t}){\xi \over \gamma +t}+{4 B\over \mu^2 (t+\gamma)^2} \label{eq_Deltatplus1_learningrate}\\
&=& {(\gamma+t-1)\xi \over (\gamma +t)^2}+{1 \over (\gamma +t)^2}({{4B \over \mu^2}-\xi}) \nonumber \\
&\leq& {(\gamma+t-1)\xi \over (\gamma +t)^2}  \label{eq_Deltatplus1_cal} \\
&<& {(\gamma+t-1)\xi \over (\gamma +t)^2-1} = {\xi \over \gamma +t+1}.\label{eq_Deltatplus1_finalresult}
\end{eqnarray}
where (\ref{eq_Deltatplus1}) comes directly from (\ref{eq_barwtminuswK}), (\ref{eq_Deltatplus1_learningrate}) is based on the property of learning rate defined in Theorem 2, (\ref{eq_Deltatplus1_cal}) comes from the fact that ${4B \over \mu^2} \le \xi$.

According to the definition of $\xi$, we have the inequality $\xi = \max\{{4B \over \mu^2},\gamma \Delta_0\} \leq {4B \over \mu^2}+\gamma\Delta_0$.

Substituting the above inequality into (\ref{eq_Deltatplus1_finalresult}), we have
\begin{eqnarray}\label{key_conclusion}
\mathbb{E}\|\bar{\boldsymbol w}_{t}-\boldsymbol{w}_{\cal K}^*\|^2 &\leq& {1  \over \gamma +t} ({4B \over \mu^2}+\gamma \Delta_0).
\end{eqnarray}

Using the property of $L-$smooth  of $F(\cdot)$, we can obtain
\begin{eqnarray}
\mathbb{E}[F(\bar{\boldsymbol w}_t)-F^*] &\leq& {L \over 2}\mathbb{E}\|\bar{\boldsymbol w}_{t}-\boldsymbol{w}_{\cal K}^*\|^2  \nonumber\\
&\leq& {L \over 2} {1  \over \gamma +t} ({4B \over \mu^2}+\gamma \Delta_0)\nonumber\\
&\leq& {\kappa \over \gamma +t}({2B \over \mu}+{\mu \gamma \over 2}\Delta_0)
\end{eqnarray}
where $\kappa={L \over \mu}$. This concludes the proof.

\section{Proof of Theorem \ref{Theorem_TSASC_convergence_rate}}\label{ProofTheorem}
According to Theorem 2, the model trained by a set $\cal M$ of edge servers can always converge to the optimal weight $\boldsymbol{w}_{\mathcal{M}}^*$, and we can write
\begin{eqnarray}
\mathbb{E}(||{\bar{\boldsymbol{w}}_{{\cal M},t}}-\boldsymbol{w}_{\mathcal{M}}^*||^2)&\leq& \Delta,
\end{eqnarray}
where
\begin{eqnarray}
\Delta&=&\dfrac{4}{\mu^2 (\gamma+t)}({\sum_{k \in \cal M}}{p_k^2{\sigma_k^2} \over n_k}+48G^2(2\tau^2+\sum_{k=1}^{K}p_k e_k^2) \nonumber \\
&&+6L D^*_{\cM}(F)+{\mu^2 (\gamma+1) \over 4}\|{\boldsymbol{w}}_{0}-\boldsymbol{w}_{\mathcal{M}}^*\|^2)
\label{eq_DeltaTheorem3Proof}
\end{eqnarray}

Combining (\ref{eq_DeltaTheorem3Proof}) with (\ref{conc_theorem1_2}), we have
\begin{eqnarray}
\mathbb{E}(F(\bar{\boldsymbol{w}}_{{\cal M},t}))-F^*
&\leq&L\Delta+L||\boldsymbol{w}^*-\boldsymbol{w}_{\mathcal{M}}^*||^2.
\end{eqnarray}

This concludes the proof.

\end{document}